\documentclass[twoside, 11pt]{article}

\usepackage{jmlr2e}
\usepackage{bm}
\usepackage{algorithm}
\usepackage{algorithmic}
\usepackage{graphicx}
\usepackage{subfigure}
\usepackage{multirow}%
\usepackage{amsmath, amssymb}
\usepackage{color}
\usepackage{mathrsfs}
\usepackage{upgreek}
\usepackage{helvet}
\usepackage{courier}
\usepackage{bm}
\usepackage{algorithm}
\usepackage{algorithmic}
\usepackage{graphicx}
\usepackage{subfigure}
\usepackage{multirow}%
\usepackage{amsmath, amssymb}
\usepackage{color}
\usepackage{mathrsfs}
\usepackage{epstopdf}
\usepackage{wrapfig}
\usepackage{picinpar}
\usepackage{url}
\def\aje{\textcolor{black}}

\def\bmu{\mbox{{\boldmath $\mu$}}}
\def\bvarpi{\mbox{{\boldmath $\varpi$}}}
\def\ww {\mbox{{\boldmath $\omega$}}}
\def\diag{\mbox{diag}}

\def\0{{\bf 0}}
\def\1{{\bf 1}}

\def\A{{\cal A}}

\def\K{{\bf K}}

\def\bI{{\bf I}}

\def\s{{\bf s}}

\def\bg{{\bf g}}
\def\bo{{\bf o}}

\def\D{{\mathcal D}}

\def\c{{\bf c}}
\def\u{{\bf u}}
\def\v{{\bf v}}
\def\w{{\bf w}}
\def\tw{{\bf {w}}}

\def\x{{\bf x}}
\def\y{{\bf y}}
\def\z{{\bf z}}
\def\f{{\bf f}}

\def\bnu{\mbox{{\boldmath $\nu$}}}

\def\bphi{\mbox{{\boldmath $\phi$}}}

\def\ba{\mbox{{\boldmath $\alpha$}}}

\def\bzeta{\mbox{{\boldmath $\zeta$}}}
\def\bD{\mbox{{\boldmath $\Delta$}}}

\def\bxi{\mbox{{\boldmath $\xi$}}}

\def\d{{\bf d}}

\def\a{\alpha}

\def\L{{\mathcal L}}
\def\bA{\bf A}
\def\bB{\bf B}
\def\bX{{\bf X}}
\def\mQ {\mathcal {Q}}
\def\mG {\mathcal {G}}

\def\R{{\mathbb R}}

\def\etal{{\em et al.\/}\, }

\def\ivor{\textcolor{black}}
\def\revise{\textcolor{black}}
\newtheorem{deftn}{Definition}
\newtheorem{thm}{Theorem}
\newtheorem{prop}{Proposition}

\jmlrheading{xx}{20xx}{xx-xx}{10/10}{xx/xx}{Mingkui Tan, Ivor W.
Tsang and Li Wang}

\ShortHeadings{Towards Ultrahigh Dimensional Feature Selection for
Big Data}{Tan, Tsang and Wang} \firstpageno{1}

\begin{document}

\title{Towards Ultrahigh Dimensional Feature Selection \\for  Big Data}

\author{\name Mingkui Tan \email tanmingkui@gmail.com \\
       \addr School of Computer Engineering\\
       Nanyang Technological University\\
       642274,  Singapore
       \AND
\name Ivor W. Tsang \email ivor.tsang@gmail.com \\
       \addr Center for Quantum Computation $\&$ Intelligent Systems\\
         University of
Technology, Sydney
       \AND
       \name Li Wang \email liwangucsd@gmail.com \\
       \addr Department of Mathematics\\
        University of California\\
        San Diego, USA}
\editor{Sathiya Keerthi}

\maketitle

\begin{abstract}

In this paper, we present a new adaptive feature scaling scheme for
ultrahigh-dimensional feature selection on Big Data. To solve this
problem effectively, we first reformulate it as a convex
semi-infinite programming (SIP) problem and then propose an
efficient \emph{feature generating paradigm}. \aje{In contrast with}
traditional gradient-based approaches that conduct optimization on
all input features, the proposed method iteratively activates a
group of features and solves a sequence of multiple kernel learning
(MKL) subproblems of much reduced scale. To further speed up the
training, we propose to solve the MKL subproblems in their primal
forms through a modified accelerated proximal gradient approach. Due
to such \aje{an} optimization scheme, some efficient cache techniques are
also developed. The feature generating paradigm can guarantee \aje{that the solution converges} globally under mild conditions and achieve
lower feature selection bias. Moreover, the proposed method can
tackle two challenging tasks in feature selection: 1) group-based
feature selection with complex structures and 2) nonlinear feature
selection with explicit feature mappings.  Comprehensive experiments
on a wide range of synthetic and real-world datasets \aje{containing} tens of
million data points with $O(10^{14})$ features demonstrate the
competitive performance of the proposed method over state-of-the-art
feature selection methods in terms of generalization performance and
training efficiency.

\end{abstract}


\section{Introduction}\label{Introduction}

With the rapid development of the \emph{Internet}, \emph{Big Data}\aje{, with large volumes} and ultrahigh dimensionality\aje{, has} emerged in various
machine learning applications, such as text mining and information
retrieval~\citep{deng2011hierarchical,Ping2011,Zhang2012}. For
instance, a collaborative email-spam filtering task with 16 trillion
($10^{13}$) unique features has been studied
in~\citep{Weinberger2009}. \aje{Ultrahigh-dimensional} data also widely
\aje{appears} in many nonlinear machine learning tasks. To tackle the
intrinsic nonlinearity of \emph{Big Data}, researchers \aje{have} proposed to
achieve fast training and prediction through linear techniques using
explicit feature mappings~\citep{Chang2010,Maji2009}. However, most
of the explicit feature mappings dramatically expand the
dimensionality of the data. For instance, the commonly used
$2$-degree polynomial kernel feature mapping has a dimensionality of
$O(m^2)$ where $m$ denotes the number of input
features~\citep{Chang2010}. Even with a \aje{modest} $m$, the
dimensionality of the induced feature space is very \aje{large}. Other
typical feature mappings include the spectrum-based feature mapping
for string \aje{kernels}~\citep{sonnenburg2007large,sculley2006spam},
\aje{the} histogram intersection kernel feature
expansion~\citep{wu2012efficient}, \aje{etc}.

The ultrahigh dimensionality not only incurs the \aje{large} memory
requirements and high computational \aje{costs for training} but also
deteriorates the \aje{generalizability} due to the ``curse of
dimensionality"~\citep{Duda2000,Guyon2003,Dell2006,Dasgupta2007,Blum2007}.
Fortunately, in many datasets with ultrahigh dimensions, most of the
features are irrelevant to the output. Accordingly, selecting the
most informative features and dropping irrelevant features can
vastly improve generalization performance~\citep{Ng98}. Moreover,
for ultrahigh-dimensional problems, a sparse classifier is useful
for faster predictions. \aje{Last}, in many applications such as
bioinformatics~\citep{Guyon2003}, a small number of features are
expected to interpret the results for further biological analysis.

\aje{In recent} decades,  numerous feature selection methods have
been proposed for classification
tasks~\citep{Guyon2002,chapelle2008multi}. In general, existing
feature methods can be classified into two categories\aje{:} filter
methods and wrapper methods~\citep{Kohavi97,Ng98,Guyon2002}. Filter
methods, such as the signal-to-noise ratio method~\citep{Golub1999}
and spectral feature filtering \citep{Zhao2007}, \aje{have the
advantage} of low computational requirements but \aje{cannot} identify
the optimal feature subset \aje{for} the predictive model of interest. \aje{ In contrast}, wrapper methods, which select the discriminative
features by incorporating inductive learning rules, can achieve
better performance than filter methods~\citep{XuNMMKL,Guyon2003},
but \aje{incur a much higher} computational cost. How to scale \aje{these} wrapper methods to \emph{Big Data} is a very challenging issue and is also the major focus of this paper.

As a typical wrapper method, the \aje{SVM-based} recursive feature
elimination (SVM-RFE) has shown good performance on \aje{a} gene selection
task in Microarray data analysis~\citep{Guyon2002}. Specifically,
using a recursive feature elimination scheme, SVM-RFE obtains nested
subsets of features based on the weights of \aje{the} classifier.
Unfortunately, the nested feature selection strategy is ``monotonic"
and suboptimal in identifying the most informative feature
subset~\citep{XuNMMKL,tan10}. To address this drawback,
non-monotonic feature selection methods have gained \aje{much}
attention~\citep{XuNMMKL,Chan2007}. Basically, the non-monotonic
feature selection requires the convexity of the objective such that
a global solution exists. To this end, \cite{Chan2007} proposed two
convex relaxations to the $\ell_0$-norm sparse SVM, {QSSVM}
and {SDP-SSVM}, which are solved by convex quadratically constrained
quadratic programming (QCQP) and \aje{semidefinite} programming (SDP),
respectively. The resultant models are convex, \aje{and} thus they belong to
the non-monotonic feature selection methods. However, these two
models are too expensive to \aje{solve}, especially for high
dimensional problems. \aje{In addition}, Xu \etal proposed another
non-monotonic feature selection method, namely
{NMMKL}~\citep{XuNMMKL}. Since it involves a QCQP problem with many
quadratic constraints, it is still computationally \aje{unfeasible for
high-dimensional} problems.

Focusing on the logistic loss, recently, some researchers \aje{have} proposed
\aje{selecting} features using greedy
strategies~\citep{Tewari2011,Lozano2011}, which iteratively include
one feature into a feature subset. For example,  a group orthogonal
matching pursuit ({GOMP}) is proposed in~\citep{Lozano2011}. A more
general greedy scheme is presented in~\citep{Tewari2011}. Although
promising performance has been observed, these greedy methods have
some drawbacks. \aje{First}, since only one feature is involved in each
iteration, these methods are very expensive when there are a large
number of features to be selected. More critically,  due to the
absence of \aje{an} appropriate regularizer in the objective function, \aje{overfitting may occur}, which will seriously deteriorate
the generalization performance~\citep{Lozano2011,Tewari2011}.

To avoid \aje{overfitting}, one can add a regularizer to the loss
function \aje{given} a set of labeled patterns $\{\x_{i},
y_{i}\}_{i=1}^{n}$ where $\x_{i}\in \R^{m}$ is an instance of $m$
dimensions and $y_{i}\in\{\pm1\}$ is the output label. To select the
most important features, we can learn a sparse decision function
$d(\x)=\w'\x$ by solving \aje{the following}:
\begin{eqnarray}
\label{eq:l0} \min_\w~~\|\w\|_{0}+
C\sum\limits_{i=1}^{n}l(-y_{i}\w'\x_{i}),
\end{eqnarray}
where  {$\w \in \R^{m}$} is the weight vector, $\|\w\|_{0}$ denotes
the $\ell_0$-norm that counts the number of nonzeros in $\w$,
$l(\cdot)$ is a convex loss function, and $C>0$ is a regularization
parameter. Unfortunately, solving this problem is NP-hard due to the
$\ell_0$-norm regularizer. Instead, many researchers resort to
learning a sparse decision rule through an $\ell_1$-convex
relaxation \aje{as follows}~\citep{Bradley1998,Zhu2003L1svm,Fung2004}:
\begin{eqnarray}\label{eq:ell_one}
\label{eq:l1} \min_\w~~\|\w\|_{1}+
C\sum\limits_{i=1}^{n}l(-y_{i}\w'\x_{i}),
\end{eqnarray}
where  $\|\w\|_{1}  = \sum_{j=1}^{m}|w_j|$ is the $\ell_1$-norm on
$\w$. The $\ell_1$-regularized problem can be efficiently solved
because of its convexity. Recently, many optimization methods have
been proposed to solve \aje{this} including Newton
methods~\citep{Fung2004}, proximal gradient
methods~\citep{Yuan2011},  coordinate descent
methods~\citep{Yuan2010jmlr,Yuan2011}, \aje{etc}. Interested readers
can find more details of these methods
in~\citep{Yuan2010jmlr,Yuan2011} and references therein. Moreover,
recently, \aje{much} attention has been paid \aje{to} online learning methods
and stochastic gradient descent ({SGD}) methods for dealing with
\emph{Big Data}
challenges~\citep{xiao2009dual,duchi2009efficient,langford2009sparse,Shwartz2013}.

However, there are several deficiencies regarding the $\ell_1$-norm
regularized model and existing $\ell_1$-norm methods. \aje{First}, since
the $\ell_1$-norm regularization shrinks the regressors, feature
selection bias \aje{will inevitably exist in} $\ell_1$-norm
methods~\citep{zhang2008sparsity,ZhangJMLR2010,Lin2010,zhang2010nearly}.
Let $L(\w) = \sum_{i=1}^{n}l(-y_{i}\w'\x_{i})$ be the empirical loss
on the training data. Then\aje{,} $\w^*$ is an optimal solution to
(\ref{eq:ell_one}) if and only if it satisfies the following
optimality conditions~\citep{Yuan2010jmlr}:
\begin{equation}\label{eq:w_cond}
  \left\{
   \begin{array}{ll}
   \nabla_j  L(\w^*) = - {1}/{C}   &\text{if} ~~w_j^*>0,  \\
   \nabla_j  L(\w^*) = {1}/{C}   &\text{if} ~~w_j^*<0,  \\
    -{1}/{C} \leq \nabla_j  L(\w^*) \leq  {1}/{C}   &\text{if} ~~w_j^*=0.\\
   \end{array}
  \right.
\end{equation}
According to the above  conditions, one can achieve different levels
of sparsity by changing the regularization parameter $C$.
Specifically,  with a small $C$, minimizing $\|\w\|_1$ in
(\ref{eq:l1}) would favor selecting only a few features. However,
the sparser the solution, the larger the predictive risk (or
empirical loss)~\citep{Lin2010}. In an extreme case where C
is chosen to be \aje{very small (close to zero)}, none of the features
will be selected according to the condition (\ref{eq:w_cond}), which
leads to a poor prediction model. To avoid this problem, we can
learn an accurate prediction model using a larger $C$ (to reduce the
empirical loss), which, however, will include more features
according to (\ref{eq:w_cond}). In other words, sparsity and
unbiased solutions cannot be achieved  simultaneously  in
(\ref{eq:ell_one}) by changing the tradeoff parameter $C$.
Following~\citep{figueiredo2007gradient,ZhangJMLR2010}, one remedy
is to \aje{perform de-biasing} with the detected features by \aje{retraining}, which
is equivalent to \aje{setting} $C$ to $\infty$. However, such \aje{de-biasing
methods that involve many times of retraining} are not efficient.
Moreover, when tackling \emph{Big Data} \aje{with} ultrahigh dimensions, the
$\ell_1$-regularization \aje{will} be inefficient or \aje{unfeasible}. For
example, when the dimensionality is \aje{approximately} $10^{12}$, one needs
\aje{approximately} 1 TB \aje{of} memory to store the weight vector $\w$, which is
intractable for existing $\ell_1$-methods including online learning
methods and {SGD} methods~\citep{langford2009sparse,Shwartz2013}.
\aje{Last}, due to the scale variation of $\w$, it is also \aje{nontrivial}
to control the number of features \aje{while regulating} the decision
function.

In \citep{tan10}, the conference version of this paper,  an
$\ell_0$-norm sparse SVM model is introduced. Its nice optimization
scheme has brought significant benefits to several applications,
namely image retrieval~\citep{Mohammad2011}, multi-label prediction~\citep{Quanquan2011}, feature selection for
multivariate performance measures~\citep{Mao2013}, feature
selection for logistic regression~\citep{Tan2013logistic},
 and graph-based feature
selection~\citep{Quanquan2011UAI}. However, several issues \aje{remain}
to be solved. \aje{First}, the tightness of the convex relation
 remains unclear. \aje{Second}, the adopted
optimization strategy is incapable of dealing with very large-scale
and \aje{ultrahigh-dimensional} problems. \aje{Third}, the presented feature
selection strategy is limited to linear feature selections. However,
in many applications, one needs to tackle features with complex
structures.

Regarding the above issues, in this paper, we propose an adaptive
feature scaling (AFS) for feature selection by introducing a
continuous feature scaling vector $\d \in [0,1]^m$. To enforce sparsity, we impose an explicit $\ell_1$-constraint $||\d||_1\leq
B$, where the scalar $B$ represents the least number of features to
be selected. The solution to the \aje{resulting} optimization problem is
 \aje{nontrivial} due to the additional constraint. Fortunately, by
transforming it \aje{into} a convex semi-infinite programming (SIP) problem,
an efficient optimization scheme can be developed. In summary, this
paper makes the following extensions and improvements.
\begin{itemize}
\item
 A {Feature Generating Machine} ({FGM}) is
proposed to efficiently solve the SIP problem.\footnote{The C++ and
MATLAB source codes \aje{for} the proposed methods are publicly available
at: \url{http://c2inet.sce.ntu.edu.sg/Mingkui/robust-FGM.rar.}}
Instead of performing the optimization on all input features, \aje{this method} iteratively {infers} the most informative features and then solves a
reduced subproblem using multiple kernel learning (MKL) methods \aje{in which} each base kernel is defined on a set of the most informative
features.
\item
A major advantage of this scheme is that the feature selection bias
in the $\ell_1$-regularized methods can be alleviated by separately
controlling the complexity and sparsity of the decision function.
Particularly, the proposed optimization scheme mimics the
\aje{retraining} strategy to reduce the feature selection bias with
little effort.
\item
To speed up the training on \emph{Big Data}, we  propose to solve
the primal form of the MKL  subproblem through a modified
accelerated proximal gradient method.  Accordingly, the \aje{large} memory
requirement and heavy computational \aje{costs} can be significantly
reduced. The convergence rate of the modified APG is provided.
Several cache techniques are proposed to further enhance the
efficiency.
\item
The feature generating paradigm is also extended to group feature
selection with complex structures and nonlinear feature selection
with  explicit feature mappings.
\end{itemize}

The remainder of this paper is organized as follows. In
Section~\ref{sec:formulation}, we start by presenting the adaptive
feature scaling (AFS) for the linear feature selection task, group
feature selection task and the corresponding convex reformulations.
After that, in Section~\ref{sec:MKL}, we present the feature
generating machine (FGM) to solve the \aje{resulting} optimization
problems, which \aje{include} two core steps\aje{. These are} the worst-case
analysis step and the subproblem optimization step. In Section
\ref{sec:worst_case}, we illustrate the details of the worst-case
analysis for a number of learning tasks \aje{including} feature
selection with complex group structures and nonlinear feature
selection with explicit feature mappings. We detail the subproblem
optimization in Section \ref{sec:learn}. In Section
\ref{sec:kernel}, we extend FGM to do nonlinear feature selection
with kernels. The connections \aje{to} related studies are discussed in
Section~\ref{sec:related}. We conduct comprehensive experiments in
Section~\ref{sec:exp} and conclude this work in
Section~\ref{sec:Discussion}.

\section{Feature Selection Through Adaptive Feature Scaling}\label{sec:formulation}

Throughout \aje{this} paper,  we denote the transpose of \aje{a} vector/matrix by
the superscript $'$, a vector with all entries equal to one as
$\textbf{1}\in \R^{n}$ and the zero vector as $\textbf{0}\in
\R^{n}$. In addition, we denote a dataset by ${\bX} = [{{\x}_1},
..., {{\x}_n}]' = [{\f}^1, ..., {\f}^m]$, where ${\x}_i \in \R^m$
represents the $i$th instance and $\f^j \in \R^n$ denotes the $j$th
feature vector. We use $|\mG|$ to denote cardinality of an index set
$\mG$ and $|v|$ to denote the absolute value of a real number $v$.
For simplicity, we denote $\v \succeq \a$ if $v_i\geq \a, \forall i$
and $\v \preceq \a$ if $v_i\leq \a, \forall i$. We also denote
$\|\v\|_p$ as the $\ell_p$-norm of a vector and $\|\v\|$ as the
$\ell_2$-norm of a vector. Given a vector ${\v} = [\v_1', ...,
\v_p']'$ where $\v_i$ denotes a sub-vector of $\v$, we denote
$\|\v\|_{2,1} = \sum_{i=1}^{p}\|\v_i\|_2$ as the mixed
$\ell_1$/$\ell_2$ norm~\citep{Bach2010convex} and $\|\v\|_{2,1}^2 =
(\sum_{i=1}^{p}\|\v_i\|_2)^2$.  Accordingly, we call
$\|\v\|_{2,1}^2$ an $\ell_{2,1}^2$ regularizer. Following
\citep{Rakotomamonjy2008}, we define $\frac{x_i}{0} = 0$ if $x_i=0$
and $\infty$ otherwise. \aje{Last}, {$\bA\odot \bB$ represents} the
element-wise product between two matrices $\bA$ and $\bB$.

\subsection{A New AFS Scheme for Feature
Selection}\label{sec:linear} In SVM,  we learn a linear decision
function $d(\x)=\w'\x-b$ by solving the following problem:
\begin{eqnarray}\label{eq:struct_risk}
\min_\w ~~\frac{1}{2}\|\w\|^2+
C\sum\limits_{i=1}^{n}l(-y_{i}(\w'\x_{i}-b)),
\end{eqnarray}
where $\w=[w_1, \ldots,  w_m]'\in \R^m$ denotes the weight of the
decision hyperplane, $b$ denotes the shift  from the origin, $C>0$
represents the regularization parameter and $l(\cdot)$ denotes a
convex loss function. In this paper, we concentrate on the squared
hinge loss \[l(-y_{i}(\w'\x_{i}-b)) =
\frac{1}{2}\max(1-y_{i}(\w'\x_{i}-b),0)^2\] and the logistic loss
\[l(-y_{i}(\w'\x_{i}-b)) = \log(1+\exp(-y_{i}(\w'\x_{i}-b))).\]

The $\ell_2$-regularizer $||\w||^2$, which is used to avoid the
\aje{overfitting} problem~\citep{liblinear}, cannot induce \aje{a} sparse
solution. To address this issue, we introduce a feature scaling
vector $\d \in [0, 1]^{m}$ to scale the importance of features.
Given an instance $\x_i$, we impose $\sqrt{\d} =[\sqrt {d_1},
\ldots, \sqrt{d_m}]'$ on its features~\citep{Vishy10} \aje{resulting in
the following rescaled instance:}
\begin{eqnarray}\label{eq:d}
\widehat{\x}_{i} = (\x_{i}  \odot \sqrt\d).
\end{eqnarray} With this
scaling scheme, the $j$th feature is selected if and only if $d_j>
0$. Similar feature scaling \aje{schemes} can be found
in~\citep{Weston2000,Chapelle2002,Grandvalet2002,Alain2003,Varma09,Vishy10}.
However, unlike these scaling schemes, $\d$ in (\ref{eq:d}) is
bounded in $[0,~1]^m$.

In many real-world applications, one may intend to select a desired
number of features with acceptable generalization performance. For
example, in Microarray data analysis, due to expensive
\aje{biodiagnosis} and limited resources, biologists prefer to select
less than 100 genes from hundreds of thousands of
genes~\citep{Guyon2002,Golub1999}. Motivated by such prior
knowledge, to select features, we explicitly constrain the
$\ell_1$-norm of $\d$ by \aje{the following}:
\begin{eqnarray}\label{eq:l1d}
  \sum\limits_{j=1}^{m}d_{j} = ||\d||_1 \leq B, ~d_{j}\in [0, 1], ~j=1,
\cdots, m,
\end{eqnarray}
where the integer $B$ represents the least number of features to be
selected. Let $\mathcal{D}=\big\{\d\in \mathbb{R}^m
\big|\sum_{j=1}^{m}d_{j}\leq B, ~d_{j}\in [0, 1], ~j=1, \cdots,
m\big\}$ be the domain of $\d$. Note that the compact domain
$\mathcal{D}$ contains \aje{an} infinite number of elements. \aje{Herein, for simplicity, we concentrate on} the squared hinge loss. Accordingly, the
proposed AFS can be formulated as \aje{follows}:
\begin{eqnarray}\label{eq:fgssvm}
\min\limits_{\d\in\D }\min\limits_{\tw,
\bxi,b}&{}&\frac{1}{2}\|\tw\|_{2}^{2}+
{\frac{C}{2}\sum\limits_{i=1}^{n}\xi^{2}_{i}}
\\ \text{s.t.} &{}&  {y_{i}\left(\tw'(\x_{i}  \odot \sqrt\d)-b\right)  \geq 1 -\xi_{i}}, \;\; \;\;i=1, \cdots, n, \nonumber
\end{eqnarray}
where $C$ is the regularization parameter that trades off between
the model complexity and the fitness of the decision function, and
$b/\|\tw\|$ determines the offset of the hyperplane from the origin
along the normal vector $\w$. This problem is a non-convex
optimization problem. For a fixed $\d$, it indexes a set of
features. Accordingly, the inner minimization problem \aje{with respect to} $\tw$ and $\bxi$ is a standard SVM problem \aje{as follows}:
\begin{eqnarray}\label{eq:fgssvm:appendix}
\min\limits_{\tw, \bxi, b}&{}&\frac{1}{2}\|\tw\|_{2}^{2}+
{\frac{C}{2}\sum\limits_{i=1}^{n}\xi^{2}_{i}}
\\ \text{s.t.} &{}&  {y_{i}\left(\tw'(\x_{i}  \odot \sqrt\d)-b\right)  \geq 1 -\xi_{i}}, \;\; \;\;i=1, \cdots, n, \nonumber
\end{eqnarray}
which can be solved in its dual form. By introducing the Lagrangian
multiplier $\alpha_i \geq 0$ to each constraint
${y_{i}\left(\tw'(\x_{i} \odot \sqrt\d)-b\right)  \geq 1 -\xi_{i}}$,
the Lagrangian function is \aje{as follows}:
\begin{eqnarray}\label{eq:fgssvm:lag}
\L(\w, \bxi,b, \ba) = \frac{1}{2}\|\tw\|_{2}^{2}+
\frac{C}{2}\sum\limits_{i=1}^{n}\xi^{2}_{i}
-\sum_{i=1}^{n}{\alpha_i} \left(y_{i}\left(\tw'(\x_{i} \odot
\sqrt\d)-b\right) - 1+ \xi_{i}\right).
\end{eqnarray}
By setting the derivatives of $\L(\w, \bxi,b, \ba)$ w.r.t. $\w$,
$\bxi$ and $b$  to $\0$, we get \aje{the following:}
\begin{eqnarray}\label{eq:relation}
\w = \sum_{i=1}^{n}{\alpha_i} {y_{i}(\x_{i} \odot \sqrt\d)}, \ba =
C\bxi,~\text{and}~\sum_{i=1}^n \alpha_i y_i = 0.
\end{eqnarray}
 Substituting these
results into (\ref{eq:fgssvm:lag}), we obtain the dual form of
(\ref{eq:fgssvm:appendix}) as \aje{follows}:
\begin{eqnarray}\label{eq:fgssvm:subdual}
\max_{\ba\in
 \mathcal{A}}~~ -\frac{1}{2}
\bigg\|\sum_{i=1}^{n}\alpha_{i}y_{i}(\x_{i}\odot
 \sqrt\d)\bigg\|^{2}-\frac{1}{2C}\ba'\ba + \ba'\1,
\end{eqnarray}
where $\A = \{\ba|\sum_{i=1}^n \alpha_i y_i = 0, \ba\succeq \0\}$ is
the domain of $\ba$. For convenience, let $\c(\ba) =
\sum_{i=1}^{n}\alpha_{i}y_{i}\x_{i} \in \R^m$\aje{; then}, we have
$\bigg\|\sum_{i=1}^{n}\alpha_{i}y_{i}(\x_{i}\odot
\sqrt\d)\bigg\|^{2} = \sum_{j=1}^{m} d_j [c_j(\ba)]^2$ where
$c_j(\ba)$ denotes the $j$th coordinate of $\c(\ba)$ and is a
function of $\ba$. For simplicity, let $$f(\ba, \d)=\frac{1}{2}
\sum_{j=1}^{m} d_j [c_j(\ba)]^2+\frac{1}{2C}\ba'\ba - \ba'\1.$$
Apparently,  $f(\ba, \d)$ is linear in $\d$ and concave in $\ba$,
and both $\mathcal{A}$ and $\mathcal{D}$ are compact domains.
Problem (\ref{eq:fgssvm}) can be equivalently reformulated as the
following problem:
\begin{eqnarray}\label{eq:fgssvm_dual}
 \min\limits_{\d\in \mathcal{D}}\max\limits_{\ba\in
 \mathcal{A}}~~-f(\ba, \d),
\end{eqnarray}
However, this problem is still difficult to \aje{address}. According
to the minimax saddle-point theorem~\citep{Sion1958}, we immediately
have the following relation.
\begin{thm}\label{thm:minimax}
Since both $\mathcal{A}$ and $\D$ are convex compact sets, according
to the minimax saddle-point theorem~\citep{Sion1958} the following
equality holds by interchanging the order of $\min_{\d\in \D}$ and
$\max_{\ba \in \mathcal{A}}$ in (\ref{eq:fgssvm_dual})\aje{:}
\begin{eqnarray}\label{eq:minimax}
\min\limits_{\d\in \mathcal{D}}\max\limits_{\ba\in
 \mathcal{A}}~~-f(\ba, \d) =
\max\limits_{\ba\in
 \mathcal{A}}\min\limits_{\d\in \mathcal{D}}~~-f(\ba, \d).
\end{eqnarray}
\end{thm}
Based on the above equality, hereafter we address
(\ref{eq:fgssvm_dual}) by solving the following  minimax problem
instead:
\begin{eqnarray}\label{eq:fgssvm_dual2}
\min\limits_{\ba\in \mathcal{A}}\max\limits_{\d\in
\mathcal{D}}~~f(\ba, \d).
\end{eqnarray}

\subsection{AFS  for Group Feature
Selection} \label{sec:cutting}

The  {AFS} scheme for linear feature selection can be easily
extended for group feature \aje{selection} where the features are
organized by a group structure $\mG = \{ \mG_1, ..., \mG_p \}$\aje{. Here, }$p=|\mG|$ denotes the number of groups and $\cup_{j=1}^{p}
\mG_j = \{1, ..., m\}$\aje{, and }$\mG_j \subset \{1, ..., m\}, j = 1,
..., p$ denotes the index set of feature supports belonging to the
$j$th group of features. In group feature selection, a feature
in one group is selected if and only if this group is
selected~\citep{yuan2006model,meier2008group}. Let $\w_{\mG_j} \in
\R^{|\mG_j|}$ and $\x_{\mG_j} \in \R^{|\mG_j|}$ be the components of
$\w$ and $\x$ related to $\mG_j$, respectively. The group feature
selection is usually achieved by solving the following non-smooth
group lasso problem~\citep{yuan2006model,meier2008group}:
\begin{eqnarray}\label{eq:group_risk}
\min_\w ~~\lambda\sum_{j=1}^p||\w_{\mG_j}||_2+
\sum\limits_{i=1}^{n}l(-y_{i}\sum_{j=1}^p\w_{\mG_j}'\x_{i}{_{\mG_j}}),
\end{eqnarray}
where $\lambda$ denotes the trade-off parameter.  To solve this
problem, many efficient algorithms have been proposed, such as
accelerated proximal gradient descent methods~\citep{Liu2010,
Jenatton2011,Bach2010convex}, block coordinate descent
methods~\citep{Qin2010,Jenatton2011} and active set
methods~\citep{Francis2009HKL,Roth2008}. However, the same issues \aje{with}
the $\ell_1$-regularization, namely\aje{,} the scalability issue for
\emph{Big Data} and feature selection bias, will \aje{occur when
performing} group feature selection via solving (\ref{eq:group_risk}).
Moreover, when dealing with groups with complex structures,  the
number of groups can be exponential in the number of features $m$.
As a result, solving (\ref{eq:group_risk}) could be very expensive.

To extend the AFS for linear feature selection to group feature
selection, we introduce a group scaling vector $\widehat
\d=[\widehat{d}_1, \ldots, \widehat{d}_p]'\in\widehat{\mathcal{D}}$
to scale the groups where
{$\widehat{\mathcal{D}}=\big\{\widehat{\d}\in \mathbb{R}^p
\big|\sum_{j=1}^{p}\widehat{d}_{j}\leq B, \widehat{d}_{j}\in [0, 1],
~j=1, \cdots, p\big\}$}. Without loss of generality, we first assume
that there \aje{are no overlapping elements} among groups, namely, $\mG_i
\cap \mG_j = \emptyset, \forall i\neq j$. Accordingly, we have $\w =
[\w_{\mG_1}', ..., \w_{\mG_p}']' \in \R^m$. By taking the shift term
$b$ into consideration, the decision function is expressed as \aje{follows}:
\begin{eqnarray}
q(\x)=\sum\limits_{j=1}^{p}
{\sqrt{\widehat{d}_j}}\w_{\mG_j}'\x_{\mG_j}-b,  \nonumber
\end{eqnarray}
Focusing on the squared hinge loss, the \aje{AFS-based} group feature
selection can be formulated as the following problem:
\begin{eqnarray}\label{eq:fgssvm_group}
\min\limits_{\widehat \d\in \widehat{\D}}\min\limits_{ \tw, \bxi, b}&&\frac{1}{2}\|\w\|_{2}^{2}+\frac{C}{2}\sum\limits_{i=1}^{n}\xi^{2}_{i} \\
\text{s.t.} &{}& y_{i} \left(\sum\limits_{j=1}^{p}
\sqrt{\widehat{d}_j}\w_{\mG_j}'\x_{i\mG_j} -b \right) \geq
1-\xi_{i}, \;\;\xi_i\geq 0, \;\;i=1, \cdots, n. \nonumber
\end{eqnarray}
With similar deductions in Section \ref{sec:linear}, the above
problem can be transformed \aje{into} the following minimax problem:
\begin{eqnarray}\label{eq:dual_group}
\min\limits_{\widehat\d\in \widehat{\D}}\max\limits_{\ba\in
\mathcal{A}}-\frac{1}{2} \sum\limits_{j=1}^{p}\widehat{d}_j
\bigg\|\sum_{i=1}^{n}\alpha_{i}y_{i}\x_{i\mG_j}\bigg\|^{2}-\frac{1}{2C}\ba'\ba
+ \ba'\1,
\end{eqnarray}
which will \aje{reduce} to the linear case if $|\mG_j| = 1, \forall j
= 1, ..., p$.  Without loss of generality, \aje{we hereafter} drop the hat
from $\widehat{\d}$ and $\widehat{\D}$. Let $$\c_{\mG_j}(\ba) =
\sum_{i=1}^{n}\alpha_{i}y_{i}\x_{i\mG_j}$$ and define
$$f(\ba, \d)=\frac{1}{2} \sum_{j=1}^{p} d_j
\left\|\c_{\mG_j}(\ba)\right\|^2+\frac{1}{2C}\ba'\ba- \ba'\1.$$
\aje{Last}, we arrive at the following unified minimax optimization
problem:
\begin{eqnarray}\label{eq:fgssvm_dual_group}
\min\limits_{\ba\in \mathcal{A}}\max\limits_{\d\in
\mathcal{D}}~~f(\ba, \d),
\end{eqnarray}
 where
${\mathcal{D}}=\big\{{\d}\in \mathbb{R}^p \big|
\sum_{j=1}^{p}{d}_{j}\leq {B},~{d}_{j}\in [0, 1], ~j=1, \cdots,
p\big\}$. When $|\mG_j| = 1, \forall j = 1, ..., p$, we have $p=m$,
and problem (\ref{eq:fgssvm_dual_group}) is reduced to problem
(\ref{eq:fgssvm_dual2}).

\subsection{Group Feature Selection with Complex Structures} \label{sec:complex}
The \aje{above AFS scheme} can be extended to deal with group features
with complex structures, such as groups with overlapping
features or even more complex structures. When dealing with groups
with overlapping features, a heuristic \aje{method} is to explicitly augment
${\bX} = [\f^1, ..., \f^m]$ to make the groups \aje{nonoverlapping} by
repeating the overlapping features. For example, suppose ${\bX} =
[\f^1, \f^2, \f^3]$ with groups $\mG = \{\mG_1, \mG_2\}$ where
$\mG_1 = \{1,2\}$ and $\mG_2 = \{2,3\}$, and $\f^2$ is an
overlapping feature. To avoid the overlapping element, we duplicate
$\f^2$ and construct an augmented dataset \aje{such that} ${\bX}_{au} = [\f^1,
\f^2, \f^2, \f^3]$. After \aje{this}, the group index sets become  $\mG_1
= \{1,2\}$ and $\mG_2 = \{3,4\}$. This
 feature augmentation strategy can be extended to groups that \aje{have} more complex structures, such as tree structures and graph
structures~\citep{Francis2009HKL}. Here, we only study tree-structured groups, which \aje{are} defined as follows.
\begin{deftn}\label{defn:tree}
Tree-structured set of
groups~\citep{Jenatton2011,Kim2020tree,Kim2012}. A super set of
groups $\mG \triangleq \{\mG_h\}_{\mG_h\in \mG}$ with $|\mG| = p$ is
said to be tree-structured in $\{1, . . . , m\}$, if $\cup \mG_h =
\{1, . . . , m\}$ and if for all $\mG_g, \mG_h \in \mG$, $(\mG_g
\cap \mG_h \neq \varnothing ) \Rightarrow (\mG_g \subseteq \mG_h
~~\textrm{or}~~ \mG_h\subseteq \mG_g) $. For such a set of groups,
there exists a (\aje{nonunique}) total order relation $\preceq$ such
that:
\begin{eqnarray}
\mG_g \preceq \mG_h \Rightarrow \{ \mG_g \subseteq \mG_h
~\textrm{or}~ \mG_g \cap \mG_h = \varnothing\}. \nonumber
\end{eqnarray}
\end{deftn}

Similar to the overlapping case, we can augment the overlapping
elements of all groups along the tree structures resulting in the
augmented dataset ${\bX}_{au} = [{\bX}_{\mG_1}, ...,
{\bX}_{\mG_p}]$ where ${\bX}_{\mG_i}$ \aje{represents} the  data columns
selected by ${\mG_i}$ and $p$ denotes the number of all possible
groups. However,  this simple idea may bring \aje{large} challenges for
optimization\aje{;} particularly when there are \aje{a large} number of overlapping groups. For instance, in graph-based group structures, the number of
groups $p$ can be exponential in $m$~\citep{Francis2009HKL}. How to
avoid this difficulty is one of the focuses of this paper.

\section{Feature Generating Machine}\label{sec:MKL}
Under the proposed AFS scheme, both the linear feature selection and
\aje{the} group feature selection can be cast as the minimax problem
(\ref{eq:fgssvm_dual_group}).  By bringing in an additional variable
$\theta\in \R$ \aje{and} following \citep{Royset}, this problem can be
further formulated as a semi-infinite programming (SIP)
problem~\citep{kelley1960cpm,Royset}:
\begin{eqnarray}\label{eq:group_unify}
\min\limits_{\ba \in \mathcal{A},\theta \in \R} \theta, ~~
\text{s.t.}~~~~\theta\geq f(\ba, \d),~~\forall~\d\in
 \mathcal{D}.
\end{eqnarray}
In (\ref{eq:group_unify}), each nonzero $\d \in \mathcal{D}$ defines
a quadratic constraint (w.r.t. $\ba$) $\theta \geq f(\ba,\d)$.
Notice that there are \aje{an} infinite number of $\d$'s in  $\mathcal{D}$.
Consequently, there are \aje{an} infinite number of constraints involved in
(\ref{eq:group_unify}) making it difficult to \aje{solve}.

\subsection{Optimization Strategies by Feature Generation}
Before solving (\ref{eq:group_unify}), we first consider its
optimality condition. Specifically, let $\mu_{h}\geq 0$ be the dual
variable for each constraint $\theta\geq f(\ba, \d)$, then the
Lagrangian function of (\ref{eq:group_unify}) can be written as \aje{follows}:
\begin{eqnarray}\label{eq:lang_dual_all}
\mathcal{L}(\theta,\ba,\textbf{$\bmu$})={\theta} - \sum_{ \d_{h}\in
\mathcal{D}}\mu_{h}\left(\theta-f(\ba, \d_{h})\right). \nonumber
\end{eqnarray}
By setting its derivative w.r.t. $\theta$ to zero,  we have
$\sum\mu_{h}=1$. Let $\mathcal{M}=\{\bmu|\sum\mu_{h}={1},
\mu_{h}\geq0,  h = 1, ..., |\mathcal{D}|\}$ be the domain of $\bmu$.
We define \aje{the following:}
$${f}_\text{m}(\ba) = \max_{\d_{h} \in \mathcal{D}}f(\ba, \d_{h}).$$
Then\aje{,} the KKT conditions of (\ref{eq:group_unify}) can be written as \aje{follows}:
\begin{eqnarray}\label{eq:fgssvm_MKL_all}
 &&\sum\limits_{\d_{h}\in \mathcal{D}}\mu_h \nabla_{\ba}f(\ba, \d_{h})
 = \0, ~~~~ \text{and} ~~~~\sum\limits_{\d_{h}\in \mathcal{D}}
 \mu_{h}=1. \\
 &&{\mu_h(f(\ba, \d_{h}) - f_m(\ba)) = 0,}~~~ \mu_h\geq 0, ~~  h = 1, ..., |\mathcal{D}|.
\end{eqnarray}
Although there are many constraints in problem
(\ref{eq:lang_dual_all}), most of them are nonactive at optimality if $\bX$ only \aje{contains} a small number of relevant features
w.r.t. the output $\y$. Actually, according to the above conditions,
we have $\mu_h = 0$ if $f(\ba, \d_{h}) < {f}_\text{m}(\ba)$, which
 induces sparsity among
$\mu_h$'s. Motivated by this observation, we design an efficient
optimization scheme \aje{that} iteratively \emph{infers} the active
constraints and then solves a subproblem of reduced scale with the
selected constraints. Accordingly, the computational burden brought
by the infinite number of constraints can be avoided. The details of
\aje{this algorithm are} presented in Algorithm~\ref{Alg:cut_algo}, which is
also known as the cutting plane
algorithm~\citep{kelley1960cpm,Mutapcic09}.

\begin{algorithm}
\caption{Cutting Plane Algorithm for Solving (\ref{eq:group_unify})
(\textbf{Outer Iterations}).} \label{Alg:cut_algo}
\begin{algorithmic}[1]
\STATE Initialize $\ba^{0}=C\textbf{1}$ and $\mathcal{C}_0 =
\emptyset$.
Set iteration index $t=1$. \STATE \label{step:meb} {Feature Inference}: \\
~~Do worst-case analysis to \emph{infer} the most violated ${\d_{t}}$ based on $\ba^{t-1}$. \\
~~Set $\mathcal{C}_t=\mathcal{C}_{t-1}\bigcup \{\d_{t}\}$. \STATE
\label{step:terminate} {subproblem Optimization}:
\\~~Solve subproblem (\ref{eq:QCQP_sub}) obtaining the optimal solution $\ba^t$ and $\mu^t$.
\STATE Let $t=t+1$. Repeat \aje{steps} 2-3 until convergence.
\end{algorithmic}
\end{algorithm}

Basically, Algorithm~\ref{Alg:cut_algo}
involves two major steps: the feature inference step (also known as
the {worst-case analysis}) and the {subproblem} optimization step.
The whole procedure iterates until \aje{particular} stopping conditions are
achieved.
Specifically, the worst-case analysis \aje{infers}
the most-violated $\d_t$ based on $\ba^{t-1}$ \aje{and adds it to} the
active constraint set $\mathcal{C}_t$. As will be shown later, in
general\aje{,} each active $\d_t \in \mathcal{C}_t$ contains $B$\textbf{
new features}. In this sense, we refer \aje{to} Algorithm~1 as the
\emph{Feature Generating Machine} (FGM). Recall that, if there is no
feature being selected,  the empirical loss is $\bxi = \1$.
Therefore, we initialize $\ba^{0} = C\1$ according to the relation
in (\ref{eq:relation}). Once an active ${\d_t}$ is identified, we
update $\ba^t$ by solving the following subproblem with the
constraints defined in $\mathcal{C}_t$:
\begin{eqnarray}\label{eq:QCQP_sub}
\min\limits_{\ba \in \mathcal{A},\theta\in \R}\theta,
~~~~\text{s.t.}~~~~ f(\ba, \d_h)- \theta\leq 0,~\forall~\d_h\in
 \mathcal{C}_t.
\end{eqnarray}
Since $|\mathcal{C}_t|=t$, problem (\ref{eq:QCQP_sub}) involves \aje{only} $t$ constraints and thus is easier to \aje{address}.

\aje{Last}, after obtaining the optimal solution $\d^*$ to
(\ref{eq:QCQP_sub}), the selected features (or the selected feature
groups) are associated with the nonzero entries in $\d^*$.

\subsection{Convergence Analysis}\label{sec:convergence}
Before the introduction of the worst-case analysis and the solution
to the subproblem, we first conduct \aje{a} convergence analysis of
Algorithm~\ref{Alg:cut_algo}.

 Without loss of generality, let $\mathcal{A}\times
\mathcal{D}$ be the  domain for problem (\ref{eq:group_unify}).  In
the $(t+1)$th iteration, we find a new constraint $\d_{t+1}$ based
on $\ba_{t}$ and add it \aje{to} $\mathcal{C}_{t}$, i.e.,
$f(\ba_{t},\d_{t+1})=\max_{\d \in \mathcal{D}} f(\ba_{t},\d)$.
Apparently, we have $\mathcal{C}_t\subseteq \mathcal{C}_{t+1}$. For
convenience, we define \aje{the following:}
\begin{eqnarray}
 \begin{array}{l}
\beta_{t} = \max\limits_{1\leq i\leq {t}} f(\ba_{t},\d_{i})=
\min\limits_{\ba\in \mathcal{A}}\max\limits_{1\leq i\leq {t}}
  f(\ba,\d_{i}).
  \end{array}
\end{eqnarray}
and
\begin{eqnarray}
\varphi_{t} = \min\limits_{1\leq j \leq {t}} f(\ba_{j},\d_{j+1}) =
\min\limits_{1\leq j \leq {t}}(\max_{\d\in
\mathcal{D}}f(\ba_{j},\d)).~~
\end{eqnarray}
  Following~\citep{YeSemi}, we have the following lemma\aje{:}
\begin{lemma}
 Let $(\alpha^{*},\theta^{*})$ be a globally optimal solution  of
(\ref{eq:group_unify}), \aje{and let }$\{\theta_{t}\}$ and $\{\varphi_{t}\}$ \aje{be defined as} above, then:
$\theta_{t}  \leq \theta^{*} \leq \varphi_{t}.$ 
With the number of \aje{iterations} $t$ increasing, $\{\theta_{t}\}$ \aje{will monotonically increase} and the sequence $\{\varphi_{t}\}$ \aje{will monotonically decrease}.
\end{lemma}

\begin{proof}
According to the definition, we have $\theta_{t}
 = \beta_t$ Moreover, $\theta^{*} \!\!\!\!~=~\!\!\!\! \min_{\ba\in \mathcal{A}}
\max_{\d\in \mathcal{D}}  f(\ba,\d)$. For a fixed feasible $\ba$, we
have $\max_{\d \in \mathcal{C}_{t}} f(\ba,\d) \leq \max_{\d \in
\mathcal{D}}  f(\ba,\d) $, then \[ \min_{\ba\in \mathcal{A}}\max_{\d
\in \mathcal{C}_{t}} f(\ba,\d) \leq \min_{\ba\in
\mathcal{A}}\max_{\d \in \mathcal{D}} f(\ba,\d), \] that is, $\theta_{t}
\leq \theta^{*}.$ On the other hand, for $\forall j=1,\cdots,k$,
$f(\ba_{j},\d_{j+1}) = \max_{\d\in \mathcal{D}}f(\ba_{j},\d)$\aje{; thus,} $(\ba_{j},f(\ba_{j},\d_{j+1}))$ is a feasible solution  of
(\ref{eq:group_unify}).  Then $\theta^{*} \leq f(\ba_{j},\d_{j+1})$
for $j=1,\cdots, {t}$, and hence we have \aje{the following:} \[ \theta^{*} \leq
\varphi_{t}= \min_{1\leq j \leq {t}} f(\ba_{j},\d_{j+1}).\] With
increasing \aje{iterations} $t$, the subset $\mathcal{C}_{t}$ \aje{will
monotonically increase. Thus,} $\{\theta_{t}\}$ \aje{will monotonically
increase,} while $\{\varphi_{t}\}$ \aje{will monotonically decrease}. \aje{QED}
\end{proof}

The following conclusion shows that FGM converges to a global
solution of (\ref{eq:group_unify}).

{\begin{thm}\label{thm:global} Assume that in Algorithm~\ref{Alg:cut_algo}, the subproblem (\ref{eq:QCQP_sub}) and the
worst-case analysis in step 2 can be solved. Let
$\{(\ba_{t},\theta_{t})\}$ be the sequence generated by Algorithm~1.
If Algorithm~1 terminates at iteration $(t+1)$, then
$\{(\ba_{t},\theta_{t})\}$ is the global optimal solution of
(\ref{eq:group_unify}); otherwise, $(\ba_{t},\theta_{t})$ converges
to a global optimal solution $(\ba^*,\theta^*)$ of
(\ref{eq:group_unify}).
\end{thm}}

\begin{proof}
We can measure the convergence of FGM by the gap difference \aje{between the}
series $\{\theta_{t}\}$ and $\{\varphi_{t}\}$. Assume in \aje{the} $t$th
iteration \aje{that} there is no update of
 $\mathcal{C}_{t}$, i.e.\aje{,} $\d_{{t}+1}~=~\arg\max_{\d\in \mathcal{D}}
f(\ba_{t},\d) \in \mathcal{C}_{t},$ then $\mathcal{C}_{t} =
\mathcal{C}_{{t}+1}$. In this case, $(\ba_{t},\theta_{t}) $ is the
globally optimal solution  of (\ref{eq:group_unify}). Actually,
since $\mathcal{C}_{t} = \mathcal{C}_{{t}+1}$ in Algorithm~1, there
will be no update of $\ba$, i.e.\aje{,} $\ba_{{t}+1} = \ba_{t}$. Then\aje{, we have the following:}
\begin{eqnarray}
&&f(\ba_{t},\d_{{t}+1}) = \max_{\d\in \mathcal{D}}f(\ba_{t},\d) =
\max_{\d\in \mathcal{C}_{t}}f(\ba_{t},\d)=\max_{1\leq i\leq {t}}
f(\ba_{t},\d_{i})=\theta_{t} \nonumber\\
&&\varphi_{t} = \min\limits_{1\leq j \leq {t}}
f(\ba_{j},\d_{j+1})\leq \theta_{t}. \nonumber
\end{eqnarray}

According to Lemma 1, we have $\theta_{t}  \leq \theta^{*} \leq
\varphi_{t}$, \aje{and} thus we have $\theta_{t} = \theta^{*} = \varphi_{t}$,
and $(\ba_{t},\theta_{t})$ is the global optimum of
(\ref{eq:group_unify}).

Suppose \aje{that} the algorithm does not terminate in finite steps. Let
$\mathcal{X}=\mathcal{A}\times [\theta_1,\theta^*]$, \aje{then} a limit point
$(\bar{\ba},\bar{\theta})$ exists for $(\ba_t, \theta_t)$ since
$\mathcal{X}$ is compact. \aje{In addition,} we also have $\bar{\theta}\leq
\theta^*$. For each $t$, let $\mathcal{X}_t$ be the feasible region
of \aje{the} $t$th subproblem, which \aje{has} $\mathcal{X}_t\subseteq
\mathcal{X}$ and $(\bar{\ba},\bar{\theta}) \in
\cap_{t=1}^{\infty}\mathcal{X}_t\subseteq \mathcal{X}$. Then\aje{,} we have
$f(\bar{\ba},\d_t)- \bar{\theta} \leq 0, ~\d_t\in C_t$ for each
given $t=1,\cdots$.

To show \aje{that} $(\bar{\ba},\bar{\theta})$ is \aje{the} global optimal of problem
(\ref{eq:group_unify}), we only need to show \aje{that}
$(\bar{\ba},\bar{\theta})$ is a feasible point of problem
(\ref{eq:group_unify}), i.e., $\bar{\theta}\geq
f(\bar{\ba},\d)~\text{for all}~~\d\in \mathcal{D}$\aje{. Thus,}
$\bar{\theta}\geq \theta^*$ and we must have $\bar{\theta} =
\theta^*$. Let $v(\ba,\theta)=\min_{\d\in \mathcal{D}}
(\theta-f(\ba,\d)) = \theta-\max_{\d\in \mathcal{D}}f(\ba,\d)$. Then
$v(\ba,\theta)$ is continuous w.r.t. $(\ba,\theta)$. By applying the
continuity property of $v(\ba,\theta)$, we have \aje{the following:}
\begin{equation*}
\begin{aligned}
v(\bar{\ba},\bar{\theta}) &=
v(\ba_t,\theta_t)+(v(\bar{\ba},\bar{\theta})-v(\ba_t,\theta_t))\\
& =  ( \theta_t-f(\ba_t,\d_{t+1}) ) +(v(\bar{\ba},\bar{\theta})-v(\ba_t,\theta_t))\\
 &  \geq ( \theta_t-f(\ba_t,\d_{t+1}) ) -(\bar{\theta}-f(\bar{\ba},\d_t)
)+(v(\bar{\ba},\bar{\theta})-v(\ba_t,\theta_t))\rightarrow 0
 ~~(\text{when}~t\rightarrow \infty),
\end{aligned}
\end{equation*}
where we use the continuity of $v(\ba,\theta)$. \aje{QED}
\end{proof}

\section{Efficient Worst-Case Analysis}\label{sec:worst_case}
According to Theorem~\ref{thm:global}, the global convergence  of
Algorithm~{\ref{Alg:cut_algo}} is dependent \aje{on} the exact solution to
the worst-case analysis. In the following, we detail the exact
worst-case analysis for several feature selection tasks.

\subsection{Worst-Case Analysis for Linear Features}\label{sec:violated}
For simplicity, \aje{we hereafter} drop the superscript $t$ from
$\ba^{t}$.  The worst-case analysis for the feature selection is to
solve the following maximization problem:
\begin{eqnarray}\label{eq:Max_d}
 \max\limits_{\d}~~\frac{1}{2}  \left\|\sum\limits_{i=1}^{n}\alpha_{i}y_{i}(\x_{i}\odot
 \sqrt{ \d} )\right\|^{2},  ~~\text{s.t.}~ \sum_{j=1}^m d_i \leq
 B, \0 \preceq \d \preceq \1.
\end{eqnarray}
In general, solving this problem is \aje{difficult}.  Recall that
${\c(\ba)}=\sum_{i=1}^{n}\alpha_{i}y_{i}\x_{i} \in \R^m$\aje{. We} have
$\|\sum_{i=1}^{n}\alpha_{i}y_{i}(\x_{i}\odot
 \sqrt\d )\|^{2}=
\|\sum_{i=1}^{n}(\alpha_{i}y_{i}\x_{i})\odot \sqrt\d
\|^{2}=\sum_{j=1}^{m}c_j(\ba)^{2} d_{j}$. Based on this relation, we
can define a {feature score} as $$s_j = [c_j(\ba)]^{2}$$ to measure
the importance of features. \aje{Then,} problem (\ref{eq:Max_d}) can be
formulated as a linear programming problem \aje{as follows}:
\begin{eqnarray}\label{eq:lin_d}
 \max\limits_{\d}~~\frac{1}{2}\sum\limits_{j=1}^{m} s_j d_j,  ~~\text{s.t.} ~~ \sum_{j=1}^m d_i \leq
 B,~~ \0 \preceq \d \preceq \1.
\end{eqnarray}
Based on this formulation, the optimal solution to this problem can
be obtained without any numeric optimization solver. To address it,
we can first find the $B$ features with the largest feature score
$s_{j}$ and then set the corresponding $d_{j}$ to 1 and the \aje{rest}
to 0. It is easy to verify that $\d$ is the optimal solution to
(\ref{eq:lin_d}). Moreover,   as long as there are more than $B$
features with $s_j>0$, we have $||\d||_0 = B$.

\subsection{Worst-Case Analysis for Group Features}
The worst-case analysis for linear feature selection  can be
trivially extended for group feature selection. Suppose that the
features are organized into groups by $\mG = \{\mG_1, ..., \mG_p\}$,
and \aje{that there are} no overlapping features among groups, namely\aje{, that} $\mG_i
\cap \mG_j = \emptyset, ~\forall i \neq j$. To infer the most-active
groups, one just needs to solve the following optimization problem
\begin{eqnarray}\label{eq:Max_d_Strture}
 \max\limits_{\d\in\mathcal{D}}~~\sum\limits_{j=1}^{p} d_j
\bigg\|\sum_{i=1}^{n}\alpha_{i}y_{i}\x_{i\mG_j}\bigg\|^{2} =
\max\limits_{\d\in\mathcal{D}} \sum\limits_{j = 1}^{p} {d}_j
\c_{\mG_j}'\c_{\mG_j},
\end{eqnarray}
where $\c_{\mG_j} = \sum_{i=1}^{n}\alpha_{i}y_{i}\x_{i \mG_j}$ for
group $\mG_j$.  Let $s_{j} = \c_{\mG_j}'\c_{\mG_j}$ be the score for
group $\mG_j$. Easily, the optimal solution to
(\ref{eq:Max_d_Strture}) can be obtained by finding the $B$ features
with the largest $s_{j}$'s and setting their $d_{j}$'s to 1 and the
\aje{rest} to 0. Notice that if $|\mG_j| = 1$, $\forall j \in \{1, ...,
p\}$ problem (\ref{eq:Max_d_Strture}) is reduced to problem
(\ref{eq:lin_d}), where $\mG = \{1, ..., m\}$ and $s_{j} =
[c_j(\ba)]^2$ for $j\in \mG$. In this sense, we unify the worst-case
analysis of these feature selection tasks in Algorithm~\ref{Alg:most_d}.

\begin{algorithm}[h]
\caption{Worst-Case Analysis. } \label{Alg:most_d}
\begin{algorithmic}
\STATE Given $\ba$, $B$,   the training set $\{\x_{i}, y_{i}\}_{i=1}^{n}$ and the group index set $\mG = \{\mG_1, ..., \mG_p\}$.\\
1: Calculate ${\c}=\sum_{i=1}^{n}\alpha_{i}y_{i}\x_{i}$.\\
2: Calculate the feature score $\s$, where $s_{j} = \c_{\mG_j}'\c_{\mG_j}$.\\
3: Find the $B$ largest $s_{j}$'s. \\4:  {Set $d_{j}$
corresponding to the $B$ largest $s_{j}$'s to 1 and the \aje{rest} to 0}. \\
5:  Return $\d$.
\end{algorithmic}
\end{algorithm}

\subsection{Worst-Case Analysis for Groups with Complex
Structures}\label{sec:str} Algorithm~\ref{Alg:most_d} can be easily
extended to group features with complex structures, such as groups
with overlapping features or groups with tree-structures. Recall
that $p=|\mG|$\aje{. The} worst-case analysis takes $O(mn+p\log(B))$
complexity where the term $O(mn)$ is \aje{in reference to} the complexity for
computing $\c$, and  the term  $O(p\log(B))$ is w.r.t. the
complexity of sorting $s_{j}$. In general, the second term is
negligible if $p = O(m)$. However, if $p$ is extremely large, the
computational cost \aje{of} computing and sorting $s_{j}$ will be
unbearable. For instance, if the feature groups are organized in
graph or tree structures, $p$ can be so large that $p\gg
m$~\citep{Jenatton2011}.

Notice that we just need to find the $B$ groups with the largest
$s_{j}$'s, \aje{and} thus we can address the above difficulty by implementing
Algorithm~\ref{Alg:most_d} in an incremental way. Specifically, we
calculate the feature score $s_{j}$ for each group one by one and
maintain a cache $\c_B$ to store the \aje{indexes}  and scores of the $B$
feature groups of the largest scores among {those groups} that \aje{have}
been traversed. After computing $s_{j}$ for a new group ${\mG_j}$,
we update $\c_B$ if $s_{j} > s_B^{min}$, where $s_B^{min}$ denotes
the smallest score in $\c_B$.

Using the above techniques, if the groups follow the tree-structures
defined in Definition~\ref{defn:tree}, the \aje{entire} computational cost
of the worst-case analysis can be significantly reduced to
$O(n\log(m)+B\log(p))$.

\begin{remark}\label{lemma:group}
Given a set of groups $\mG = \{\mG_1, . . . , \mG_p\}$ that is
organized as a tree structure in Definition~\ref{defn:tree}, suppose
$ \mG_h \subseteq \mG_g $, then $s_h < s_g$. Furthermore, $\mG_g$
and all its decedent $\mG_h \subseteq \mG_g$ will not be selected if
$s_g < s_B^{min}$. Therefore, the computational cost of the
worst-case analysis can be reduced to $O(n\log(m)+B\log(p))$ for a
balanced tree structure.
\end{remark}

\section{Efficient Subproblem Optimization}\label{sec:learn}
After  updating $\mathcal{C}_t$ via the worst-case analysis, we are
ready to solve the subproblem (\ref{eq:QCQP_sub}). Recall that any
$\d_h \in \mathcal{C}_t$ indexes a set of features. For convenience,
we define ${\bX}_{h} \triangleq [{\x^1_{h}}, ..., {\x^n_{h}}]' \in
\R^{n\times B}$, where $\x^i_{h}$ denotes the $i$th instance with
the features indexed by $\d_h$.

\subsection{Subproblem Optimization via MKL}\label{sec:dual}
Regarding (\ref{eq:QCQP_sub}), let $\mu_{h}\geq 0$ be the dual
variable for each constraint defined by $\d_h$\aje{. Then, }the Lagrangian
function can be written as \aje{follows}:
\begin{eqnarray}\label{eq:lang_dual}
\mathcal{L}(\theta,\ba,\textbf{$\bmu$})={\theta} - \sum_{ \d_{h}\in
\mathcal{C}_t}\mu_{t}\left(\theta-f(\ba, \d_{h})\right). \nonumber
\end{eqnarray}
By setting its derivative w.r.t. $\theta$ to zero,  we have
$\sum\mu_{t}=1$. Let $\bmu$ be the vector of $\mu_{t}$'s, and
$\mathcal{M}=\{\bmu|\sum\mu_{h}={1}, \mu_{h}\geq0\}$ be its domain.
By applying the minimax saddle-point theorem~\citep{Sion1958},
$\mathcal{L}(\theta,\ba,\textbf{$\bmu$})$ can be rewritten as \aje{follows}:
\begin{small}
\begin{eqnarray}\label{eq:fgssvm_MKL}
\max\limits_{\ba\in\mathcal{A}}\min\limits_{\bmu\in \mathcal{M}}~-
\sum\limits_{\d_{h}\in \mathcal{C}_t}\mu_h f(\ba, \d_{h})
=\min\limits_{\bmu\in \mathcal{M}}\max\limits_{\ba\in
 \mathcal{A}}~-\frac{1}{2}(\ba \odot \y)'\! \Big( \sum\limits_{\d_{h}\in
\mathcal{C}_t}\mu_{h} {}
 {{\bX}_{h}{\bX}_{h}^{'}}+\frac{1}{C}\bI\Big)\!(\ba \odot \y),
\end{eqnarray}
\end{small}

 \noindent where the equality holds since the
objective function is concave in $\ba$ and convex in $\bmu$. Problem
(\ref{eq:fgssvm_MKL}) is a multiple kernel learning (MKL)
problem~\citep{lanckriet2004lkm,Rakotomamonjy2008} with
$|\mathcal{C}_t|$ base kernel matrices ${\bX}_{h}{\bX}_{h}'$.
Several existing {MKL} approaches can be adopted to solve this
problem, such as the SimpleMKL~\citep{Rakotomamonjy2008}, which
\aje{solves} the non-smooth problem using the \aje{subgradient}
methods~\citep{Rakotomamonjy2008,Nedic2009new}. Unfortunately, it is
expensive to calculate the \aje{subgradient} w.r.t. $\ba$ for large-scale
problems. The minimax subproblem (\ref{eq:fgssvm_MKL}) can be
directly solved by proximal gradient
methods~\citep{Nemirovski2005,Tseng2008} or SQP
methods~\citep{Royset}. However, they are
 expensive when $n$ is very large.

Based on the definition of ${\bX}_{h}$, we have $
 \sum_{\d_{h}\in \mathcal{C}_t}\mu_{h} {}
 {\bX}_{h}{\bX}_{h}^{'} =  \sum_{\d_{h}\in \mathcal{C}_t}\mu_{h} {}
 {{\bX} \diag(\d_h)}  {\bX}^{'} =    {}
 {{\bX} \diag(\sum_{\d_{h}\in \mathcal{C}_t}\mu_{h} \d_h)}
 {\bX}^{'}$ for the linear feature
selection task. After solving (\ref{eq:fgssvm_MKL}), we have \aje{the following:}
\begin{eqnarray}\label{eq:d_optimum_d}
\d^* =
 \sum_{\d_{h}\in \mathcal{C}_t}\mu^*_{h} \d_h,
\end{eqnarray}
where $\bmu^* = [\mu_1^*, ..., \mu_h^*]'$ denotes the optimal
solution to (\ref{eq:fgssvm_MKL}). This relation also holds for the
group feature selection tasks. Since $\sum_{h=1}^{|\mathcal{C}_t|}
\mu^{*}_h = 1$, we have ${\d}^* \in \mathcal{D} =
\big\{\d\big|\sum_{j=1}^{m}d_{j}\leq B, ~d_{j}\in[0, 1], ~j=1,
\cdots, m\big\}$, where the nonzero entries are associated with
selected features/groups. Since each $\d_h$ selects at most $B$
features/groups, after $t$ iterations, the number of selected
features/groups is no more than $t B$.

\begin{remark}
Suppose Algorithm 1 stops after $t$ iterations, \aje{then} the number of
selected features/groups is constrained in $[B, t B]$, namely
$B\leq||{\d}^*||_0 \leq t B$.
\end{remark}

\subsection{Subproblem Optimization  in the Primal} \label{sec:duall21}
Solving the MKL problem (\ref{eq:fgssvm_MKL}) is very expensive for
\emph{Big Data} where $n$ is very large. Recall that, after $t$
iterations, $\mathcal{C}_t$ includes at most $t B$ features where
$tB\ll n$. Based on this observation, in the following, we propose
to solve it in the primal form w.r.t. $\w$ where $||\w||_0 \leq
tB$. Without loss of generality, \aje{we hereafter} assume that $ t=
|\mathcal{C}_t|$ after  $t$th iterations. 

Let ${\bX}_{h} \in \R^{n\times B}$ denote the data subset with
features selected by $\d_h \in \mathcal{C}_t$. Let ${{\ww}}_h \in
\R^B$ denote the weight vector w.r.t. ${{\bX}_{h}}$ and $\ww =
[\ww_1', ..., \ww_t']' \in \R^{tB}$ be a supervector concatenating
all $\ww_h$'s where $tB \ll n$. For convenience, we define the loss
function w.r.t. the squared hinge loss as \aje{follows:}
$$P(\ww,b) = \frac{C}{2}\sum_{i=1}^{n}\xi_i^2$$ where
 ${\xi_i = \max(1-y_{i}(\sum_{h}{}\ww_h'\x_{ih}-b),0)}$ and the loss function w.r.t. the logistic
loss as \aje{follows:}
 $$P(\ww,b) = C\sum_{i=1}^{n} \log(1+\exp(\xi_i)),$$ where $\xi_i = -y_i (\sum_{h=1}^{t}\ww_h'\x_{ih}-b)$. We
have the following conclusion regarding the subproblem
(\ref{eq:fgssvm_MKL}).
\begin{thm}\label{thm:primal_dual}
Let $\x_{ih}$ denote the $i$th instance of ${\bX}_h$\aje{. Then}, the {MKL}
subproblem (\ref{eq:fgssvm_MKL}) can be equivalently addressed by
solving an $\ell_{2,1}^2$-regularized problem \aje{as follows}:
\begin{eqnarray}\label{eq:l21}
\min\limits_{\ww}~~\frac{1}{2}(\sum_{h=1}^t {\|\ww_h\|}{})^2 +
P(\ww,b).
\end{eqnarray}
Furthermore, the dual optimal {solution} $\ba^*$ can be recovered
from the optimal solution $\xi^*$. To be more specific, $\a_i^* =
C\xi_i^*$ holds for the square-hinge loss and $\a_i =
\frac{C\exp(\xi_i^*)}{1+\exp(\xi_i^*)}$  holds  for the logistic
loss.
\end{thm}

The proof can be found in Appendix A.

According to Theorem \ref{thm:primal_dual}, we can solve the {MKL}
subproblem (\ref{eq:fgssvm_MKL}) in its primal form (\ref{eq:l21})
and recover $\ba^*$ to do the worst-case analysis based on $\bxi^*$
for the squared hinge loss and logistic loss, respectively.

Rather than solving (\ref{eq:fgssvm_MKL}), in this paper, we solve
its primal form (\ref{eq:l21}) instead. For the convenience of
presentation, we define \aje{the following:}
\begin{eqnarray}\label{eq:Fw}
F(\ww,b) =
 \Omega(\ww)
+ P(\ww,b),
\end{eqnarray}
where  $\Omega(\ww) = \frac{1}{2}(\sum_{h=1}^t {\|\ww_h\|}{})^2$.
Furthermore, let $\nabla P(\v) =
\partial_{\v} P(\v,v_b)$ and $\nabla_{b} P(\v,v_b) =
\partial_{b} P(\v,v_b).$ $F(\ww,b)$ is a non-smooth function
w.r.t $\ww$. In addition, $P(\ww,b)$ has \aje{a} block coordinate Lipschitz
gradient w.r.t $\ww$ and $b$ where $\ww$ is deemed \aje{to be} a block
variable. \aje{This is because} it is separable w.r.t $\ww$ and $b$. In addition, it
is known that $F(\ww,b)$ is at least Lipschitz continuous for both
the logistic loss and the squared hinge loss~\citep{Yuan2011}. Then\aje{,}
the following relation holds:
\begin{eqnarray}
P(\ww,b) \leq P(\v,v_b)+\langle \nabla P(\v),\ww-\v \rangle +
\langle \nabla_{b} P(v_b),b-v_b \rangle + \frac{L}{2}\|\ww-\v||^2 +
\frac{L_b}{2}\|b-v_b||^2, \nonumber
\end{eqnarray}
where $L$ and $L_b$ denote the Lipschitz constants  \aje{with respect to}  $\ww$ and $b$, respectively. Therefore, we minimize $F(\ww,b)$ \aje{with respect to} $\ww$ and $b$ in a block coordinate descent
manner~\citep{tseng2001convergence}. In this paper, we propose to
minimize $F(\ww,b)$ \aje{using} an accelerated proximal gradient ({APG})
method~\citep{beck2009fast, Toh2009}. \aje{APG} iteratively
minimizes a quadratic approximation to $F(\ww, b)$ w.r.t. $\ww$.
Specifically, given a point $[\v',v_b]'$, the quadratic
approximation to $F(\ww, b)$ at this point w.r.t. $\ww$ is written
as \aje{follows}:
\begin{eqnarray}\label{eqn:prox}
Q_{\tau}(\ww,\v,v_b) &=&P(\v,v_b)+\langle \nabla P(\v),\ww-\v
\rangle + \Omega(\ww) +
\frac{\tau}{2}\|\ww-\v||^2\nonumber\\
&=& \frac{\tau}{2} \|\ww-\u\|^2 +  \Omega(\ww) + P(\v,v_b)  -
\frac{1}{2\tau}\|\nabla P(\v)\|^2,
\end{eqnarray}
where $\tau$ is a positive constant and $\u = \v - \frac{1}{\tau}
\nabla P(\v)$.  To minimize $Q_{\tau}(\ww,\v,v_b)$ w.r.t. $\ww$, we
need to solve the following Moreau Projection problem:
\begin{eqnarray}\label{eqn:prox-close}
 \min\limits_{\ww}~~ \frac{\tau}{2} \|\ww-\u\|^2 +
\Omega(\ww).
\end{eqnarray}
According to \citep{Martins2010}, this problem has a unique
closed-form solution, which can be calculated \aje{using} Algorithm~\ref{alg:SG}. For convenience, let $\u_h$ be the \aje{component corresponding} to $\ww_h$, namely\aje{,} $\u = [\u_1', ..., \u_t']'$\aje{; then,} we have the following proposition.
\begin{prop}\label{prop:moro}
Let $S_{\tau}(\u,\v)$ be the optimal solution to problem
(\ref{eqn:prox-close}) at point $\v$, then $S_{\tau}(\u,\v)$ is
unique and can be calculated as follows:
\begin{equation}\label{eq:proj}
  [S_{\tau}(\u,\v)]_h =  \left\{
   \begin{array}{cc}
  \frac{{o}_h}{\|\u_h\|}{\u_h}, & \textrm{if}~~ {o}_h>0, \\
   \0, & {\textrm{otherwise}},  \\
   \end{array}
  \right.
  \end{equation}
where $[S_{\tau}(\u,\v)]_h \in \R^{B}$ \aje{denotes} the corresponding
component w.r.t. $\u_h$ and $\bo \in \R^t$ \aje{is} an intermediate
variable. Let $\widehat{\bo} = [{\|\u_1\|} , ..., {\|\u_t\|}]' \in
\R^t$, the intermediate vector $\bo$ can be calculated via a
soft-threshold operator: $
 o_h = [{\textrm{soft}}({\widehat{\bo}},\varsigma)]_h =  \left\{
   \begin{array}{ll}
  \widehat{o}_h-\varsigma, & \textrm{if}~~~~ \widehat{o}_h>\varsigma, \\
   0, & {\textrm{Otherwise}}.  \\
   \end{array}
  \right.
$. Here\aje{,} the threshold value $\varsigma$ can be calculated in Step 4
of Algorithm~\ref{alg:SG}.
\end{prop}
\begin{proof}
The proof can be adapted from the results in Appendix F of
\citep{Martins2010}.
\end{proof}
\begin{algorithm}[H]
\begin{algorithmic}\label{alg:Moreau}
\STATE Given \aje{a} point $\v$,  $s = \frac{1}{\tau}$ and the number of kernels $t$.\\
 1: Calculate $\widehat{o}_h = {\|\bg_h\|} {}$
   for all $h = 1, ..., t$.\\
 2: Sort $\widehat{\bo}$ to obtain $\bar{\bo}$ such that $\bar{o}_{(1)} \geq ... \geq \bar{o}_{(t)}$.\\
 3: Find $\rho = \max\Big\{t\Big|\bar{o}_h - \frac{s}{1+h s}\sum\limits_{i=1}^{h} \bar{o}_i>0, h = 1, ..., t\Big\}$. \\
 4: Calculate the threshold value $\varsigma = \frac{s}{1+\rho s}\sum\limits_{i=1}^{\rho}
 \bar{o}_i$. \\
 5: Compute $\bo = {soft}(\widehat{\bo},\varsigma)$. \\
 6: Compute and output $S_{\tau}(\u,\v)$ via equation (\ref{eq:proj}).
\end{algorithmic}\caption{Moreau Projection $S_{\tau}(\u,\v)$.} \label{alg:SG}
\end{algorithm}

\begin{remark}
For the Moreau Projection in Algorithm~\ref{alg:SG}, a sorting with
$O(t)$ is needed. In our problem setting, in general, $t$ is not
very large. Accordingly, the Moreau Projection can be efficiently
calculated.
\end{remark}

Now we tend to minimize $F(\ww,b)$ \aje{with respect to} $b$. Since there is no
regularizer on $b$, \aje{this} is equivalent to \aje{minimizing} $P(\ww,v_b)$ w.r.t.
$b$. The updating can be done by $b = v_b - \frac{1}{\tau_b}
\nabla_{b} P(\v,v_b)$,
 which is essentially \aje{a} steepest
descent update. After that, we can use the Armijo line
search~\citep{Nocedal2006NO} to find a step size $\frac{1}{\tau_b}$
such that,
\begin{eqnarray}\label{eq:line_search_b}
P(\ww,b) \leq P(\ww,v_b) - \frac{1}{2\tau_b} |\nabla_{b}
P(\v,v_b)|^2,
\end{eqnarray} where $\ww$ is the minimizer to
$Q_{\tau}(\ww,\v,v_b)$. Since the above line search is performed on
a single variable only, it can be efficiently performed.
 With the calculation of $S_{\tau}(\bg)$ in
Algorithm~\ref{alg:SG} and the updating rule of $b$ above, we
\aje{propose} to solve (\ref{eq:l21}) through a modified {APG} scheme
using a block coordinate scheme. The details of the modified {APG}
algorithm \aje{are} presented in Algorithm~\ref{alg:apg}.

In Algorithm~\ref{alg:apg}, $L_t$ and $L_{bt}$ denote the Lipschitz
constants of  $P(\ww,b)$ w.r.t. $\ww$ and $b$\aje{, respectively,} at the $t$\aje{th} iteration of Algorithm~\ref{Alg:cut_algo}. In practice, we
estimate $L_0$ by $L_0 = 0.01 n C$, which will be further adjusted
by the line search. When $t>0$, $L_t$ is estimated by $L_t = \eta
{L_{t-1}}$.  The warm-start for \aje{the} initialization of $\ww$ and $b$ in
Algorithm~\ref{alg:apg} can greatly accelerate the convergence
speed. The internal variables $L^{k}$ and $L_b^{k}$ will be useful
in the proof of the convergence rate. Specifically, at the $t$th
iteration of Algorithm~\ref{Alg:cut_algo}, a sublinear convergence
rate of Algorithm~\ref{alg:apg} is guaranteed.\footnote{Regarding
Algorithm~\ref{alg:apg},  a linear convergence rate can be obtained
for the logistic loss under mild conditions. The details can be
found in Appendix C. }

\begin{thm}\label{thm:apg_con}
Let $L_t$ and $L_{bt}$ be the Lipschitz constant of {$P(\ww,b)$}
w.r.t. $\ww$ and $b$ respectively. Let $\{({\ww^{k}}',b^k)'\}$ be
the sequences generated by Algorithm~\ref{alg:apg} and $L =
\max(L_{bt}, L_t)$\aje{. For} any $k\geq 1$, we have \aje{the following}:
\begin{small}
\begin{eqnarray*}\label{eq:conver}
F(\ww^k,b^k) - F(\ww^*,b^*) \leq \frac{2
L_t||\ww^0-\ww^*||^2}{\eta(k+1)^2} + \frac{2
L_{bt}(b^0-b^*)^2}{\eta(k+1)^2} \leq \frac{2
L||\ww^0-\ww^*||^2}{\eta(k+1)^2} + \frac{2
L(b^0-b^*)^2}{\eta(k+1)^2}.
\end{eqnarray*}
\end{small}
\end{thm}
The proof can be found in Appendix B.

According to Theorem \ref{thm:apg_con}, if $L_{bt}$ is very
different from $L_t$, the block coordinate updating scheme in
Algorithm~\ref{alg:apg} can achieve an improved convergence speed
over batch updating w.r.t. $(\ww',b)'$.

\begin{algorithm}[H]
\begin{algorithmic}
\STATE Initialization: Initialize the Lipschitz constant $L_t
=L_{t-1}$,
set $\ww^0 = \v^1 = [\ww_{t-1}', \0']'$  and $b^{0} = v_b^1 = b_{t-1}$ by warm start, $\tau_0 = L_t$, $\eta\in (0,1)$, parameter $\varrho^{1} = 1$ and $k=1$.\\
1: Set $\tau = \eta \tau_{k-1}$.\\
  ~~~~~~For $j=0, 1, ...,$ \\
  ~~~~~~~~~~Set $\u = \v^k - \frac{1}{\tau}\nabla p(\v^k)$, compute $S_{\tau}(\u,\v^k)$.  \\
  ~~~~~~~~~~If $F(S_{\tau}(\u,\v^k), v_b^k) \leq Q(S_{\tau}(\u,\v^k),\v^k,v_b^k )$, \\
  ~~~~~~~~~~~~~~~Set $\tau_k = {\tau}$, stop;\\
  ~~~~~~~~~~Else\\
  ~~~~~~~~~~~~~~~$\tau = \min\{\eta^{-1}\tau,L_t\}$.\\
  ~~~~~~~~~~End\\
 ~~~~~~End\\
2: Set $\ww^{k} = S_{\tau_k}(\u,\v^k)$ and $L^{k} = \tau_k$. \\
3: Set $\tau_b = \eta \tau_{k}$.\\
  ~~~~~~For $j=0, 1, ...$ \\
  ~~~~~~~~~~Set $b = v_b^k  - \frac{1}{\tau_b} \nabla_{b}
P(\v,v_b^k)$.  \\
  ~~~~~~~~~~If $P(\ww^{k},b) \leq P(\ww^{k},v_b^k) - \frac{1}{2\tau_b} |\nabla_{b} P(\v,v_b^k )|^2$, \\
  ~~~~~~~~~~~~~~~Stop;\\
  ~~~~~~~~~~Else\\
  ~~~~~~~~~~~~~~~$\tau_b = \min\{\eta^{-1}\tau_b,L_t\}$.\\
  ~~~~~~~~~~End\\
 ~~~~~~End\\
4: Set $b^{k} = b$ and $L_{b}^{k} = \tau_b$. Go to Step 8 if the stopping condition \aje{is achieved}. \\

5: Set $\varrho^{k+1}=\frac{1+\sqrt{(1+4(\varrho^k)^2)}}{2}$.

6: Set $\v^{k+1} = \ww^k+\frac{\varrho^{k}-1}{\varrho^{k+1}}(\ww^k-\ww^{k-1})$ and $v_b^{k+1} = b^k+\frac{\varrho^{k}-1}{\varrho^{k+1}}(b^k-b^{k-1})$.\\

7: Let $k=k+1$ and go to step 1.\\
8:  Return and output $\ww_t = \ww^{k}$,  $b_t = b^{k}$ and $L_t =
\eta {\tau_k}$.
\end{algorithmic}\caption{Accelerated Proximal Gradient for Solving Problem (\ref{eq:l21}) (\textbf{Inner Iterations}).} \label{alg:apg}
\end{algorithm}

\textbf{{Warm Start}}: From Theorem \ref{thm:apg_con}, in general,
the number of iterations needed in {APG} to achieve an
$\epsilon$-solution is $O(\frac{||\ww^0-\ww^*||}{\sqrt{\epsilon}})$.
Since {FGM} incrementally includes a set of features into the
subproblems, \aje{a} warm start of $\ww^0$ can be very useful \aje{in improving} its efficiency. To be more specific, when a new active constraint is
added into the constraint set we can use the optimal solution of
the last iteration (denoted by $[{\ww_1^{*}}', ...,
{\ww_{t-1}^{*}}']$) as an initial guess \aje{for} the next cutting plane
iteration. Specifically, at the $t$th iteration we use $\ww^{-1} =
\ww^0 = [{\ww_1^{*}}', ..., {\ww_{t-1}^{*}}', \0']'$ as the starting
point, which can greatly accelerate the convergence speed.

\subsection{De-biasing of FGM}
Based on Algorithm \ref{alg:apg}, we show that {FGM} resembles the
\emph{\aje{retraining}} process and can achieve de-biased solutions.
First, we revisit the de-biasing process for $\ell_1$-minimization.

\textbf{De-biasing for $\ell_1$-methods}.  To reduce the solution
bias, a de-biasing process  is often adopted in $\ell_1$-methods.
For example, in the sparse recovery
problem~\citep{figueiredo2007gradient}, after solving the
$\ell_1$-regularized problem, a least-square problem (\aje{that} drops
the $\ell_1$-regularizer) is solved with the detected supports (or
features). One can also use the de-biasing to reduce the solution
bias of $\ell_1$-SVM. It is worth mentioning that,  when dealing
with classification tasks, due to the rounding errors of the labels,
a suitable regularizer is important to avoid \aje{overfitting}.
Alternatively, we can use the standard SVM  to do the de-biasing
with a \aje{relatively} large $C$ with the features, which is called \emph{\aje{retraining}}. Interestingly, when $C$ goes to infinity, it is
equivalent to \aje{minimizing} the empirical loss without any regularizer,
which, however, may cause \aje{an overfitting} problem.

\textbf{De-biasing effect of FGM}.  In the worst-case analysis,
{FGM} includes $B$ elements (features/groups) that violate the
optimality condition the most. With a \aje{sufficiently} small $B$, these
features/groups will be the most informative features. After that,
FGM solves an $\ell_{2,1}^2$-regularized problem~(\ref{eq:l21})
through Algorithm~\ref{alg:apg} with these features.  Recall that the parameters $B$ and $C$ in FGM are adjusted separately. When
doing the minimization on the $\ell_{2,1}^2$-regularized problem, we
can use a relatively large $C$ to \aje{more strongly} penalize the empirical loss.
Accordingly, with a suitable $C$, each outer iteration of {FGM} can
be deemed as \aje{a} de-biasing process. Moreover, since the solution is
de-biased, it will\aje{, in turn,} benefit the \emph{inference} step in the
worst-case analysis to select \emph{better} features, \aje{and} vice versus.

\subsection{Stopping Conditions}\label{sec:stop}
Suitable stopping conditions of {FGM} are important to reduce the
risk of \aje{overfitting} and improve the training efficiency. The
stopping criteria of {FGM} include: 1) the stopping conditions for
the outer cutting plane iterations in Algorithm~\ref{Alg:cut_algo} \aje{and}
2) the stopping conditions for the inner APG iterations in Algorithm~\ref{alg:apg}.

\subsubsection{Stopping Conditions for Outer iterations}\label{sec:stop_out}
We first introduce the stopping conditions w.r.t. the outer cutting
plane iterations, e.g., Algorithm~1. Recall that the optimality
condition for the SIP problem is $\sum_{\d_{t}\in \mathcal{D}}\mu_t
\nabla_{\ba}f(\ba, \d_{t})
 = \0$ and $\mu_t(f(\ba, \d_{t}) - f_m(\ba)) = 0, \forall \d_t \in \mathcal{D}$.
A direct stopping condition can be written as \aje{follows}:
\begin{eqnarray}\label{eq:stop_original}
f(\ba, \d)  \leq {f}_\text{m}(\ba) + \epsilon, ~~~\forall \d \in
\mathcal{D},
\end{eqnarray}
where ${f}_\text{m}(\ba) = \max_{\d_{h} \in \mathcal{C}_t}f(\ba,
\d_{h})$\aje{,} and $\epsilon$ is a small tolerance value. To check this
condition, we \aje{simply} need to find a new $\d_{t+1}$ via the worst-case
analysis. If $f(\ba, \d_{t+1}) \leq f_m(\ba) + \epsilon$, the
stopping condition in (\ref{eq:stop_original}) is achieved. In
practice, due to the scale variation of ${f}_\text{m}(\ba)$ for
different problems, it is \aje{nontrivial} to set the tolerance
$\epsilon$. Notice that we perform the subproblem optimization in
the primal, and the objective value $F(\ww_t)$ monotonically
decreases. Therefore, in this paper, we propose to use the \aje{following} relative function value difference as the stopping condition instead:
\begin{eqnarray}\label{eq:stop_con}\label{eq:stop_out}
\frac{F(\ww_{t-1},b) - F(\ww_{t},b)}{F(\ww_{0},b)} \leq
\epsilon_{c},
\end{eqnarray}
 where $\epsilon_{c}$ is a small tolerance value.

In some applications, one may need to select a desired number of
features. In such cases, we can terminate Algorithm~1 after a
maximum number of $T$ iterations \aje{when} at most $TB$ features \aje{are}
selected.

\subsubsection{Stopping Conditions for Inner
iterations}\label{sec:stop_in}
 \textbf{Exact
and Inexact {FGM}}: In each iteration of Algorithm~1, one needs to
do the inner master  problem minimization in (\ref{eq:l21}). Its
optimal condition is $\nabla_{\ww}F(\ww) = \0$. In practice, to
achieve a solution with high precision to meet this condition is
expensive yet unnecessary. Alternatively, one  achieves an
$\epsilon$-accurate solution.

However, \aje{an} inaccurate solution may affect the convergence of
{FGM}. Let $\widehat{\ww}$ and $\widehat{\bxi}$ be the exact
solution to (\ref{eq:l21}). According to Theorem~\ref{thm:primal_dual}, the exact solution of $\widehat{\ba}$ to
(\ref{eq:fgssvm_MKL}) can be obtained by \aje{setting} $\widehat{\ba} =
\widehat{\bxi}$.  Now suppose ${\ww}$ is an $\epsilon$-accurate
solution to (\ref{eq:l21})\aje{, and let} $\bxi$ be the corresponding loss\aje{; then, }we have $\alpha_i = \widehat{\alpha}_i + \epsilon_i$, where
$\epsilon_i$ is \aje{the} gap between $\widehat{\ba}$ and ${\ba}$. When
performing the worst-case analysis in Algorithm~\ref{Alg:most_d}, we
need to calculate \aje{the following:}
\begin{eqnarray}{\c}=\sum_{i=1}^{n}\alpha_{i}y_{i}\x_{i}
=\sum_{i=1}^{n}(\widehat{\alpha}_{i} + \epsilon_i) y_{i}\x_{i}
=\widehat{\c} + \sum_{i=1}^{n} \epsilon_i y_{i}\x_{i} = \widehat{\c}
+\Delta \widehat{\c},
\end{eqnarray}
where $\widehat{\c}$ denotes the exact feature score w.r.t.
$\widehat{\ba}$, and $\Delta \widehat{\c}$ denotes the error \aje{in}
${\c}$ \aje{caused} by the inexact solution. Apparently, we have \aje{the following:}
\begin{eqnarray}
|\widehat{c}_j-{c}_j| = |\Delta \widehat{c}_j| = O(\epsilon),
~\forall j=1, ..., m.
\end{eqnarray}
Notice that we just need to find significant features with the
largest $|c_j|$. Accordingly, an $\epsilon$-accurate solution with a
sufficiently small $\epsilon$ can still choose the significant
features. Therefore, the convergence of {FGM} will not be affected
with a  sufficiently small $\epsilon$. Let $\{\ww^k\}$ be the inner
iteration sequence, in this paper, we set the stopping condition of
the inner problem as \aje{follows:}
\begin{eqnarray}\label{eq:stop_inner}
\frac{F(\ww^{k-1}) - F(\ww^k)}{F(\ww^{k-1})}\leq \epsilon_{in},
\end{eqnarray}
where $\epsilon_{in}$ is a small tolerance value. In practice, we
usually set $\epsilon_{in} = 0.001$.

\subsection{Cache for Efficient Implementations}\label{sec:Implement}

The optimization scheme of FGM allows \aje{the use of} some cache techniques
to improve the optimization efficiency.

\textbf{Cache for features}. Different from the cache used in kernel
{SVM}~\citep{ref05a} \aje{that} caches kernel entries, we cache the
features in {FGM}. In gradient-based methods, one needs to calculate
$\w'\x_i$ for each instance to compute the gradient of the loss
function, which takes $O(mn)$ complexity in general on all the input
features. Unlike these methods, the gradient computations in the
modified {APG} algorithm of FGM are w.r.t. the selected features
only. Therefore, we can cache these features to accelerate the
feature retrieval. To cache these features, we \aje{need} $O(tBn)$
additional memory where $tB \ll m$ for \aje{high-dimensional} problems.
However, the operation complexity for feature retrieval can be
significantly reduced from $O(nm)$ to $O(tBn)$. The cache of
features is particularly important for nonlinear feature selection
with explicit feature mappings, where the data with expanded
features can be too large to be loaded into the main memory.

\textbf{Cache for inner products}. The cache technique can \aje{also be}
used to accelerate the Algorithm~\ref{alg:apg}.  To make a
sufficient decrease \aje{in} the objective value, in Algorithm~\ref{alg:apg} a line search is performed to find a suitable step
size. When doing the line search, one may need to calculate the loss
function $P(\ww)$ many times, where $\ww = S_{\tau}(\bg) = [\ww_1',
..., \ww_t']'$.  The computational cost will be very high if $n$ is
very large. However, according to equation (\ref{eq:proj}), we have \aje{the following:}
$$\ww = S_{\tau}(\bg_h)= \frac{{o}_h}{||\bg_h||} \bg_h = \frac{{o}_h}{||\bg_h||} ( \v_h - \frac{1}{\tau} \nabla
P(\v_h)),$$ where only ${o}_h$ is affected by the step size. Then\aje{,}
the calculation of $\sum_{i=1}^{n} \ww'\x_i$ follows:
\begin{eqnarray}
&&\sum\limits_{i=1}^{n} \ww'\x_i = \sum\limits_{i=1}^{n}
\left(\sum\limits_{h=1}^{t}\ww_h'\x_{ih}\right) =
\sum\limits_{i=1}^{n} \left(\sum\limits_{h=1}^{t}
\frac{{o}_h}{||\bg_h||}\left( \boxed{\v_h'\x_{ih}} - \frac{1}{\tau}
\boxed{\nabla P(\v_h)'\x_{ih}}\right)\right). \nonumber
\end{eqnarray}
According to the above calculation rule,  we can make a fast
computation of $\sum_{i=1}^{n}\ww'\x_i$ by caching $\v_h'\x_{ih}$
and $\nabla P(\v_h)'\x_{ih}$ for the $h^{th}$ group  of each
instance $\x_i$. Accordingly, the complexity of computing
$\sum_{i=1}^{n} \ww'\x_i$ can be reduced from $O(ntB)$ to $O(nt)$.
\aje{In other words}, no matter how many line search steps \aje{need} to be
conducted, we only need to scan the selected features once.
Therefore, the computational cost can be much reduced.

\section{Nonlinear Feature Selection Through Kernels}\label{sec:kernel}
Using the kernel tricks, we can extend {FGM} to do nonlinear feature
\aje{selection}. Let $\bphi(\x)$ be a nonlinear feature mapping that maps
the \emph{nonlinear} input  features to a high-dimensional linear
feature space.   To select the features, again we introduce a
scaling vector $\d \in \mathcal{D}$ to the input features resulting
in a new feature mapping $\bphi(\x\odot \sqrt \d)$. By replacing
$(\x\odot \sqrt \d)$ in (\ref{eq:fgssvm}) with $\bphi(\x \odot \sqrt
\d)$, we obtain the kernel version of {FGM}, which can be formulated
as the following semi-infinite kernel (SIK) learning problem:

\begin{eqnarray}\label{eq:kernelFGM}
\min\limits_{\ba \in \mathcal{A},\theta} \theta &:& \theta\geq
f_{\K}(\ba, \d),~~\forall~\d\in
 \mathcal{D},
\end{eqnarray}
where $f_{\K}(\ba, \d) = \frac{1}{2}(\ba \odot \y)'\!
(\K_{\d}+\frac{1}{C}I)\!(\ba \odot \y)$ and ${\K}_{\d}^{ij}$ is
calculated as $\bphi(\x_i \odot \sqrt{\d})'\bphi(\x_j \odot
\sqrt{\d})$. This problem can be  solved by
Algorithm~\ref{Alg:cut_algo}. However, for kernelization,  we need
to solve the following optimization problem for the worst-case
analysis:
\begin{small}
\begin{eqnarray}\label{eq:kernel_max_d}
\!\!&\max\limits_{\d\in\mathcal{D}}&\frac{1}{2}\Big\|\sum\limits_{i=1}^{n}\ba_{i}y_{i}{\bphi}(\x_{i}\odot
 \sqrt{\d})\Big\|^{2} =  \max\limits_{\d\in\mathcal{D}}~ \frac{1}{2}(\ba \odot \y)'
{\K}_\d (\ba \odot \y),
\end{eqnarray}
\end{small}

\subsection{\aje{Worst-Case} Analysis for Additive Kernels}
\noindent In general, solving problem~(\ref{eq:kernel_max_d}) for
general kernels (\emph{e.g.}\aje{, a} Gaussian kernel) is very challenging.
However, this problem can be exactly solved for \textbf{additive
kernels}. A kernel $\K_{\d}$ is an additive kernel if it can be
linearly represented by a set of  base kernels $\{\K_j\}_{j=1}^{p}$,
where each  base kernel $\K_j$ is constructed by one feature or a
subset of features~\citep{Maji2009}. In other words, we can select
the optimal subset features by choosing a small subset of important
kernels.

\begin{prop}
The worst-case analysis  w.r.t. additive kernels can be exactly
solved.
\end{prop}
\begin{proof}
Notice that each base kernel $\K_j$ in an additive kernel is
constructed by one feature or a subset of features. Let $\mG =
\{\mG_1, ..., \mG_p\}$ be the index set of features that produce the
base kernel set $\{\K_j\}_{j=1}^{p}$ and ${\bphi}_j(\x_{i\mG_j})$ be
the corresponding feature map to $\mG_j$. Accordingly, for additive
kernels, the kernel selection problem can be considered as a group
feature selection problem. Similar \aje{to} the group feature selection,
we introduce a feature scaling vector $\d \in \mathcal{D} \subset
\R^p$ to scale ${\bphi}_j(\x_{i\mG_j})$. The resultant model
becomes \aje{as follows}:
\begin{eqnarray}\label{eq:additive_group}
\min\limits_{\d\in \widehat{\D}}\min\limits_{\tw, \bxi, b}&&\frac{1}{2}\|\w\|_{2}^{2}+\frac{C}{2}\sum\limits_{i=1}^{n}\xi^{2}_{i} \\
\text{s.t.} &{}& y_{i} \left(\sum\limits_{j=1}^{p}
\sqrt{{d}_j}\w_{\mG_j}'{\bphi}_j(\x_{i\mG_j}) -b \right) \geq
1-\xi_{i}, \;\;\xi_i\geq 0, \;\;i=1, \cdots, n, \nonumber
\end{eqnarray}
where $\w_{\mG_j}$ has the same dimensionality \aje{as}
${\bphi}_j(\x_{i\mG_j})$. By transforming this problem \aje{into an} SIP
problem in (\ref{eq:group_unify}), we can solve \aje{the} kernel learning
(selection) problem via FGM. The corresponding worst-case analysis
is  to solve the following problem:
\begin{eqnarray}\label{eq:max_HIK}
\max\limits_{\d\in\mathcal{D}}~~\sum_{j=1}^p d_j(\ba\odot\y)' \K_j
(\ba\odot\y)  = \max\limits_{\d\in\mathcal{D}}~
\sum\limits_{j=1}^{p}d_j s_j,
\end{eqnarray}
where $s_j =
 (\ba\odot\y)' \K_j
(\ba\odot\y)$ and $\K_j^{i,k} = {\bphi}_j(\x_{i\mG_j})'
{\bphi}_j(\x_{k\mG_j})$. This problem can be exactly solved by
choosing the $B$ kernels with the largest $s_j$'s.
\end{proof}
In the past decades, many additive kernels have been proposed based
on different application contexts, such as the general intersection
kernel in computer vision~\citep{Maji2009}, \aje{the} string kernel in text
mining and ANOVA kernels~\citep{Francis2009HKL}. Taking the general
intersection kernel \aje{as an} example, it is defined as $ k(\x,\z,a) =
\sum_{j=1}^{p} \min\{|x_j|^a,|z_j|^a\},$ where $a>0$ is a kernel
parameter. When $a=1$, it reduces to the well-known Histogram
Intersection Kernel (HIK), which has been widely used in computer
vision and text \aje{classification}~\citep{Maji2009,wu2012efficient}.

It is worth mentioning that, even though we can exactly solve the
worst-case analysis for additive kernels, the subproblem
optimization is still very challenging for large-scale problems.
First, storing the kernel matrices takes $O(n^2)$ space
complexity, which is unbearable when $n$ is very large. More
critically, solving the MKL problem with many training points is
still computationally expensive. To address these issues, we propose
to approximate a base kernel matrix using a group of approximated
features, such as random features~\citep{Vedaldi2010} and HIK expanded features~\citep{wu2012efficient}. As a result, the MKL
problem is reduced to a \textbf{group feature selection} problem\aje{. Thus,} it is scalable to \emph{Big Data} by avoiding \aje{storage of} the
base kernel matrices. Moreover, the subproblem optimization can be
efficiently solved in the primal.

\subsection{Worst-Case Analysis for \aje{Ultrahigh-Dimensional} Big Data}\label{sec:ultra}
\emph{Big Data} with ultrahigh dimensionality \aje{exists widely} in many
applications, such as \aje{in} nonlinear feature selection with explicit
feature mappings.

In general, nonlinear feature selection \aje{with} general kernels is
difficult. However, if the explicit feature mapping of a kernel is
available, the nonlinear feature selection problem can be cast as a
linear feature selection problem in the feature space. {However, the
dimensionality of the feature space may become very high. Taking the
polynomial kernel $k(\x_{i},\x_{j}) = (\gamma
\x_{i}'\x_{j}+r)^{\upsilon}$ for example, the dimension of the
feature mapping  exponentially increases with
$\upsilon$~\citep{Chang2010}, where $\upsilon$ is referred to as the
degree. When $\upsilon=2$, the 2-degree explicit feature mapping can
be expressed as \aje{follows:}
$${\bphi(\x)}= [r, \sqrt{2\gamma r}x_{1}, ..., \sqrt{2\gamma
r}x_{m}, \gamma x_{1}^2, ..., \gamma x_{m}^2, \sqrt{2} \gamma
x_{1}x_{2}, ..., \sqrt{2} \gamma x_{m-1}x_{m}].$$ Compared with the
original features, the second-order feature mapping can capture the
feature pair dependencies. Therefore, it has been widely \aje{used} in
many applications\aje{,} such as text mining and natural language
processing~\citep{Chang2010}. Unfortunately, the dimensionality of
this feature mapping is $(m+2)(m+1)/2$. Apparently, the
dimensionality will be ultrahigh for a \aje{modest} $m$. For example, if
$m = 10^{6}$, the dimensionality of the feature space is
$O(10^{12})$, where \aje{approximately} {1 TB} \aje{of memory will be} required to store the
weight vector $\w$. As a result, the existing methods are not
feasible~\citep{Chang2010}. However, this computational bottleneck
can be effectively addressed by the proposed feature generating
paradigm in which only $tB$ features (or groups) are needed to be
stored in the main memory. As a result, we only store the \aje{indexes}
and scores of the $tB$ features in a cache $\c_{B}$.

\begin{algorithm}[h]
\caption{Incremental Implementation of Algorithm~2 for \aje{Ultrahigh-Dimensional} Data. } \label{Alg:most_d_ulta}
\begin{algorithmic}
\STATE Given $\ba$, $B$, number of data groups $k$,
feature mapping $\bphi(\x_i)$ and a structured array ${\c_{B}}$.\\
1: Split $\bX$ into $k$ subgroups $\bX = [\bX^1, ..., \bX^k]$.\\
2: For $j=1, ..., k$.\\
 ~~~~~~Calculate the feature score $\s$ w.r.t. $\bX^j$ according to  $\bphi(\x_i)$.\\
 ~~~~~~Sort $\s$ and update $\c_{B}$.\\
 ~~~~~~For $i=j+1, ..., k$.  (\textbf{Optional})  \\
 ~~~~~~~~~~~Calculate the cross feature score $\s$ w.r.t. $\bX^j$ and $\bX^i$.  \\
 ~~~~~~~~~~~Sort $\s$ and update $\c_{B}$. \\
 ~~~~~~End\\
 ~~End\\
 3: Return $\c_{B}$.
\end{algorithmic}
\end{algorithm}

\aje{Ultrahigh-dimensional} \emph{Big Data} can be too \aje{large} to be
loaded into the main memory\aje{. Thus,} the worst-case analysis is still
very challenging to \aje{address}. Motivated by the incremental
worst-case analysis for complex group feature selection in Section
\ref{sec:str}, we propose to address the \emph{Big Data} challenge
in an incremental manner. The general scheme for the incremental
implementation is presented in Algorithm~\ref{Alg:most_d_ulta}.
Particularly, we partition $\bX$ into $k$ small data \aje{subsets} of lower
dimensionality \aje{such that} $\bX = [\bX^1, ..., \bX^k]$. For each small data
subset, we can load \aje{the subset} into memory and calculate the feature scores
of the features. In Algorithm~\ref{Alg:most_d_ulta}, the inner loop
w.r.t. the iteration index $i$ is used only for second-order
feature selection where the calculation of \aje{the} feature score for the
\textbf{cross-features} is required. For instance, in nonlinear
feature selection using 2-degree polynomial mapping, we need to
calculate the feature score of $x_{i}x_{j}$.

\section{Connections to Related Studies}\label{sec:related}
In this section, we discuss the connections of \aje{the} proposed methods with
related studies, \aje{such as} the $\ell_1$-regularization~\citep{Jenatton2009}, Active-set methods, SimpleMKL~\citep{Rakotomamonjy2008},
$\ell_q$-MKL, infinite kernel learning (IKL), SMO-MKL~\citep{Vishy10}, \aje{etc}.

\subsection{Relation to $\ell_1$-regularization}
\revise{Recall that the $\ell_1$-norm of a vector $\w$ can be expressed as
the following variational form~\citep{Jenatton2009}:
\begin{eqnarray}\label{eq:ell}
\|\w\|_1 = \sum_{j=1}^{m} |w_j| = \frac{1}{2} \min_{\d\succeq 0}
\sum_{j=1}^{m} \frac{w_j^2}{d_j} + d_j.
\end{eqnarray}
It is not difficult to verify that,  $d_j^* = |w_j|$ holds at the
optimum, which indicates that the scale of $d_j^*$ is proportional
to $|w_j|$.  Therefore, it is useless to impose the additional
$\ell_1$-constraint $||\d||_1 \leq B~\mathrm{or}~||\w||_1\leq B$ in
the $\ell_1$-regularized problem.}

\revise{To address this challenge, in AFS, we bound $\d \in [0,
1]^m$. To demonstrate the effect of this box constraint, we make the
following transformations.  Let $\widehat{w}_j = w_j \sqrt{d_j}$ and
$\widehat{\w} = [\widehat{w}_1, ..., \widehat{w}_m]'$, the
variational form of the problem (\ref{eq:fgssvm}) can be
equivalently written as
\begin{eqnarray}\label{eq:fgssvm2}
\min\limits_{\d\in\D }\min\limits_{\widehat{\tw},
\bxi,b}&{}&\frac{1}{2}\sum_{j=1}^m\frac{\widehat{w}_j^2}{d_j} +
{\frac{C}{2}\sum\limits_{i=1}^{n}\xi^{2}_{i}}
\\ \text{s.t.} &{}&  {y_{i}(\widehat{\tw}'\x_{i} - b) \geq 1 -\xi_{i}}, \;\; \;\;i=1, \cdots, n. \nonumber
\end{eqnarray}
For simplicity, hereby we drop the hat from $\widehat{\tw}$ and
define a new regularizer $\|\w\|_B^2$ as
\begin{eqnarray} \label{eq:Bnorm}
\|\w\|_B^2 = {\min_{\d \succeq \0}\sum_{j=1}^{m} \frac{w_j^2}{d_j}},
~~~\text{s.t.}~~~ ||\d||_1 \leq B, ~~\d \in [0, 1]^m.
\end{eqnarray}
 This new regularizer
has the following properties.
\begin{prop} \label{prop:Bnorm}
{Given a vector $\w \in \mathbb{R}^m$ with $\|\w\|_0 = {\kappa}>0$,
where ${\kappa}$ denotes the number of nonzero entries in $\w$. Let
$\d^*$ be the minimizer of (\ref{eq:Bnorm}), we have: (I)  $d^*_{j}
= 0$ if $|w_j| = 0$. (II) If ${\kappa} \leq B$, then $d^*_j = 1$ for
$|w_j| > 0$; else if $\frac{\|\w\|_1}{\max\{|w_j|\}}\geq {B}$ and
${\kappa} > B$, then we have $\frac{|w_j|}{d_j^*} =
\frac{\|\w\|_1}{B}$ for all $|w_j| >0$. (III) If ${\kappa}\leq B$,
then
 $\|\w\|_B = \|\w\|_2$; else if $\frac{\|\w\|_1}{\max\{|w_i|\}}\geq
{B} $  and ${\kappa} > B$,  $\|\w\|_B = \frac{\|\w\|_1}{\sqrt{B}}$.
}
\end{prop}
The proof can be found in Appendix D.}

Proposition~\ref{prop:Bnorm} can \aje{also be} extended to the group
feature \aje{case}. Specifically, given a $\w \in \mathbb{R}^m$ with $p$
groups $\{ \mG_1, ..., \mG_p \}$, we have $\sum_{j=1}^p
||\w_{\mG_j}||_2 = \|\v\|_1$, where $\v = [||\w_{\mG_1}||, ...,
||\w_{\mG_p}||]'\in \R^p$.} \revise{According to Proposition
\ref{prop:Bnorm}, no matter how large the magnitude of $|w_j|$ is,
$d_j$ in $\|\w\|_B^2$ is always upper bounded by 1.}  \aje{In contrast}, $\d$ in the variation form of the $\ell_1$-norm is not
upper bounded. As a result,  it is meaningless to impose the
additional $\ell_1$-constraint $||\d||_1 \leq B$ in (\ref{eq:ell}) \aje{since} $\d$ is scale-sensitive.

\revise{There are two advantages of using $\|\w\|_B^2$ over the
$\ell_1$-norm regularizer. Firstly, by using $\|\w\|_B^2$, the
sparsity and the over-fitting problem can be controlled separately.
Specifically, we can choose a proper $C$ to reduce the feature
selection bias, and set a proper stopping tolerance $\epsilon_{c}$
in (\ref{eq:stop_out}) or a proper parameter $B$ to adjust the
number of  features to be selected. Conversely, in the $\ell_1$-norm
regularized problems, the number of features is determined by the
parameter $C$, and the solution bias may happen if we intend to
select a small number of features. Secondly, by transforming the
resultant optimization problem into an SIP problem, a feature
generating paradigm has been developed. By iteratively infer the
most informative features, this scheme is particularly suitable for
dealing with \aje{ultrahigh-dimensional} \emph{Big Data} that are
infeasible for the existing $\ell_1$-norm methods, as shown in
Section \ref{sec:ultra}.}

\subsection{Connection to Existing AFS Schemes}
The proposed AFS scheme is very different from the existing AFS
schemes \revise{that have been widely studied
in~\citep{Weston2000,Chapelle2002,Grandvalet2002,Alain2003,Varma09,Vishy10}.
In these studies, a scaling vector $\d\succeq \0$ is introduced,
where $\d$ is not upper bounded. For instance, in \citep{Vishy10},
the AFS problem is reformulated as an SMO-MKL problem:
\begin{eqnarray}\label{eq:AFS_ellp}
\min\limits_{\d\succeq \0}\max\limits_{\ba\in
\mathcal{A}}~\1'\ba-\frac{1}{2} \sum\limits_{j=1}^{p} d_j
(\ba\odot\y)'\K_j(\ba\odot\y) +
\frac{\lambda}{2}(\sum_jd_j^q)^{\frac{2}{q}},
\end{eqnarray}
where $\mathcal{A} = \{\ba|0 \preceq \ba \preceq C\1,\y'\ba = 0\}$
and $\K_j$ denote a sub-kernel. When $0\leq q \leq 1$, it induces
sparse solutions, but results in non-convex optimization problems.
Moreover, the sparsity of the solution is still determined by the
regularization parameter $\lambda$. Consequently, the solution bias
inevitably exists in the SMO-MKL formulation. }

\revise{A more closely related work is the
$\ell_1$-MKL~\citep{bach2004mkl,sonnenburg2006lsm} or the SimpleMKL
problem~\citep{Rakotomamonjy2008}, which tries to learn a linear
combination of kernels. The variational regularizer of SimpleMKL can
be written as:}\revise{
$$ {\min\limits_{\d\succeq0}\sum_{j=1}^{p} \frac{||\w_j||^2}{d_j}},
~~~~\mathrm{s.t.}~~~~ ||\d||_1 \leq 1,$$ where $p$ denotes the
number of kernels and $\w_j$ represents the parameter vector of
$j$th kernel in the context of
MKL~\citep{kloft2009efficient,LpMKL,kloft2012convergence}.
Correspondingly, the regularizer $||\w||_B^2$ regarding kernels can
be expressed as:
\begin{eqnarray}\label{eq:kernel_FGM}
{\min\limits_{\d\succeq0}\sum_{j=1}^{p} \frac{||\w_j||^2}{d_j}},
~~~~\mathrm{s.t.}~~~~ ||\d||_1 \leq B, ~~\d \in [0, 1]^p.
\end{eqnarray}}
\revise{To illustrate the difference between (\ref{eq:kernel_FGM})
and the $\ell_1$-MKL, we divide the two constraints in
(\ref{eq:kernel_FGM}) by $B$, and obtain
$$ \sum_{i=1}^p \frac{d_i}{B} \leq 1, ~~0 \leq \frac{d_i}{B} \leq
\frac{1}{B}, \forall i \in \{1,..., p\}.$$ Clearly, the  box
constraint $\frac{d_i}{B} \leq \frac{1}{B}$ makes
(\ref{eq:kernel_FGM}) different from the variational regularizer in
$\ell_1$-MKL. {Actually, the $\ell_1$-norm MKL is only a special
case of $||\w||_B^2$ when $B=1$. Moreover, by extending Proposition
\ref{prop:Bnorm}, we can obtain that if $B > \kappa$, we have
$||\w||_B^2 = \sum_{i=1}^{p} ||\w_j||^2$, which becomes a non-sparse
regularizer. Another similar work is the $\ell_q$-MKL, which
generalizes the $\ell_1$-MKL from $\ell_1$-norm to $\ell_q$-norm
($q>1$)~\citep{kloft2009efficient,LpMKL,kloft2012convergence}.
Specifically, the variational regularizer of $\ell_q$-MKL can be
written as
$$ {\min_{\d\succeq0}\sum_{j=1}^{p} \frac{||\w_j||^2}{d_j}},
~~~~\mathrm{s.t.}~~~~ ||\d||_q^2 \leq 1.$$ Clearly, the  box
constraint $0 \leq \frac{d_i}{B} \leq \frac{1}{B}, \forall i \in
\{1,..., p\}$ is missing in the $\ell_q$-MKL. Notice that, when
$q>1$, the $\ell_q$-MKL cannot induce sparse solutions, and thus
cannot discard non-important kernels or features. Therefore, the
underlying assumption for $\ell_q$-MKL is that, most of the kernels
are relevant for the classification tasks. Accordingly, the bias
issue is not significant. \aje{Last}, it is worth mentioning that, when
doing multiple kernel learning, both $\ell_1$-MKL and $\ell_q$-MKL
require to compute and involve all the base kernels. Therefore, the
computational cost is unbearable for large-scale problems with many
kernels.}}

\revise{In~\citep{Gehler2008}, an infinite kernel learning method is
introduced to deal with infinite number of kernels ($p=\infty$).
Specifically, IKL adopts the $\ell_1$-MKL
formulation~\citep{Gehler2008},  and thus it is a special case of
FGM with $B=1$. To solve this problem, IKL also adopts the cutting
plane algorithm. However, it can only include one kernel  per
iteration; while FGM includes $B$ kernels per iteration. In this
sense, {IKL is also analogous to the Active-set
methods}~\citep{Roth2008,Francis2009HKL}.   Notice that, for both
methods, the worst-case analysis for large-scale problems usually
dominates the overall training complexity. For FGM, since it is able
to include $B$ kernels per iteration, it obviously reduces the
number of worst-case analysis steps, and thus has great
computational advantages over IKL.  \aje{Last}, it is worth mentioning
that,  based on the cutting plane algorithm, it is \aje{nontrivial} for
IKL to include $B$ kernels. }

\subsection{Connection to  Multiple Kernel Learning}
As previously mentioned, the subproblem (26) can be addressed by the
SimpleMKL~\citep{Rakotomamonjy2008}. In \citep{bach2004mkl,Vishy10},
an approximate solution can be efficiently obtained by a sequential
minimization optimization (SMO). \cite{sonnenburg2006lsm} proposed a
semi-infinite linear programming formulation for MKL \aje{that} allows
{MKL} to be iteratively solved with \aje{an} SVM solver and linear
programming. In addition, \cite{xu-09} proposed an extended level
method to improve the convergence of {MKL}.  More recently, an
online \aje{ultra-fast MKL algorithm called} the {UFO-MKL} is
proposed in~\citep{orabona2011ultra}. However, its $O(1/\epsilon)$
convergence rate is only guaranteed when a strongly convex
regularizer $\Omega(\w)$ is added to the standard lasso problem.
Without the  strongly convex regularizer, the convergence of
{UFO-MKL} is unclear. Therefore, it cannot be used to solve the
subproblem in FGM.

FGM is different from {MKL} in several aspects. At first, {FGM}
iteratively includes $B$ new kernels through the worst-case
analysis. Particularly, these $B$ kernels will be formed as a base
kernel for the MKL subproblem of FGM.  From the kernel learning \aje{point of} view, FGM provides a new way to construct base kernels. \aje{Second},
since FGM tends to select a subset of kernels, it is especially
suitable for MKL with many kernels. \aje{Third}, to scale MKL to
\emph{Big Data}, we propose to use the approximated features (or
explicit feature mappings) for kernels. Accordingly, the MKL problem
is reduced to a group feature selection problem, and we can solve
the subproblem in its primal form. %

\subsection{Connection to Active Set Methods}
Active set methods have been widely used to address the challenges
of \aje{an} extremely large number of features or
kernels~\citep{Roth2008,Francis2009HKL}. Basically, active set
methods iteratively include a violated variable that violates the
optimality condition of sparsity-induced problems. In this
sense, active methods can be considered a special case of FGM when
$B=1$. However, FGM is different from active-set methods in several
\aje{ways}.

\aje{First}, their motivations are different. Specifically, active set
methods start from the Lagrangian duality of sparsity-induced
problems~\citep{Francis2009HKL,Roth2008}\aje{,} while FGM starts from a
novel AFS scheme, which is reduced to solve an SIP problem.
\aje{Second}, active set methods only include one active
feature/kernel/group at each iteration. Regarding this algorithm,
when the desired number of kernels or groups becomes relatively
large, active set methods will be very computationally expensive. \aje{In contrast}, {FGM} allows \aje{us} to add $B$ new features/kernels/groups
per iteration, which can greatly improve the training efficiency by
reducing the number of worst-case \aje{analyses}. \aje{Third}, when \aje{performing} group feature selection tasks, the $B$ new feature groups obtained by the worst-case analysis will form a bigger feature group.
Particularly, FGM solves a sequence of $\ell_{2,1}^2$-regularized
non-smooth problems, which is very different from active set
methods~\citep{Francis2009HKL,Roth2008}. \aje{Last}, the de-biasing of
solutions is not investigated in active set methods.

\section{Experiments} \label{sec:exp}

In this section, we compare {FGM} with several state-of-the-art
baseline methods on several learning tasks, namely\aje{,} linear
feature selection, \aje{ultrahigh-dimensional} nonlinear feature
selection and group feature selection.\footnote{Since our focus
is on large-scale and very high-dimensional problems, some
aforementioned methods, such as {NMMKL},  {QCQP-SSVM} and {SVM-RFE}
are not included due to \aje{their} high computational cost or
\aje{their} sub-optimality for feature selection. Interested readers can refer to~\citep{tan10} for detailed comparisons. \revise{We also do
not include the $\ell_q$-MKL
~\citep{kloft2009efficient,LpMKL,kloft2012convergence} for
comparison since it cannot induce sparse solutions. Instead, we
include an $\ell_q$-variant, i.e., UFO-MKL~\citep{orabona2011ultra},
for comparison. \aje{Last}, since IKL is a special case of FGM with
$B=1$, we study its performance through FGM with $B=1$ instead.
Since it is analogous to the active-set methods, its performance can
be also observed from the results of active-set method.  }}

We organize the experiments as follows.  In Section
\ref{sec:sub_sec1}, we discuss the general experimental settings. In
Section \ref{sec:exp1}, we conduct synthetic experiments for linear
feature selection. In Section \ref{exp:shift}, we present an
experiment to study the effectiveness of the shift version of FGM
\aje{,} which incorporates \aje{a} shift of the hyperplane. In Section
\ref{sec:real-world}, we conduct linear feature selection
experiments on several real-world datasets. In Section
\ref{sec:subexpploy}, we detail the experiments on \aje{ultrahigh-dimensional} nonlinear feature selection using polynomial feature
mappings. \aje{Last}, we demonstrate the efficacy of FGM on group
feature selection in Section \ref{sec:exp2}.

\subsection{General Experimental Settings}\label{sec:sub_sec1}
On the linear feature selection task, comparisons are conducted
between {FGM} and $\ell_1$-regularized methods including
{$\ell_1$-SVM} and {$\ell_1$-LR}. \revise{For {FGM}, we study {FGM}
with SimpleMKL solver (denoted by {MKL-FGM})\footnote{\revise{For
fair comparison, when doing linear feature selections, we adopts the
LIBLinear ({CD-SVM}) as the SVM solver for SimpleMKL. The codes are
  available from:
http://c2inet.sce.ntu.edu.sg/Mingkui/FGM.htm.}}~\citep{tan10}, {FGM}
with APG method for the squared hinge loss (denoted by {PROX-FGM})
and the logistic loss (denoted by {PROX-SLR}), respectively.}

Many efficient batch training algorithms have been developed to
solve {$\ell_1$-SVM} and {$\ell_1$-LR}, such as the interior point
method, \aje{the} fast iterative shrinkage-threshold algorithm ({FISTA}),
\aje{the} block coordinate descent ({BCD}), \aje{the} {Lassplore}
method~\citep{Liu2010}, \aje{the} generalized linear model with elastic net
({GLMNET})\aje{, etc.}~\citep{Yuan2010jmlr,Yuan2011}. LIBLinear, which
adopts the coordinate descent method to solve non-smooth
optimization problems, has demonstrated state-of-the-art performance
in terms of training efficiency~\citep{Yuan2010jmlr}. In LIBLinear,
by taking \aje{advantage} of data sparsity, it achieves very fast
convergence \aje{speeds} for sparse datasets~\citep{Yuan2010jmlr,Yuan2011}.
\aje{Because of this}, we include the LIBLinear solver for
comparison\footnote{\aje{Code is} available from:
http://www.csie.ntu.edu.tw/$\sim$cjlin/liblinear/.}. \aje{In addition}, we
take the standard SVM and LR classifier of LIBLinear with all
features as the baselines\aje{;} denoted by {CD-SVM} and {CD-LR},
respectively.  We use the default stopping criteria of LIBLinear for
$\ell_1$-SVM, {$\ell_1$-LR}, {CD-SVM} and {CD-LR}.

SGD methods have gained \aje{a great deal of} attention for solving large-scale
problems~\citep{langford2009sparse,Shwartz2013}. In this experiment,
we include the proximal stochastic dual coordinate ascent with
logistic loss for comparison (denoted by {SGD-SLR}).
{SGD-SLR} has shown state-of-the-art performance among \aje{the} various
SGD methods~\citep{Shwartz2013}.\footnote{\aje{Code is} available from:
http://stat.rutgers.edu/home/tzhang/software.html.} In {SGD-SLR},
there are three important parameters\aje{:} $\lambda_1$ \aje{is used} to penalize
$||\w||_1$, $\lambda_2$ \aje{is used} to penalize $||\w||_2^2$, and the stopping
criterion \aje{is} $\emph{{min.dgap}}$. \aje{As suggested} by the package, in the
following experiment, we fix $\lambda_2$ = 1e-4 and
$\emph{{min.dgap}}$=1e-5 and change $\lambda_1$ to obtain different
levels of sparsity. All the methods are implemented in C++.

On  group feature selection tasks, we compare {FGM} with four
recently developed group lasso solvers: {FISTA}~\citep{Liu2010,
Jenatton2011,Bach2010convex}, \aje{the} block coordinate descent method
(denoted by {BCD})~\citep{Qin2010}, \aje{the} active set method (denoted
by {ACTIVE})~\citep{Francis2009HKL,Roth2008} and
UFO-MKL~\citep{orabona2011ultra}. Among \aje{these}, {{FISTA}} has been
thoroughly studied by several researchers~\citep{Liu2010,
Jenatton2011,Bach2010convex}, and we adopt the implementation of \aje{the}
{SLEP} package\footnote{\aje{Code is} available from:
http://www.public.asu.edu/\scriptsize{$\sim$}jye02/Software/SLEP/index.htm.},
where an improved line search is used~\citep{Liu2010}. We implement
the block coordinate descent method proposed by~\cite{Qin2010}
where each subproblem is formulated as a trust-region problem and
solved by a Newton's root-finding method~\citep{Qin2010}. For
UFO-MKL, an online optimization method,\footnote{\aje{Code is}
available from: http://dogma.sourceforge.net/} we stop the
training after 20 epochs. We adopt the {SLEP} solver to implement
the {ACTIVE} method. All the referred methods for group feature
selection are implemented in MATLAB for fair comparison.

\begin{table*}[htp]
\begin{center}
\begin{footnotesize}
\begin{tabular}{|c|c|c|c|c|c|c|c|c|}
\hline
 \multirow{2}{*}{Dataset} &  \multirow{2}{*}{$m$} & \multirow{2}{*}{$n_{\text{train}}$} & \multirow{2}{*}{$n_{\text{test}}$} &{\# nonzeros}&  \multicolumn{3}{c|}{Parameter Range}\\
\cline{6-8}
& & && per instance&{l1-SVM} ($C$)&{l1-LR}($C$)&{SGD-SLR}($\lambda_1$)\\
\hline
 {\tt epsilon} &  2,000  &     400,000 &     100,000 & 2,000& [5\text{e}-4, 1\text{e}-2]&[2\text{e}-3,  1\text{e}-1]&[1\text{e}-4,  8\text{e}-3]\\
\hline
 {\tt aut-avn} &     20,707 &   40,000 &     22,581& 50 &--&--&--\\
 \hline
 {\tt real-sim} &     20,958 &     32,309 &     40,000 & 52 &[5\text{e}-3, 3\text{e}-1]&[5\text{e}-3, 6\text{e}-2]&[1\text{e}-4,  8\text{e}-3]\\
\hline
{\tt rcv1} &     47,236 &   677,399  &  20,242& 74 &[1\text{e}-4,  4\text{e}-3]&[5\text{e}-5,  2\text{e}-3]&[1\text{e}-4,  8\text{e}-3]\\
\hline
{\tt astro-ph} &    99,757  &     62,369 &     32,487 & 77&[5\text{e}-3, 6\text{e}-2]&[2\text{e}-2, 3\text{e}-1]&[1\text{e}-4,  8\text{e}-3]\\
\hline
{\tt news20} &  1,355,191 &      9,996 &     10,000 &359&[5\text{e}-3, 3\text{e}-1]&[5\text{e}-2, 2\text{e}1]&[1\text{e}-4,  8\text{e}-3]\\
\hline
{\tt kddb} &  29,890,095  &     19,264,097 &     748,401 & 29&[5\text{e}-6, 3\text{e}-4]&[3\text{e}-6, 1\text{e}-4]&[1\text{e}-4,  8\text{e}-3]\\
\hline
\end{tabular}
\caption{Statistics \aje{for} the datasets used in the experiments.
{Parameter Range} lists the ranges of the
 parameters for various $\ell_1$-methods to select
different \aje{numbers} of features.  The datasets {\tt rcv1} and {\tt
aut-avn} will be used in group feature selection tasks.  }
\label{largedataset}
\end{footnotesize}
\end{center}
\vskip -0.2in
\end{table*}

Several large-scale and \aje{high-dimensional} real-world datasets are
used to verify the performance of different methods.  General
information \aje{on} these datasets, such as the average nonzero features
per instance, is listed in Table~\ref{largedataset}.\footnote{Among
these datasets, {\tt epsilon}, {\tt real-sim}, {\tt rcv1.binary},
{\tt news20.binary} and {\tt kddb} can be downloaded
from:~\emph{http://www.csie.ntu.edu.tw/$\sim$cjlin/libsvmtools/datasets/},
{\tt aut-avn} can be downloaded from
\emph{http://vikas.sindhwani.org/svmlin.html} and  {\tt Arxiv
astro-ph} is from \citep{Joachims2006}.} Among \aje{these}, {\tt epsilon},
{\tt Arxiv astro-ph}, {\tt rcv1.binary} and {\tt kddb} datasets have
 been split into training \aje{and testing sets}. For {\tt real-sim}, {\tt aut-avn} and {\tt news20.binary}, we
randomly split them into training  and testing sets, as shown in
Table~\ref{largedataset}.

All the comparisons are performed on a 2.27GHZ Intel(R)Core(TM) 4
DUO CPU running windows
sever 2003 with 24.0GB \aje{of} main memory.

\subsection{Synthetic Experiments on Linear Feature Selection}\label{sec:exp1}
In this section, we compare the performance of different methods on
two toy datasets of different scales, namely
${\bX}\in\R^{4,096\times 4,096}$ and ${\bX}\in\R^{8,192\times
65,536}$. Here\aje{,} each $\bX$ is a Gaussian random matrix with each
entry sampled from the i.i.d. Gaussian distribution
$\mathcal{N}(0,1)$. To produce the output $\y$, we first generate a
sparse vector  $\w$ with 300 nonzero entries with each nonzero
entry sampled from the i.i.d. Uniform distribution
$\mathcal{U}(0,1)$. After \aje{this}, we produce the output by {$\y =
\text{sign}(\bX\w)$}. Since only the nonzero $w_i$ contributes to
the output $\y$, we consider the corresponding feature as a relevant
feature \aje{with respect to} $\y$. Similarly, we generate the testing dataset
{${\bX}_{\text{test}}$} with output labels {$\y_{\text{test}} =
\text{sign}({\bX}_{\text{test}}\w)$}. The number of testing points
for both cases is set to 4,096.

\subsubsection{Convergence Comparison of Exact and Inexact {FGM}}
\begin{figure*}
\center \subfigure[Relative objective values w.r.t. {APG}
iterations.]{
    \label{fig:subfig_feat_ITER} 
    \includegraphics[trim = 1.5mm 0mm 6mm 2mm,  clip,  width=2.35in]{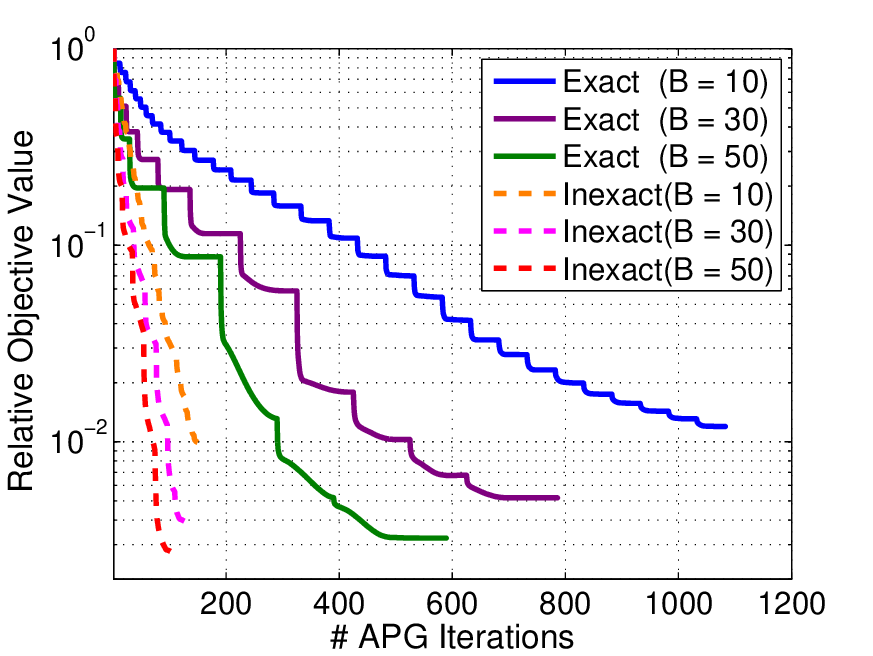}}\hspace{0.4in}
  \subfigure[Relative objective values w.r.t. outer
iterations.]{
    \label{fig:subfig_feat_diff} 
    \includegraphics[trim = 1.5mm 0mm 6mm 2mm,  clip,  width=2.35in]{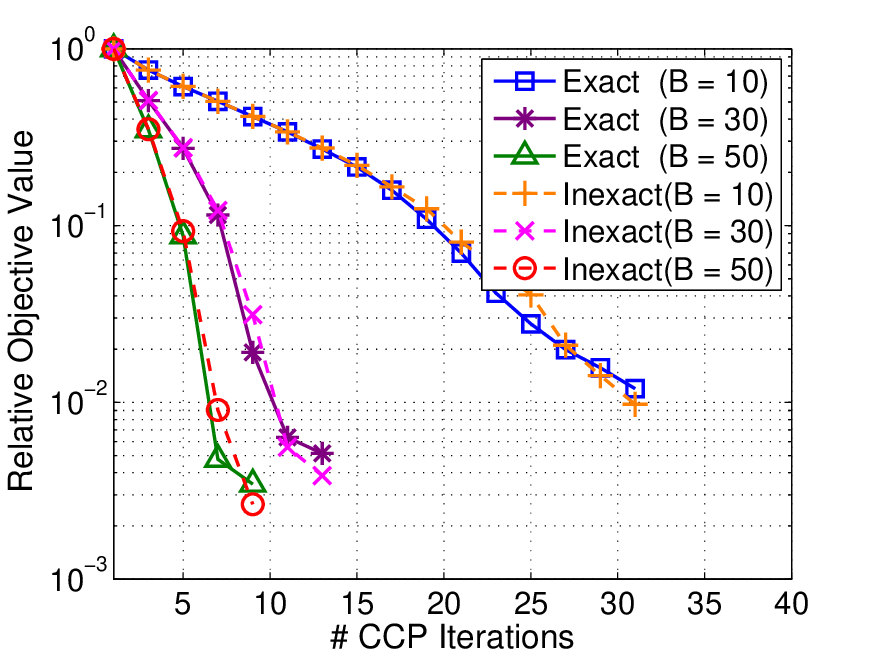}}
\caption {Convergence of Inexact {FGM} and Exact {FGM} on the
synthetic dataset.} \label{fig:toy_convergence}
\end{figure*}

In this experiment, we study the convergence of \emph{Exact} and
\emph{Inexact} {FGM} on \aje{a} small scale dataset. To study the
\emph{Exact} {FGM}, for simplicity, we set the stopping tolerance
$\epsilon_{in} = 1.0\times 10^{-6}$ in (\ref{eq:stop_inner}) for
{APG}\aje{,} while for \emph{Inexact} {FGM}, we set $\epsilon_{in} =
1.0\times 10^{-3}$. We  set $C=10$ and test different $B$'s from
$\{10, 30, 50\}$. In this experiment, only the squared hinge loss is
studied.  In Figure~\ref{fig:subfig_feat_ITER}, we report the
relative objective values  w.r.t. all the {APG} iterations for both
methods\aje{.} In Figure~\ref{fig:subfig_feat_diff}, we report the
relative objective values w.r.t. the outer iterations. We have the
following observations from Figures~\ref{fig:subfig_feat_ITER} and~\ref{fig:subfig_feat_diff}.

\aje{First}, from Figure \ref{fig:subfig_feat_ITER}, for each comparison
method, the function value sharply decreases at some iterations
where an active constraint is added. For the \emph{Exact} {FGM}, 
more {APG} iterations \aje{are required for} the tolerance $\epsilon_{in} =
1.0\times 10^{-6}$\aje{. However,} the function value does not show \aje{a} significant decrease after several {APG} iterations. \aje{In contrast}, from
Figure~\ref{fig:subfig_feat_ITER}, the \emph{Inexact} {FGM}, which
uses a relatively larger tolerance $\epsilon_{in} = 1.0\times
10^{-3}$, requires much fewer APG iterations to achieve similar
objective values \aje{as} \emph{Exact} {FGM} \aje{for} the same parameter $B$.
Particularly, from Figure~\ref{fig:subfig_feat_diff}, the
\emph{Inexact} {FGM} achieves similar objective values \aje{as}
\emph{Exact} {FGM} after each outer iteration. According to these
observations, on one hand, $\epsilon_{in}$ should be small enough
such that the subproblem can be sufficiently optimized. On the other
hand, a relatively large tolerance (e.g.\aje{,} $\epsilon_{in} = 1.0\times
10^{-3}$) can greatly accelerate the convergence speed without
degrading the performance.

Moreover, according to Figure \ref{fig:subfig_feat_diff}, {PROX-FGM}
with a large $B$ in general converges faster than with a small
$B$. Generally, by using a large $B$, \aje{lower numbers} of outer
iterations and worst-case \aje{analyses} are required, which is critical
when dealing with \emph{Big Data}. However, if $B$ is too large,
some non-informative features may be mistakenly included, and the
solution may not be exactly sparse.

\subsubsection{Experiments on Small-Scale Synthetic Dataset}\label{sec:smalltoy}
\begin{figure*}[h]
  \centering
    {\includegraphics[trim = 3.0cm 4.2cm 3.5cm 22.8cm,  clip,  width=0.94\textwidth]{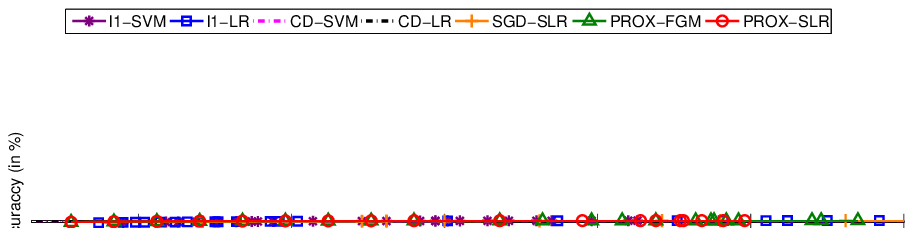}}\vspace{-6mm} \\
  \center
  \subfigure[{Testing accuracy}]{
    \label{fig:subfig_feat_accI} 
    \includegraphics[trim = 1.5mm 0mm 6mm 2mm,  clip,  width=2.35in]{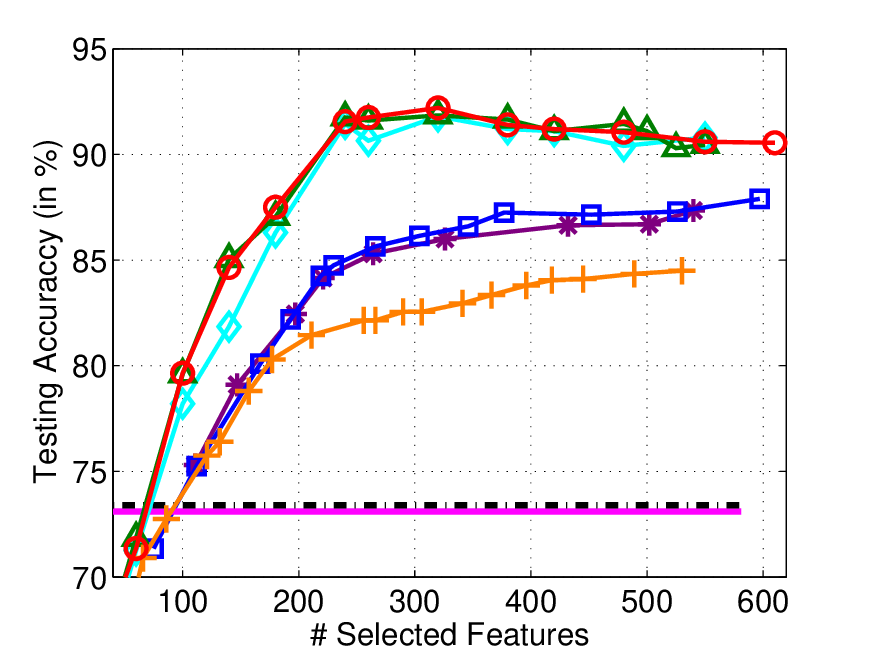}}\hspace{0.4in}
  \subfigure[{Number of recovered features}]{
    \label{fig:subfig_feat_num} 
    \includegraphics[trim = 1.5mm 0mm 6mm 2mm,  clip,
    width=2.35in]{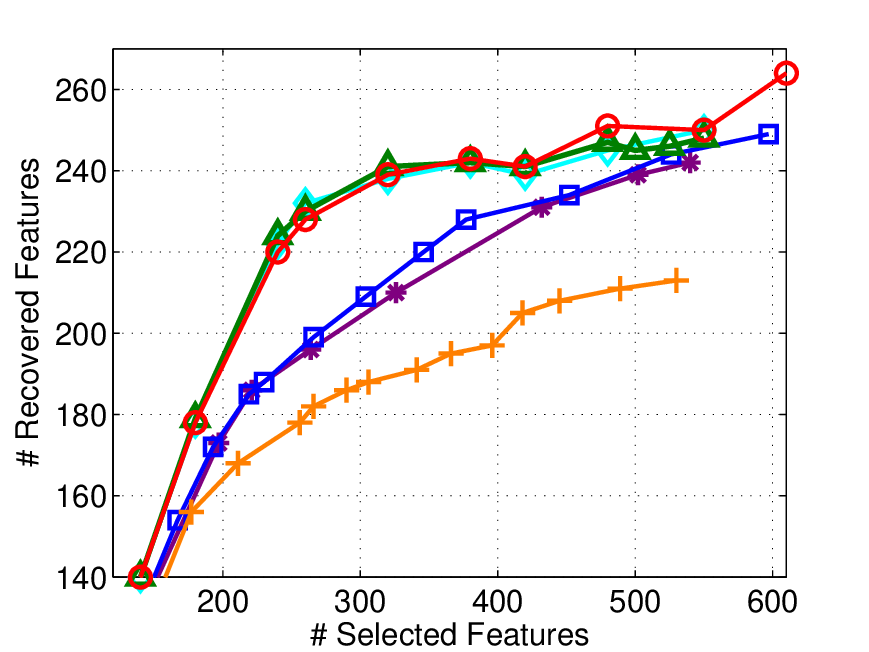}}\\
   \subfigure[{De-biased results}]{
    \label{fig:subfig_feat_unbias}
    \includegraphics[trim = 1.5mm 0mm 6mm 2mm,  clip,  width=2.35in]{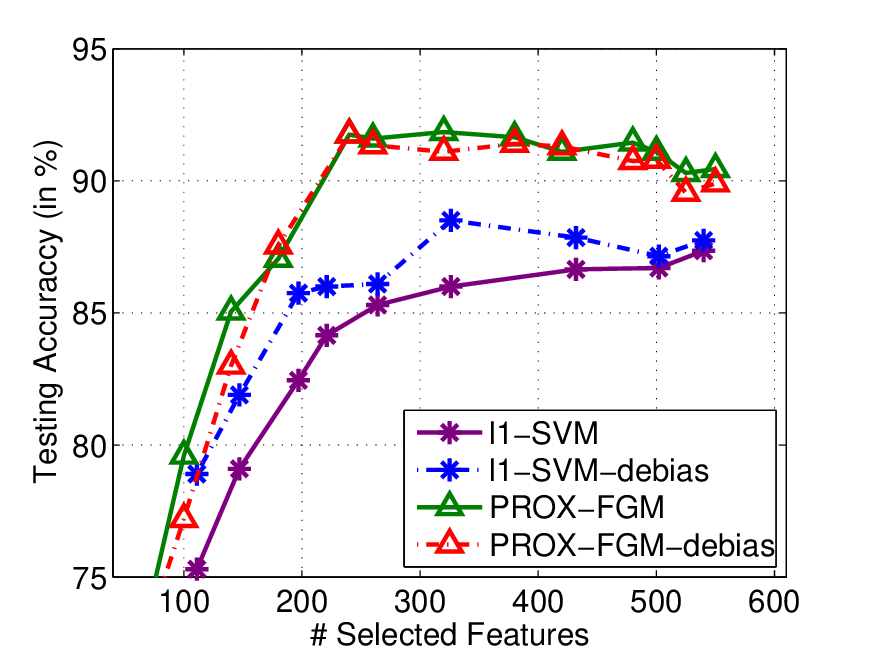}}\hspace{0.4in}
    \subfigure[{Training time}]{
    \label{fig:subfig_feat_time}
    \includegraphics[trim = 1.5mm 0mm 6mm 2mm,  clip,  width=2.35in]{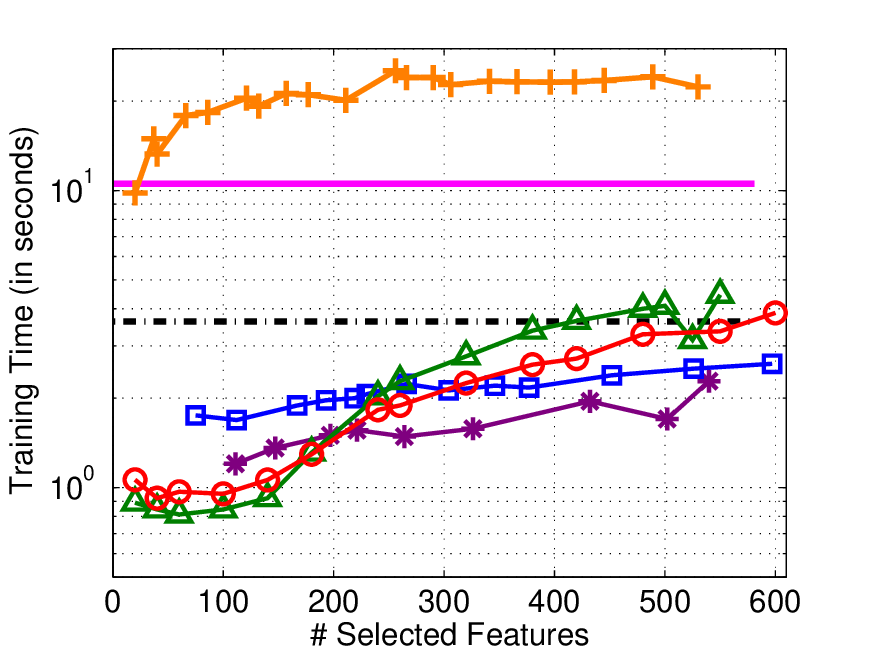}}
\caption{Experimental results on the small dataset, where {CD-SVM}
and {CD-LR} denote the results of standard SVM and LR with all
features, respectively. The training time of {MKL-FGM} is about
1,500 seconds, which is up to 1,000 times slower than {APG} solver.
We did not report it in the figures due to presentation issues.}
  \label{fig:toy_4096_acc}
\end{figure*}

In this experiment, we evaluate the performance of different methods
in terms of testing accuracies w.r.t. different \aje{numbers} of selected
features. Specifically,  to obtain sparse solutions \aje{with} different
sparsities, we vary $C \in [0.001, 0.007]$ for {l1-SVM}, $C \in$
[5\text{e}-3, 4\text{e}-2] for {l1-LR} and $\lambda_1 \in$
[7.2\text{e}-4, 2.5\text{e}-3] for {SGD-SLR}.\footnote{Here, we
carefully choose $C$ or $\lambda_1$ for these three $\ell_1$-methods
such that the numbers of selected features \aje{are} uniformly spread over the
range [0, 600]. Since the values of $C$ and $\lambda_1$ \aje{exhibit large changes} for different problems, we \aje{hereafter} only give their ranges.
Notice that, under this experimental setting, the results of
$\ell_1$-methods cannot be further improved through parameter
tunings.} \aje{In contrast} to these methods, we fix $C=10$ and choose
even numbers in $\{2, 4, ..., 60 \}$  for $B$ to obtain different
\aje{numbers} of features. It can be seen that it is much easier for {FGM}
to control the number of features \aje{being} selected. Specifically, the
testing accuracies and the number of recovered ground-truth features
w.r.t. the number of selected features are reported in Figure~\ref{fig:subfig_feat_accI} and Figure~\ref{fig:subfig_feat_num},
respectively. The training \aje{times for different methods are} listed in
Figure~\ref{fig:subfig_feat_time}.

For convenience of presentation, let $m_s$ and $m_g$ be the number
of selected features and the number of ground-truth features,
respectively. From Figure~\ref{fig:subfig_feat_accI} and Figure~\ref{fig:subfig_feat_num}, {FGM}\aje{-}based methods demonstrate better
testing accuracy than all $\ell_1$-methods when $m_s>100$.
Correspondingly, from Figure~\ref{fig:subfig_feat_num}, \aje{with} the
same number of selected features, {FGM}\aje{-}based methods include more
ground-truth features than $\ell_1$-methods when $m_s$$\geq$100. 
{SGD-SLR} shows the worst testing accuracy among the \aje{compared}
methods and recovers the least number of ground-truth features.

One of the possible reasons for the inferior performance of the
$\ell_1$-methods, as mentioned in the Introduction section, is the
solution bias \aje{caused} by the $\ell_1$-regularization. To demonstrate
this, we do \aje{retraining} to reduce the bias using {CD-SVM} with
$C=20$ with the selected features\aje{. Then, we} do the prediction using
the de-biased models. The results are reported in Figure~\ref{fig:subfig_feat_unbias}, where {l1-SVM-debias} and
{PROX-FGM-debias} denote the de-biased counterparts \aje{of} {l1-SVM} and
{PROX-FGM}, respectively. In general, \emph{if there was no feature
selection bias, both {FGM} and {l1-SVM} should have similar
testing \aje{accuracies as} their de-biased counterparts}. However, from
Figure~\ref{fig:subfig_feat_unbias}, {l1-SVM-debias} in general has
much better testing accuracy than {l1-SVM}\aje{,} while {PROX-FGM} has
similar or even better testing accuracy than {PROX-FGM-debias} and
{l1-SVM-debias}. These observations indicate that 1) the solution
bias indeed exists in $\ell_1$-methods and affects the feature
selection performance\aje{, and} 2) {FGM} can reduce the feature selection
bias.

From Figure \ref{fig:subfig_feat_time},  on this small-scale
dataset, {PROX-FGM} and {PROX-SLR} achieve comparable efficiency
\aje{as} the LIBlinear solver. \aje{In contrast}, {SGD-SLR}, which is a
typical stochastic gradient method, spends the longest \aje{time training}. 
This observation indicates that \aje{the} SGD-SLR method may not be
suitable for small-scale problems. \aje{Last}, as reported in the
caption of Figure~\ref{fig:subfig_feat_time}, {PROX-FGM} and
{PROX-SLR} are up to 1,000 times faster than {MKL-FGM} using \aje{the}
SimpleMKl solver. The reason \aje{for this} is that SimpleMKl uses the subgradient \aje{method} to address the non-smooth optimization problem with $n$
variables\aje{, while} the subproblem is
solved in the primal problem w.r.t. a small number of selected
variables \aje{in {PROX-FGM} and {PROX-SLR}}.

\aje{Last}, from Figure \ref{fig:toy_4096_acc}, if the number of
selected features is small ($m_s<100$), the testing accuracy is
worse than {CD-SVM} and {CD-LR} with all features. However, if \aje{a}
sufficient number ($m_s > 200$) of features are selected, the
testing accuracy is much better than {CD-SVM} and {CD-LR} with all
features, which verifies the importance of feature selection.

\begin{figure*}[h]
\centering
    {\includegraphics[trim = 8.5cm 8.5cm 9cm 2cm,  clip,  width=0.64\textwidth]{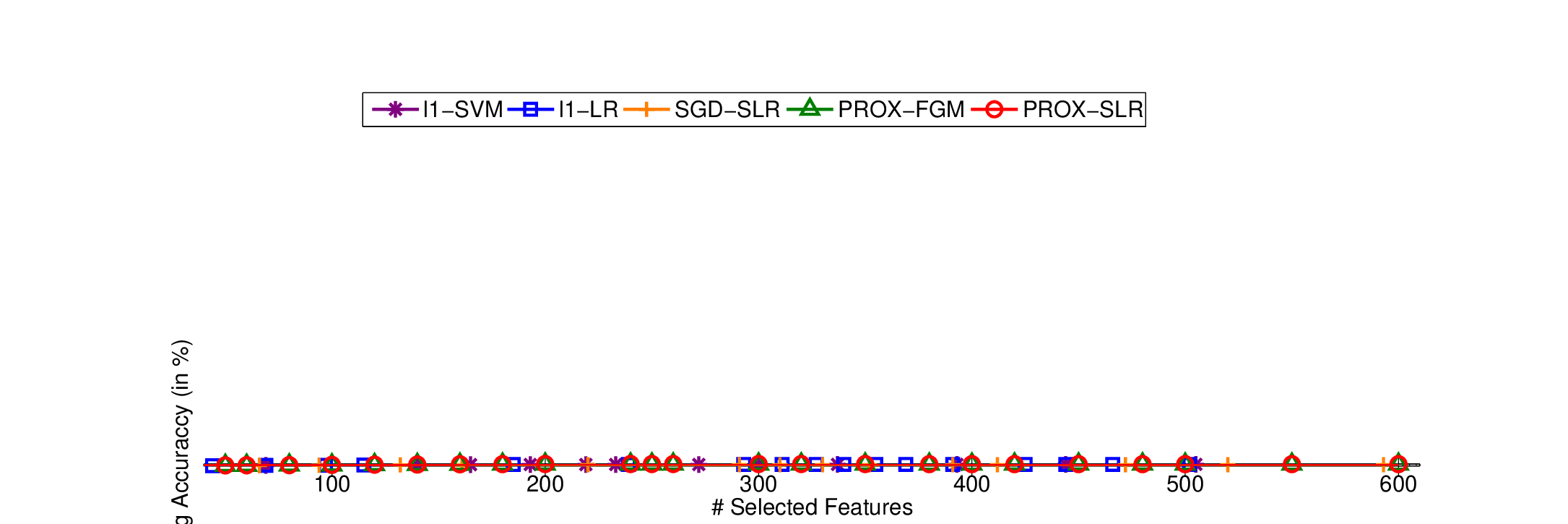}}\vspace{-6mm} \\
  \center
  \subfigure[Testing accuracy]{
    \label{fig:subfig_65536_acc} 
    \includegraphics[trim = 1.5mm 0mm 6mm 2mm,  clip,  width=2.35in]{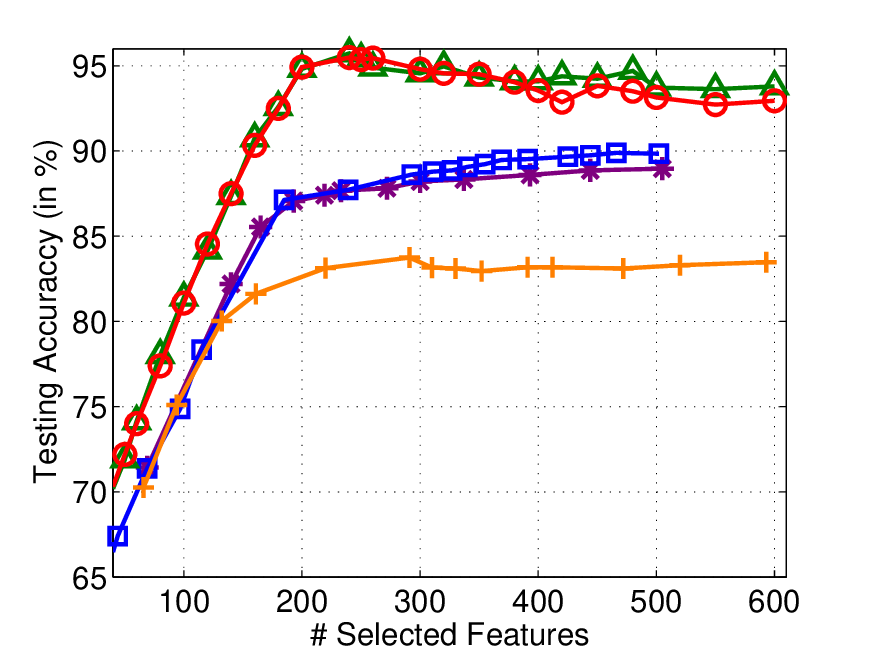}}\hspace{0.4in}
     \subfigure[Number of recovered features]{
    \label{fig:subfig_65536_spar}
    \includegraphics[trim = 1.5mm 0mm 6mm 2mm,  clip,
    width=2.35in]{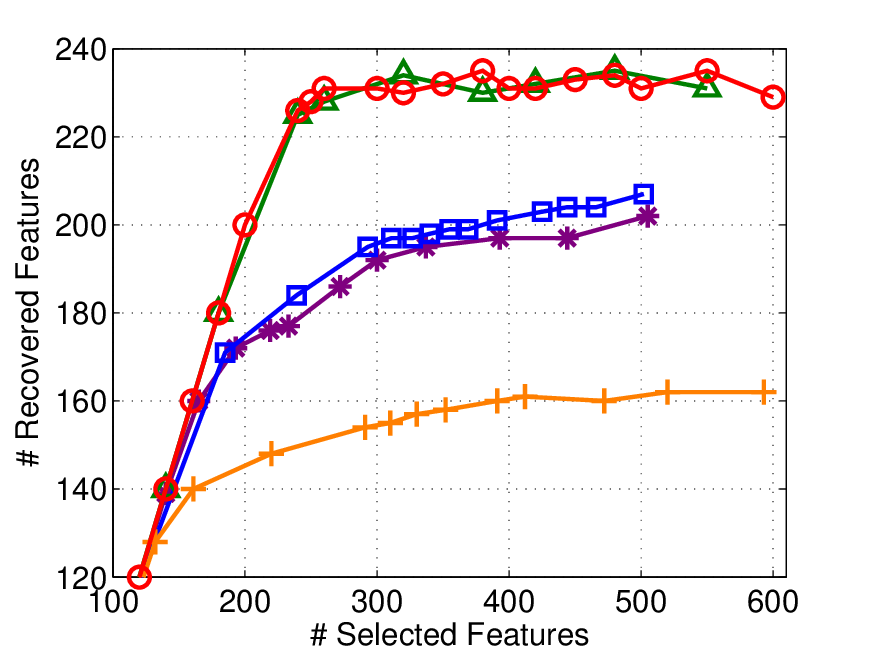}}\\
   \subfigure[{ De-biasing effect of {FGM}}]{
    \label{fig:65536_feat_unbias}
    \includegraphics[trim = 1.5mm 0mm 6mm 2mm,  clip,  width=2.35in]{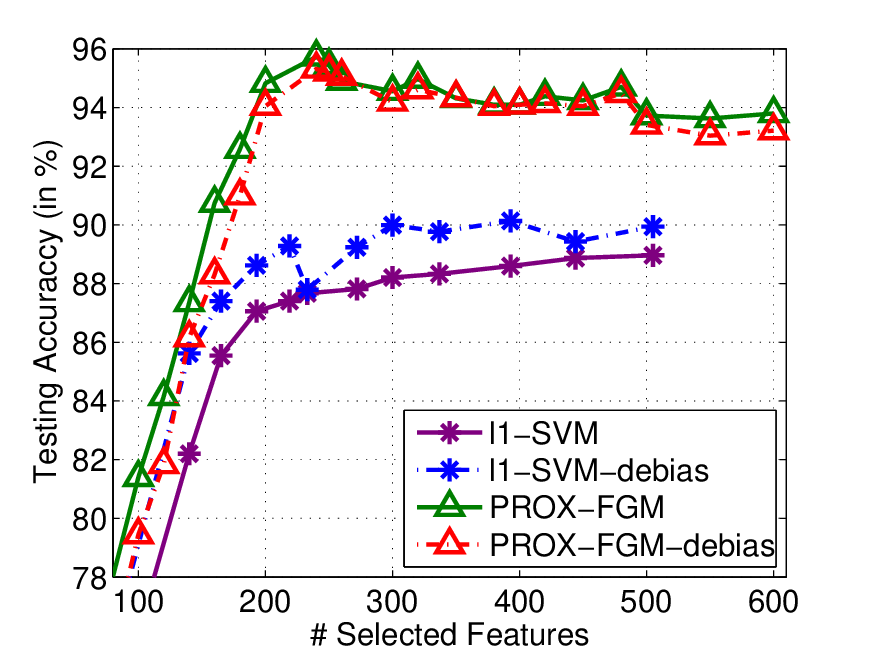}}\hspace{0.4in}
     \subfigure[Training time]{
    \label{fig:subfig_65536_time} 
    \includegraphics[trim = 1.5mm 0mm 6mm 2mm,  clip,  width=2.35in]{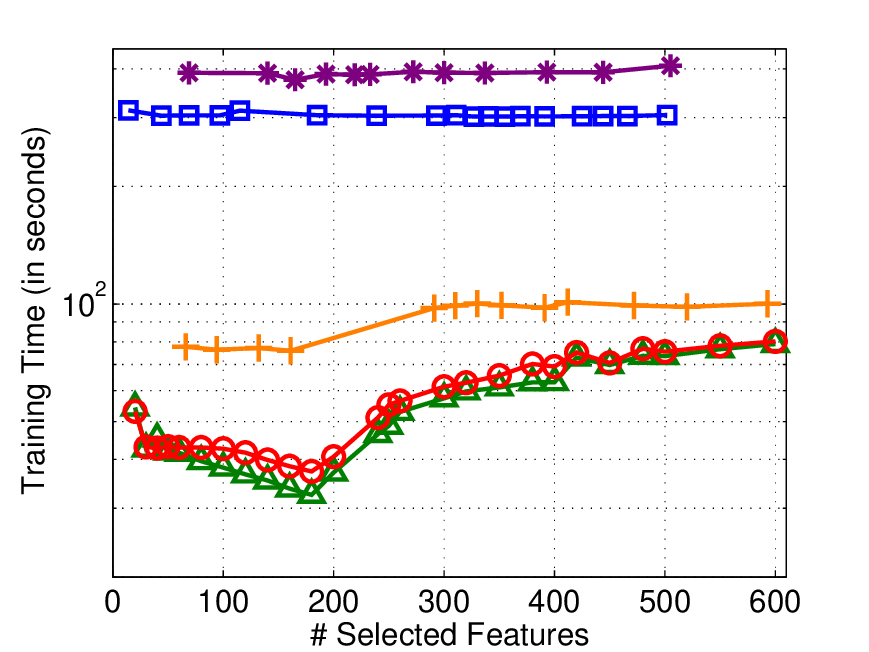}}
  \caption{Performance comparison on the large-scale synthetic dataset.}
  \label{fig:toy_feature_65536}
\end{figure*}

\subsubsection{Experiments on Large-scale Synthetic Dataset}\label{sec:largetoy}
To demonstrate the scalability of FGM, we conduct an experiment on a
large-scale synthetic dataset, namely\aje{,} ${\bX}\in\R^{8,192\times
65,536}$. Here, we do not include the comparisons with  {MKL-FGM}
due to its high computational cost. For {PROX-FGM} and {PROX-SLR},
we follow their experimental settings  above. For {l1-SVM} and
{l1-LR}, we vary $C \in [0.001, 0.004]$ and $C \in [0.005, 0.015]$\aje{, respectively,} to determine the number of features to be selected.
The testing accuracy, the number of recovered ground-truth features,
the de-biased results and the training time of the compared methods
are reported in Figure~\ref{fig:subfig_65536_acc},
\ref{fig:subfig_65536_spar}, \ref{fig:65536_feat_unbias} and
\ref{fig:subfig_65536_time}, respectively.

From Figure \ref{fig:subfig_65536_acc}\aje{,}
\ref{fig:subfig_65536_spar} and \ref{fig:65536_feat_unbias}, both
{PROX-FGM} and {PROX-SLR} outperform {l1-SVM}, {l1-LR} and {SGD-SLR}
in terms of both testing accuracy and the number of recovered
ground-truth features. From Figure~\ref{fig:subfig_65536_time},
{PROX-FGM} and {PROX-SLR} show better training \aje{efficiencies} than the
coordinate based methods (namely, LIBlinear) and the SGD based
method (namely\aje{,} SGD-SLR). Basically, {FGM} solves a sequence of small
optimization problems \aje{with} $O(ntB)$ cost and spends only a small
number of iterations to do the worst-case analysis \aje{with} $O(mn)$ cost.
\aje{In contrast}, the $\ell_1$-methods may take many iterations to
converge, and each iteration takes $O(mn)$ cost. On this large-scale
dataset, {SGD-SLR} shows \aje{a} faster training speed than LIBlinear\aje{. However, it has a highly} inferior performance in terms of testing accuracy \aje{compared to} the LIBlinear solver.

\begin{figure*}[h]
  \centering
  \subfigure[Testing accuracy]{
    \label{fig:density_acc} 
    \includegraphics[trim = 1.5mm 0mm 6mm 2mm,  clip, width=2.35in]{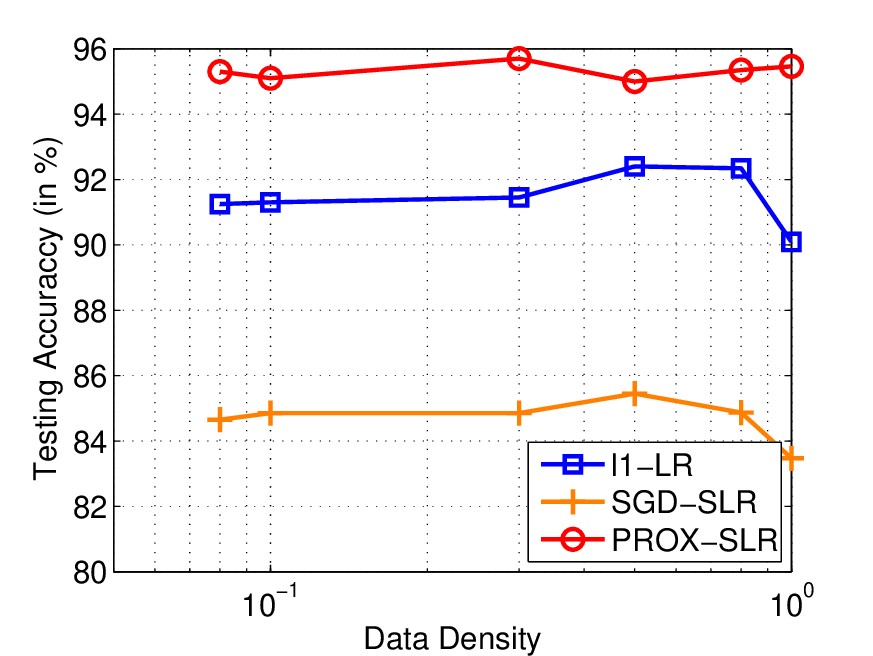}}\hspace{0.4in}
     \subfigure[Training time]{
    \label{fig:density_time} 
    \includegraphics[trim = 1.5mm 0mm 6mm 2mm,  clip, width=2.35in]{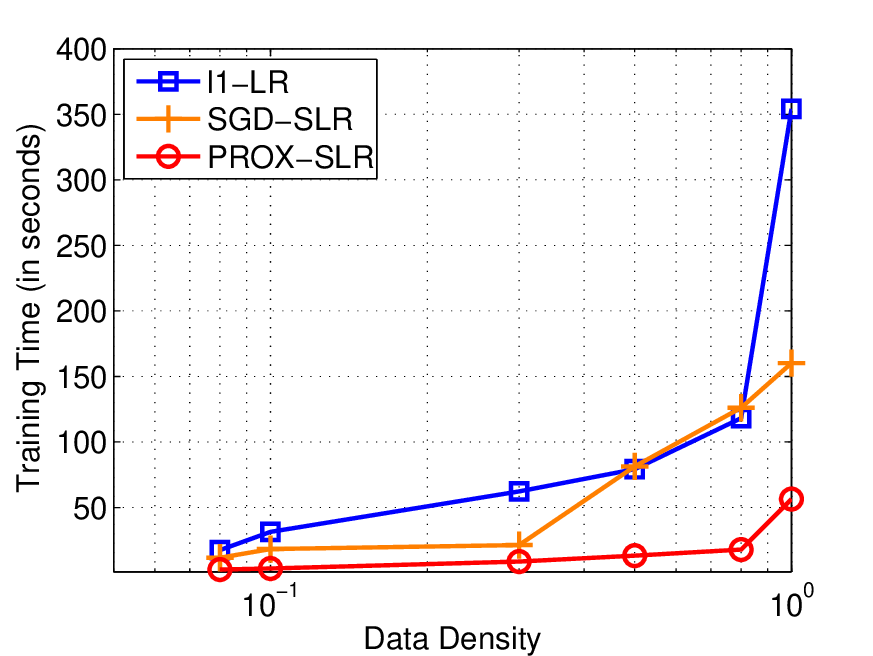}}
 \caption{Performance comparison  on the synthetic dataset
($n = 8,192$, $m=65,536$) with different data densities in $\{0.08,
0.1, 0.3, 0.5, 0.8, 1\}$. }
  \label{fig:density}
\end{figure*}

\begin{figure*}[h]
\centering \subfigure[Synthetic dataset ($b=4$)]{
\label{fig:sub_bias}
    \includegraphics[trim = 1.5mm 0mm 6mm 2mm,  clip,
    width=1.9in]{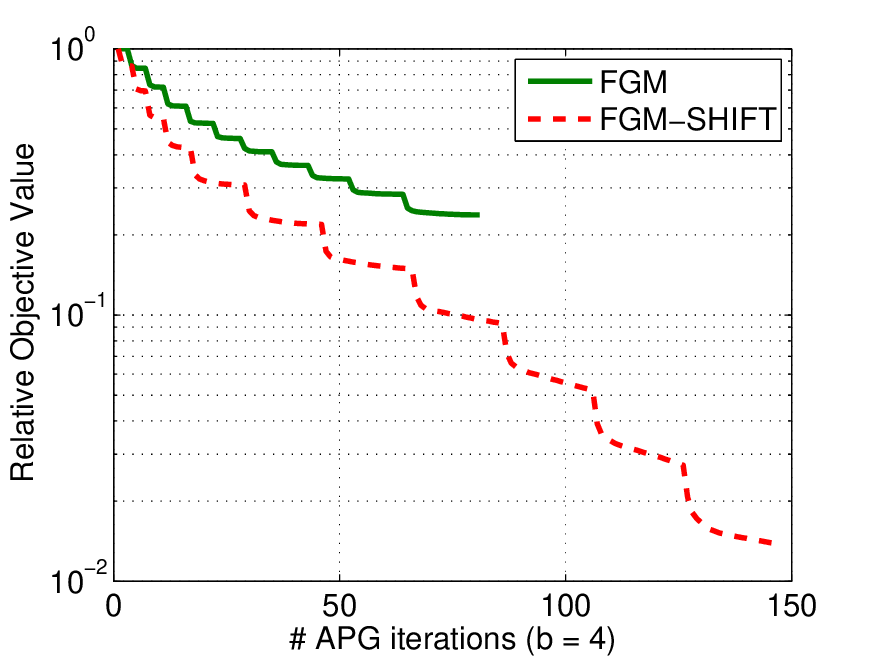}}
\subfigure[{\tt astro-ph}]{
\includegraphics[trim = 1.5mm 0mm 6mm 2mm,  clip,
width=1.9in]{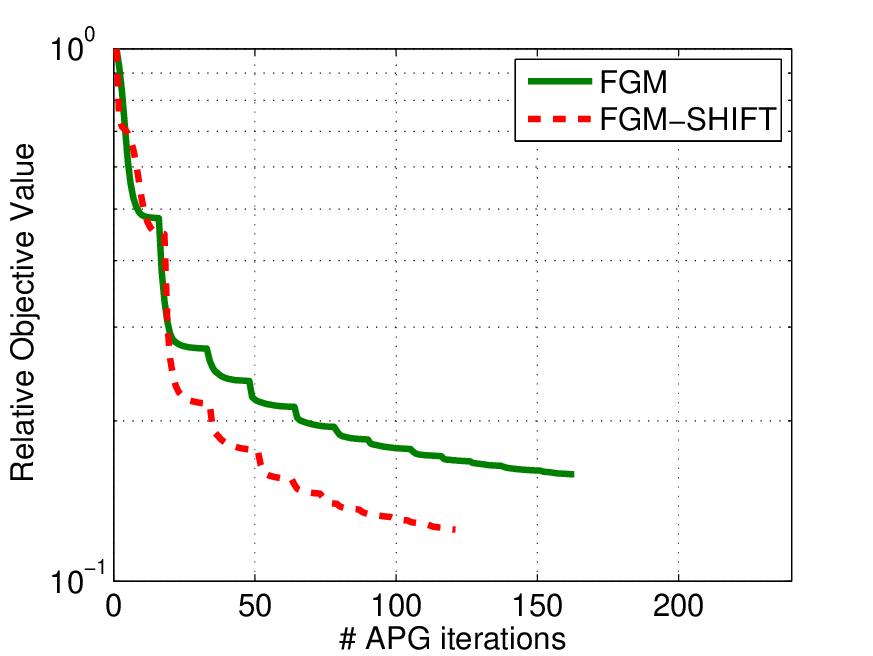}}
  \subfigure[{\tt real-sim}]{
    \includegraphics[trim = 1.5mm 0mm 6mm 2mm,  clip,  width=1.9in]{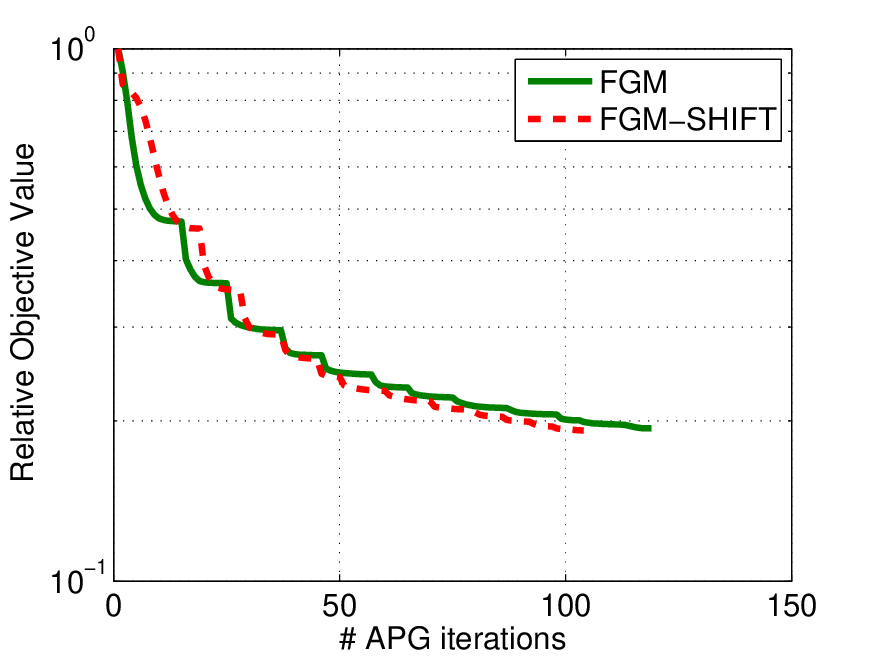}}
\caption {Relative objective values regarding each {APG} iteration,
where $b=4$ in the caption of Figure~\ref{fig:sub_bias}  denotes the
ground-truth shift of the hyperplane from the origin.}
\label{fig:toy_convergence_bias} \vskip -0.25in
\end{figure*}

In LIBlinear, the efficiency  has been improved by taking
advantage of the data sparsity. Considering this, we investigate the
sensitivity of the referred methods to the data density. To this
end, we generate datasets of different data densities by sampling
the entries from $\bX^{8,192\times 65,656}$ with \aje{a} sampling rate in
$\{0.08, 0.1, 0.3, 0.5, 0.8, 1\}$ and study the influence of the
data density \aje{on} different learning algorithms. For FGM, only
the logistic loss is studied (namely\aje{,} {PROX-SLR}). We keep the
default experimental settings for {PROX-SLR} and watchfully vary $C
\in [0.008, 5]$ for {l1-LR} and $\lambda_1 \in [\text{9.0e-4,
3e-3}]$ for {SGD-SLR}. For \aje{the} sake of brevity, we only report the
\textbf{{best accuracy}} obtained among all parameters and the
corresponding training time of {l1-LR}, {SGD-SLR} and {PROX-SLR} in
Figure~\ref{fig:density}.

From Figure \ref{fig:density_acc},  under different data densities,
{PROX-SLR} always outperforms {l1-SVM} and {SGD-SLR} in terms of the
\textbf{best accuracy}. From Figure \ref{fig:density_time}, {l1-SVM}
shows comparable efficiency with {PROX-SLR} on datasets of low data
density. However, on \aje{relatively} denser datasets,  {PROX-SLR} is much
more efficient than {l1-SVM}, which indicates that {FGM} has a
better scalability than {l1-SVM} on dense data.

\begin{figure*}[h]
  \centering
  \subfigure[Synthetic dataset ($b=4$)]{
    \includegraphics[trim = 1.5mm 0mm 6mm 2mm,  clip,  width=1.9in]{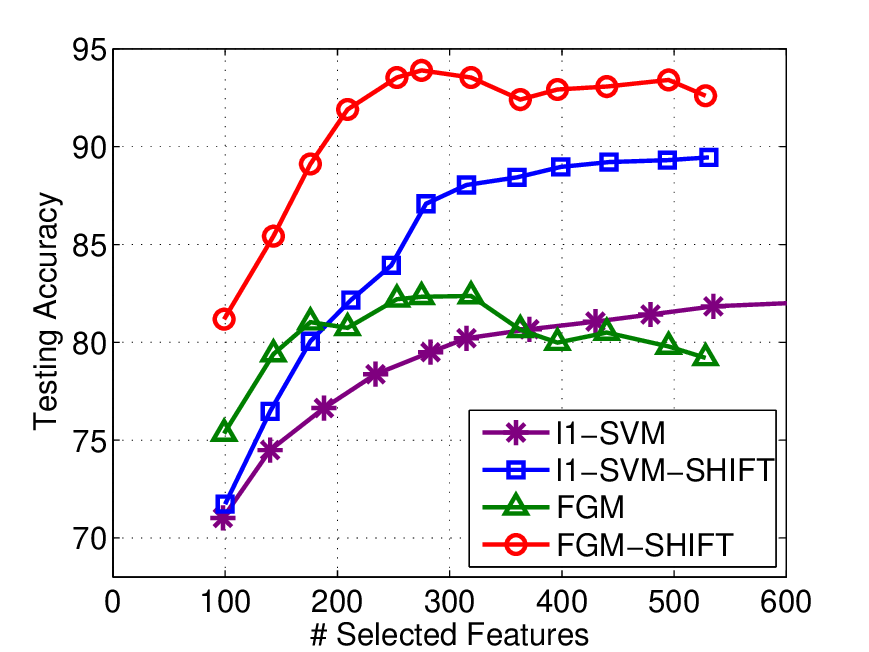}}
  \subfigure[{\tt astro-ph}]{
    \includegraphics[trim = 1.5mm 0mm 6mm 2mm,  clip, width=1.9in]{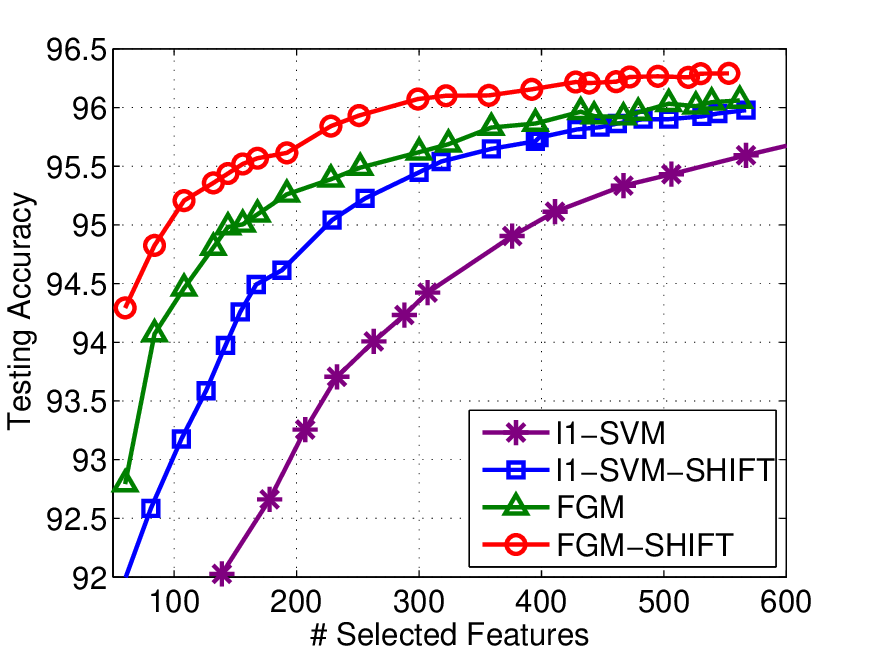}}
     \subfigure[{\tt real-sim}]{
    \includegraphics[trim = 1.5mm 0mm 6mm 2mm,  clip, width=1.9in]{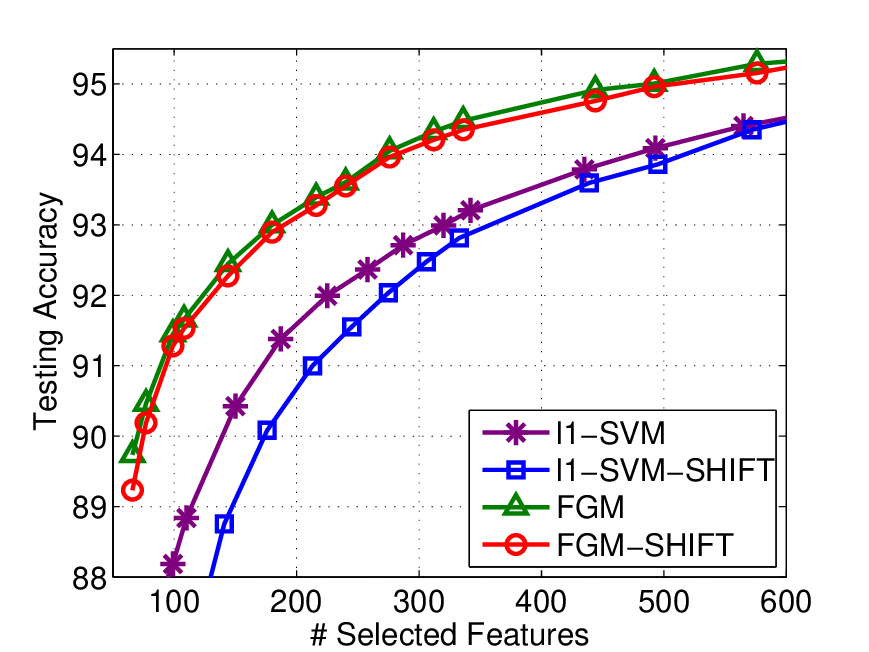}}
\caption {Testing \aje{accuracies} of different methods on  the three
datasets.} \label{fig:toy_shift}\vskip -0.05in
\end{figure*}

\subsection{Feature Selection with Shift
Consideration}\label{exp:shift}

In this section, we study the effectiveness of the shift version of
FGM (denoted by FGM-SHIFT) on a synthetic dataset and two real-world
datasets, namely\aje{,} {\tt real-sim} and {\tt astro-ph}. We follow the
data generation \aje{procedure} in Section 7.1 to generate the synthetic dataset
except that we include a shift term $b$ for the hyperplane when
generating the output $\y$. Specifically, we produce $\y$ by $\y =
\text{sign}(\bX\w - b\1)$ where $b=4$. The
 shift version of $\ell_1$-SVM by LIBlinear (denoted by l1-SVM-SHIFT) is adopted as the
baseline. In Figure~\ref{fig:toy_convergence_bias}, we report the
relative objective values of {FGM} and {FGM-SHIFT} w.r.t. the {APG}
iterations on three datasets. In Figure~\ref{fig:toy_shift}, we
report the testing accuracy versus different \aje{numbers} of selected
features.

 From Figure
\ref{fig:toy_convergence_bias}, {FGM-SHIFT} indeed achieves much
lower objective values than FGM on the synthetic dataset and \aje{the} {\tt
astro-ph} dataset, which demonstrates the effectiveness of
{FGM-SHIFT}. On the {\tt real-sim} dataset, {FGM} and {FGM-SHIFT}
achieve similar objective values, which indicates that the shift
term on {\tt real-sim} is not significant. As a result, {FGM-SHIFT}
may not significantly improve the testing accuracy.

 From Figure \ref{fig:toy_shift}, on the synthetic dataset and {\tt astro-ph} dataset,
FGM-SHIFT shows \aje{significantly} better testing accuracy than the
baseline methods, which coincides with the better objective values
of FGM-SHIFT in Figure~\ref{fig:toy_convergence_bias}. l1-SVM-SHIFT
also shows better testing accuracy than l1-SVM, which verifies the
importance of shift consideration for l1-SVM. However, on the {\tt
real-sim} dataset, the methods with shift show similar or even
inferior \aje{performance} over the methods without shift consideration,
which indicates that the shift of the hyperplane from the origin is
not significant on the {\tt real-sim} dataset. \aje{Last}, FGM and
FGM-SHIFT are always better than the counterparts of l1-SVM.

\subsection{Performance Comparison on Real-World Datasets}\label{sec:real-world} \vspace{-0.05in}
In this section, we conduct three experiments to compare the
performance of FGM with the referred baseline methods on real-world
datasets. \aje{First}, in Section \ref{real_exp}, we compare the
performance of different methods on six real-world datasets.
\aje{Second}, we study the de-biased results in Section
\ref{real_exp_debias}. \aje{Last}, we conduct \aje{a} sensitivity study of
parameters for FGM in Section \ref{exp_real_para}.

\subsubsection{Experimental Results on Real-World Datasets}\label{real_exp}

On real-world datasets, the number of ground-truth features is
unknown. We only report the testing accuracy versus different \aje{numbers}
of selected features. For FGM, we fix $C = 10$, and vary $B \in \{2,
4, ..., 60\}$ to select different \aje{numbers} of features. For the
$\ell_1$-methods, we watchfully vary the regularization parameter to
select different \aje{numbers} of features. The ranges of $C$ and
$\lambda_1$ for $\ell_1$-methods are listed in Table~\ref{largedataset}.

\begin{figure*}[h]
 \centering
    {\includegraphics[trim = 2.5cm 3.7cm 2.5cm 22.6cm,  clip,  width=0.94\textwidth]{legend_real_data.eps}}\vspace{-3mm} \
  \centering
    \subfigure[{\tt epsilon}]{
    \label{fig:subfig_feat_acc} 
    \includegraphics[trim = 1.5mm 0mm 6mm 2mm,  clip, height = 1.75in, width=2.35in]{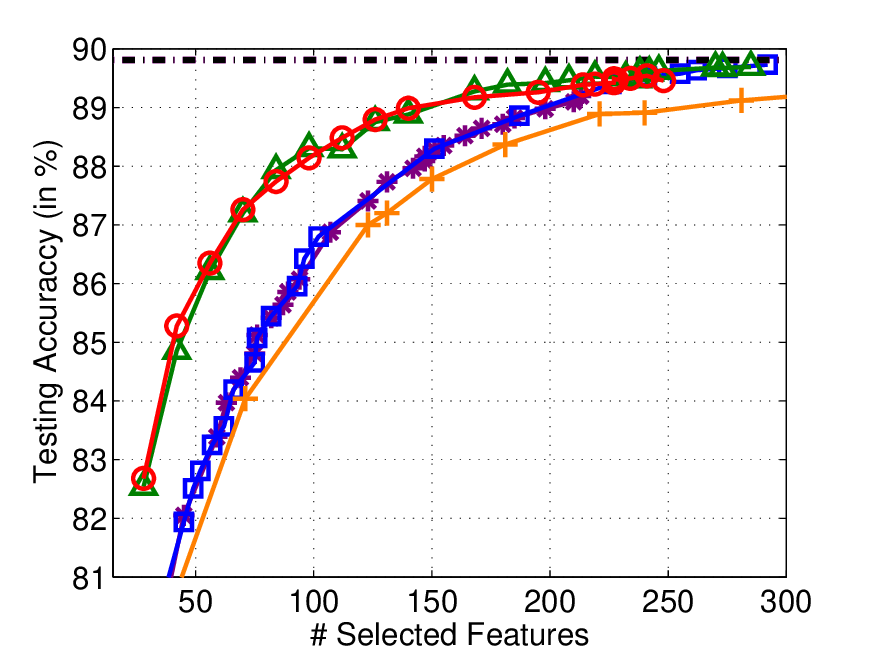}}
        \hspace{0.4in}
  \subfigure[{\tt real-sim}]{
    \includegraphics[trim = 1.5mm 0mm 6mm 2mm,  clip,  height = 1.75in,width=2.35in]{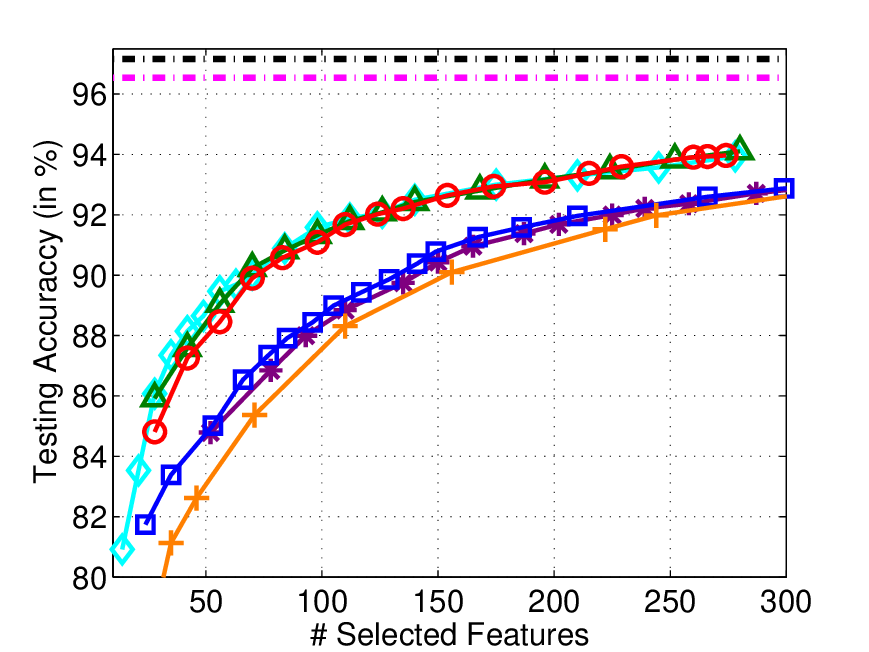}}
  \subfigure[{\tt rcv1}]{
    \includegraphics[trim = 1.5mm 0mm 6mm 2mm,  clip, height = 1.75in, width=2.35in]{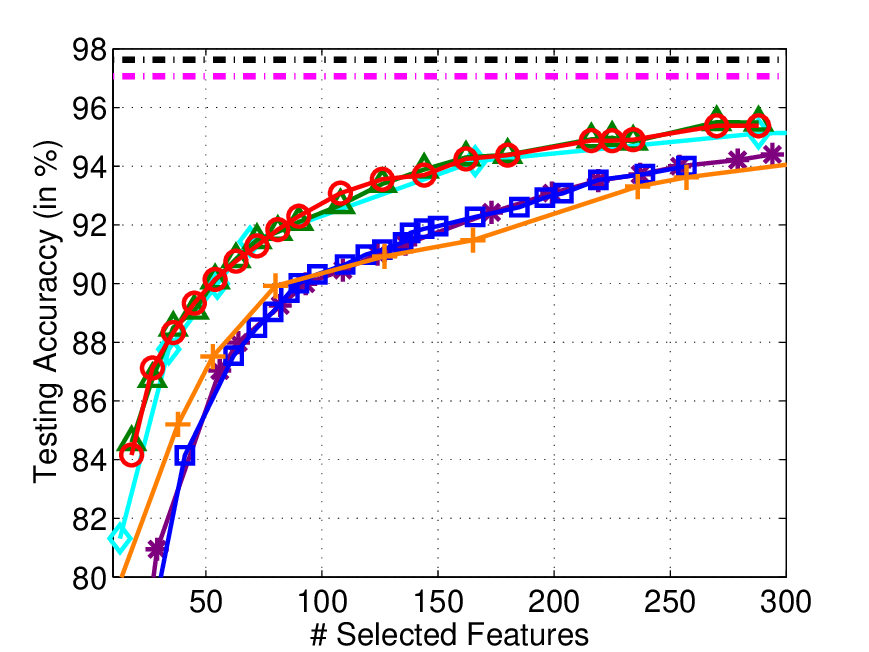}}
       \hspace{0.4in}
  \subfigure[{\tt news20}]{
    \includegraphics[trim = 1.5mm 0mm 6mm 2mm,  clip, height = 1.75in, width=2.35in]{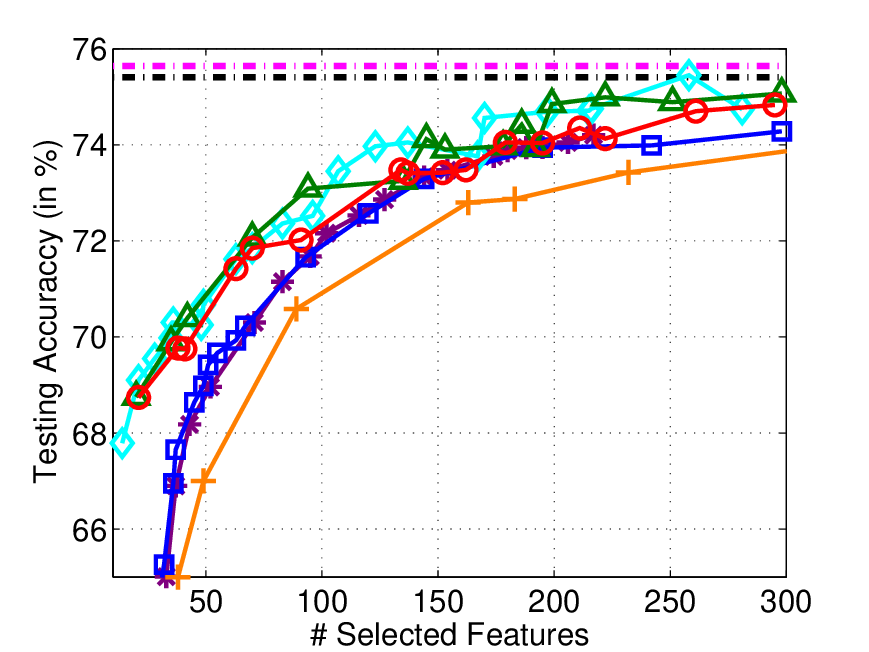}}
  \subfigure[{\tt astro-ph}]{
    \includegraphics[trim = 1.5mm 0mm 6mm 2mm,  clip, height = 1.75in, width=2.35in]{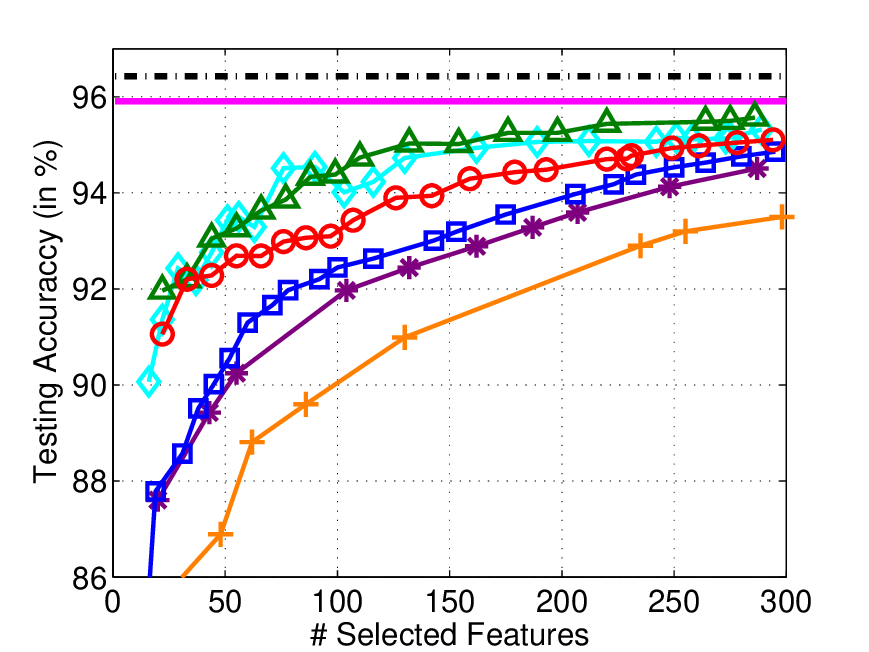}}
      \hspace{0.4in}
 \subfigure[{\tt kddb}]{
    \includegraphics[trim = 1.5mm 0mm 6mm 2mm,  clip, height = 1.75in, width=2.35in]{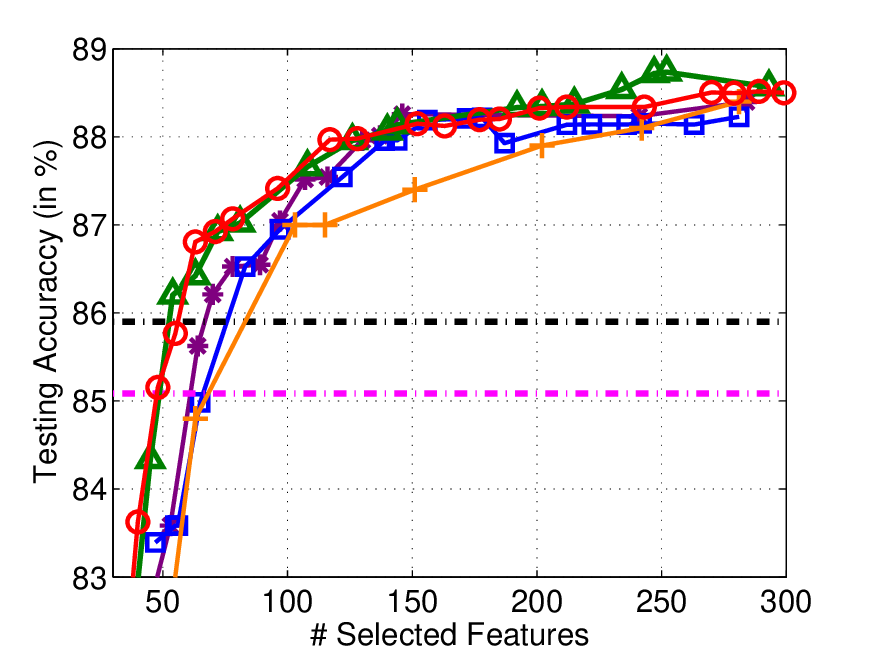}}
  \caption{Testing \aje{accuracies} on various datasets.}
  \label{fig:real_acc}
\end{figure*}

\begin{figure*}[h]
\center
    {\includegraphics[trim = 2.2cm 3.4cm 2.5cm 22.6cm,  clip,  width=0.98\textwidth]{legend_real_data.eps}}\vspace{-6mm} \\
  \centering
    \subfigure[{\tt epsilon}]{
    \label{fig:real_subfig_feat_time} 
    \includegraphics[trim = 1.5mm 0mm 6mm 2mm,  clip,  width=1.9in]{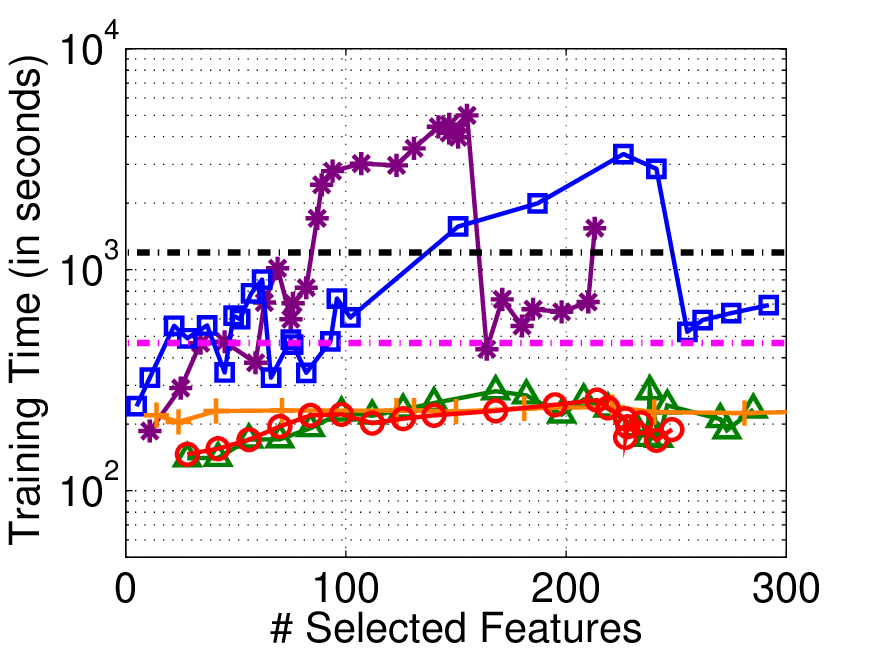}}
  \subfigure[{\tt real-sim}]{
    \includegraphics[trim = 1.5mm 0mm 6mm 2mm,  clip,  width=1.9in]{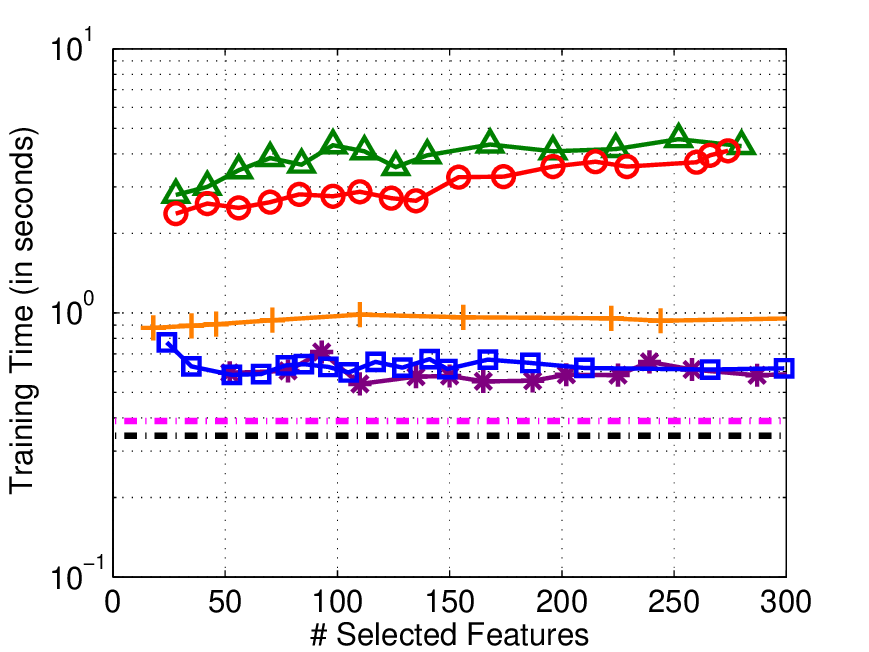}}
  \subfigure[{\tt rcv1}]{
    \includegraphics[trim = 1.5mm 0mm 6mm 2mm,  clip,  width=1.9in]{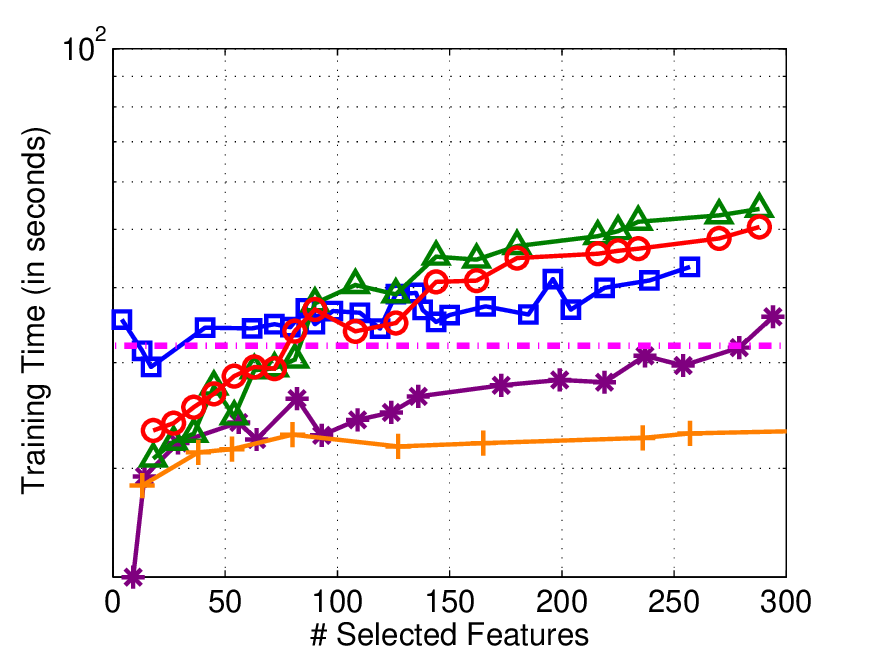}}
  \subfigure[{\tt news20}]{
    \includegraphics[trim = 1.5mm 0mm 6mm 2mm,  clip,  width=1.9in]{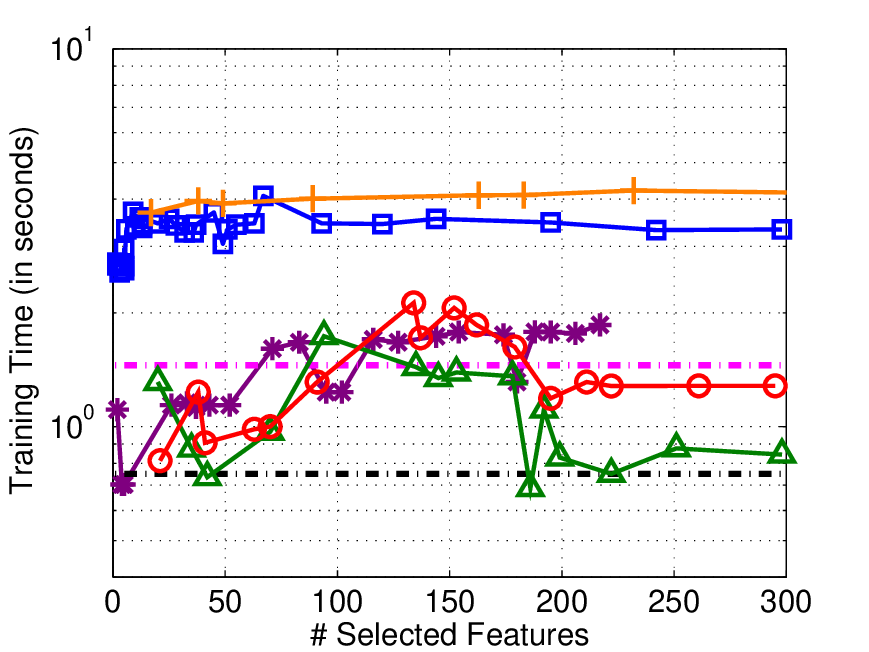}}
  \subfigure[{\tt physic}]{
    \includegraphics[trim = 1.5mm 0mm 6mm 2mm,  clip,  width=1.9in]{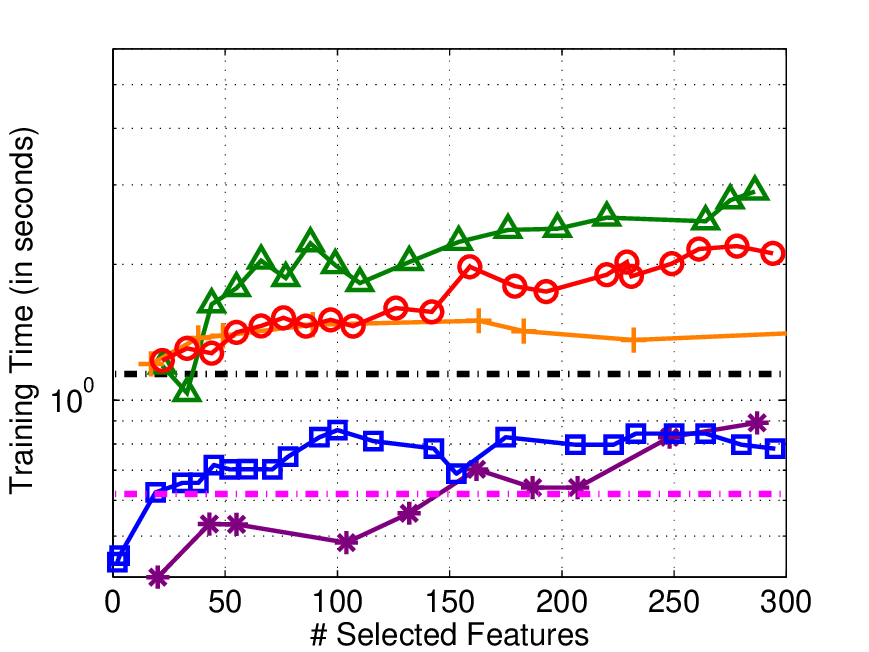}}
      \subfigure[{\tt kddb}]{
    \includegraphics[trim = 1.5mm 0mm 6mm 2mm,  clip,  width=1.9in]{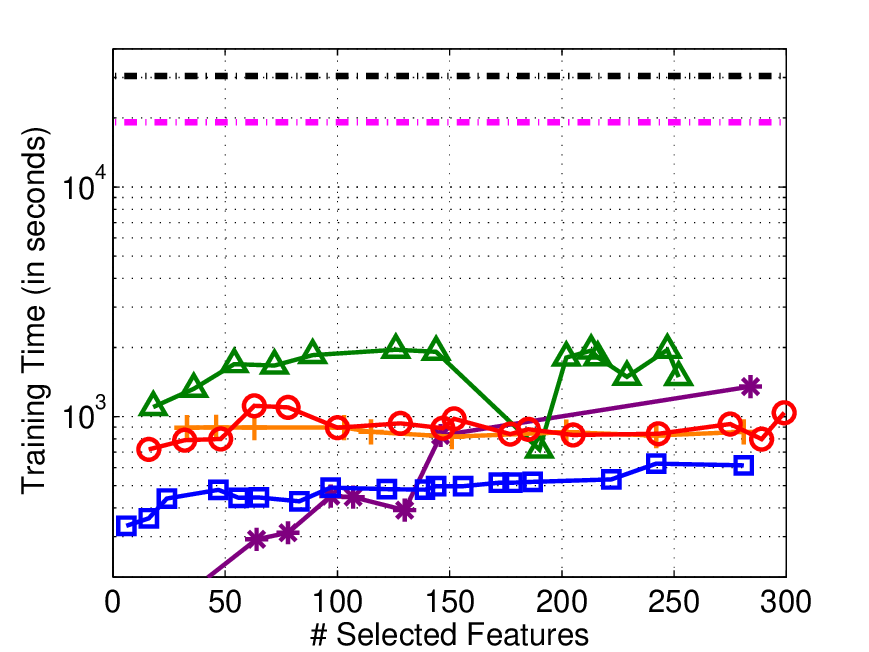}}
  \caption{Training \aje{times} on various datasets.}
  \label{fig:real_time}
\end{figure*}

The testing accuracy and training time \aje{for different methods versus}
the number of selected features are reported in Figure~\ref{fig:real_acc} and Figure~\ref{fig:real_time}, respectively.
From Figure~\ref{fig:real_acc}, on all datasets, {FGM} (including
{PROX-FGM}, {PROX-SLR} and {MKL-FGM}) obtains comparable or better
performance than the $\ell_1$-methods in terms of testing accuracy
within 300 features. Particularly, FGM shows \aje{a} much better testing
accuracy than $\ell_1$-methods on five of the studied datasets,
namely\aje{,} {\tt epsilon}, {\tt real-sim}, {\tt rcv1.binary}, {\tt Arxiv
astro-ph} and {\tt news20}.

From Figure \ref{fig:real_time}, {PROX-FGM} and {PROX-SLR} show
competitive training efficiency with the $\ell_1$-methods.
Particularly, on the large-scale dense {\tt epsilon} dataset,
{PROX-FGM} and {PROX-SLR} are much \aje{more} efficient than the LIBlinear
$\ell_1$-solvers. For {SGD-SLR}, although it demonstrates comparable
training efficiency \aje{as} {PROX-FGM} and {PROX-SLR}, it attains \aje{a} much
worse testing accuracy. In summary, {FGM}\aje{-}based methods in general
obtain better feature subsets with competitive training efficiency
\aje{as} the considered baselines on real-world datasets.

\subsubsection{De-biasing Effect of FGM}\label{real_exp_debias}

\begin{figure*}
  \centering
  \subfigure[{\tt epsilon}]{
    \label{fig:real_subfig_feat_unbias} 
    \includegraphics[trim = 1.5mm 0mm 6mm 2mm,  clip,   width=1.9in]{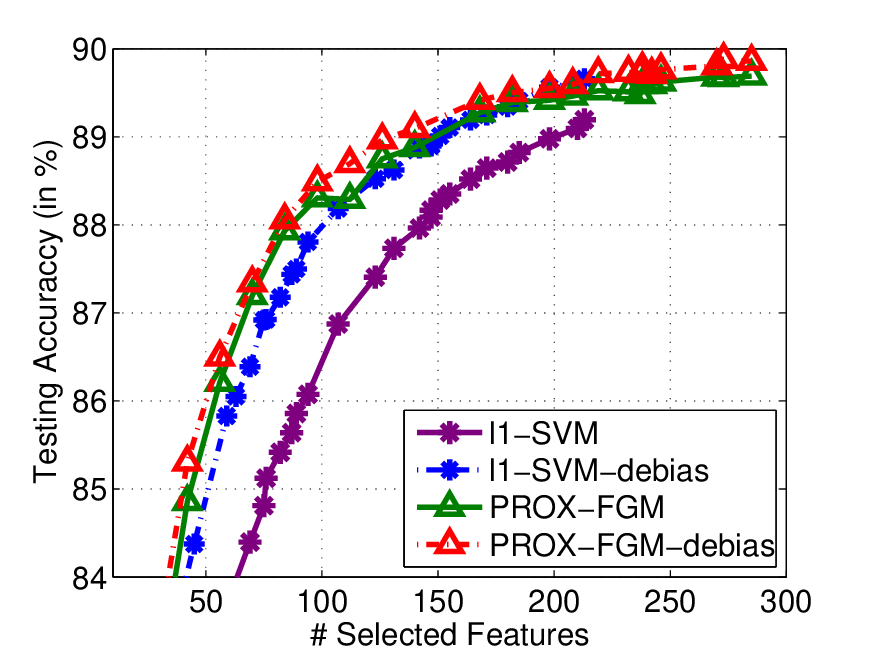}}
    \subfigure[{\tt real-sim}]{
    \includegraphics[trim = 1.5mm 0mm 6mm 2mm,  clip,   width=1.9in]{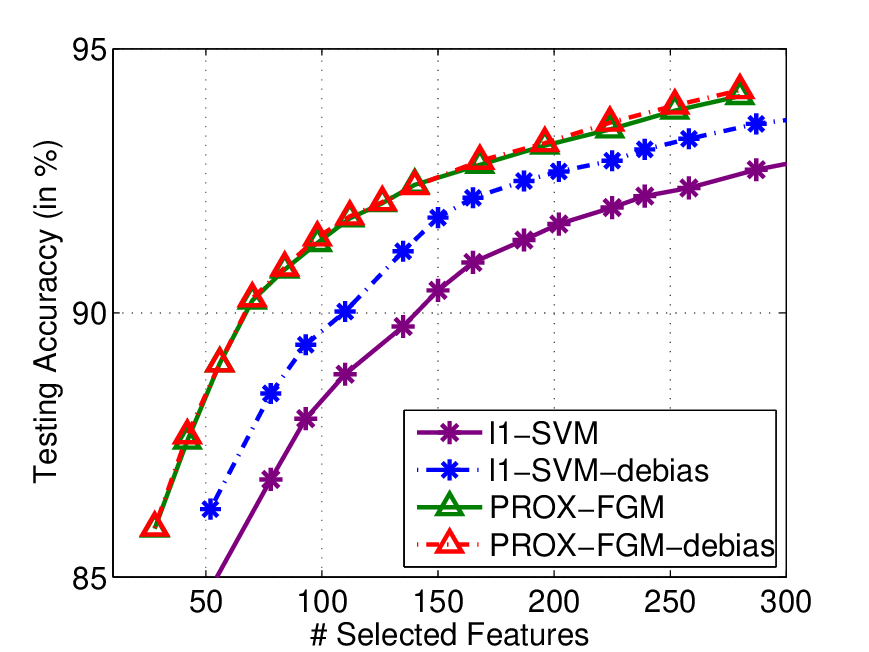}}
    \subfigure[{\tt rcv1}]{
    \includegraphics[trim = 1.5mm 0mm 6mm 2mm,  clip,   width=1.9in]{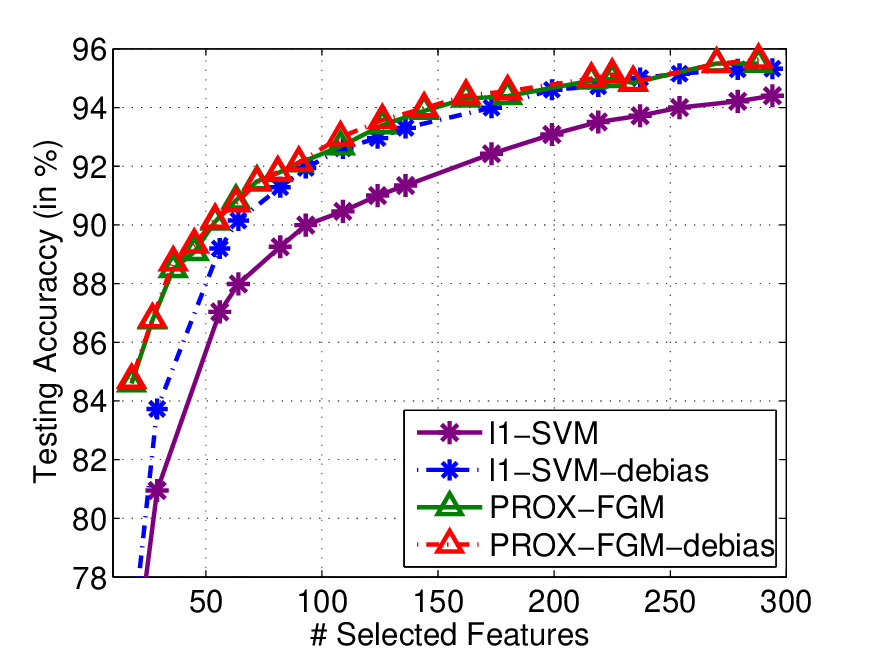}}
  \caption{De-biased results on real-world datasets.}
  \label{fig:real_acc_unbias}
\end{figure*}

In this experiment, we demonstrate the de-biasing effect of {FGM} on
three real-world datasets, namely\aje{,} {\tt epsilon}, {\tt real-sim} and
{\tt rcv1}. Here, only the squared hinge loss (namely\aje{,} PFOX-FGM) is
studied. The de-biased results are reported in Figure~\ref{fig:real_acc_unbias}, where {PROX-FGM-debias} and
{l1-SVM-debias} denote the de-biased results of {PROX-FGM} and
l1-SVM, respectively.

From Figure \ref{fig:real_acc_unbias}, {l1-SVM-debias} shows much
better results than {l1-SVM} \aje{indicating that a} feature selection
bias issue exists in {l1-SVM} on these real-world datasets. \aje{In contrast}, {PROX-FGM} achieves close or even better results compared
with its de-biased counterparts, which verifies that {PROX-FGM}
itself can reduce the feature selection bias. Moreover, on these
datasets, {FGM} shows better testing accuracy than the de-biased
l1-SVM, namely\aje{,} {l1-SVM-debias}, which indicates that the features
selected by {FGM} are more relevant than those obtained by l1-SVM
due to the reduction of feature selection bias.

\subsubsection{Sensitivity Study of Parameters}\label{exp_real_para}
In this section, we conduct \aje{a} sensitivity study of parameters for
PROX-FGM. There are two parameters in FGM, namely\aje{,} the sparsity
parameter $B$ and the regularization parameter $C$.  In \aje{these}
experiments, we study the sensitivity of these two parameters on
{\tt real-sim} and {\tt astro-ph} datasets. l1-SVM is adopted as the
baseline.

\begin{figure*}[htp]
  \centering
  \subfigure[{\tt real-sim}]{
    \label{fig:sentivity_acc} 
    \includegraphics[trim = 1.5mm 0mm 6mm 2mm,  clip, width=2.35in]{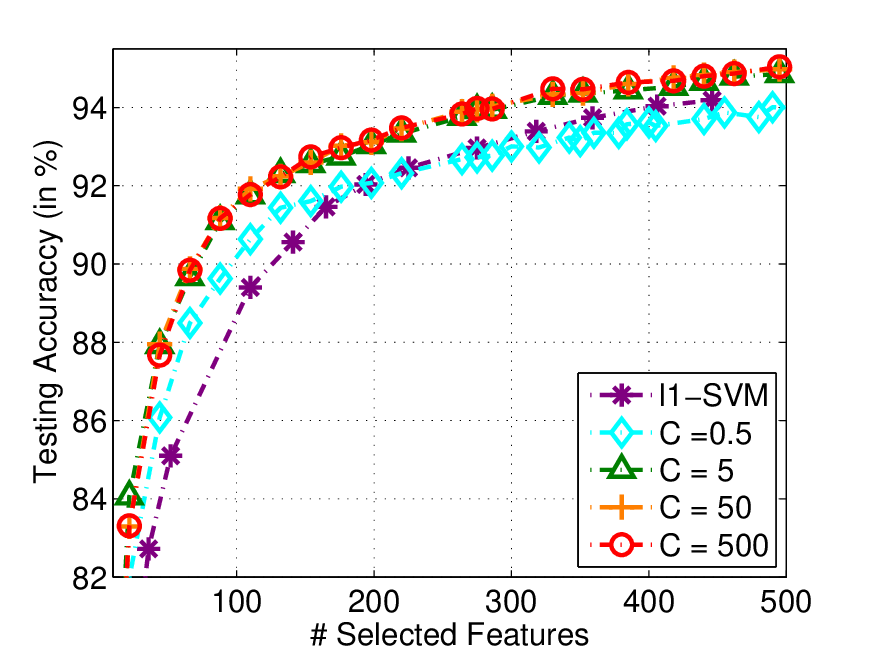}}
\hspace{0.1in}
     \subfigure[{\tt astro-ph}]{
    \label{fig:sentivity_time} 
    \includegraphics[trim = 1.5mm 0mm 6mm 2mm,  clip,   width=2.35in]{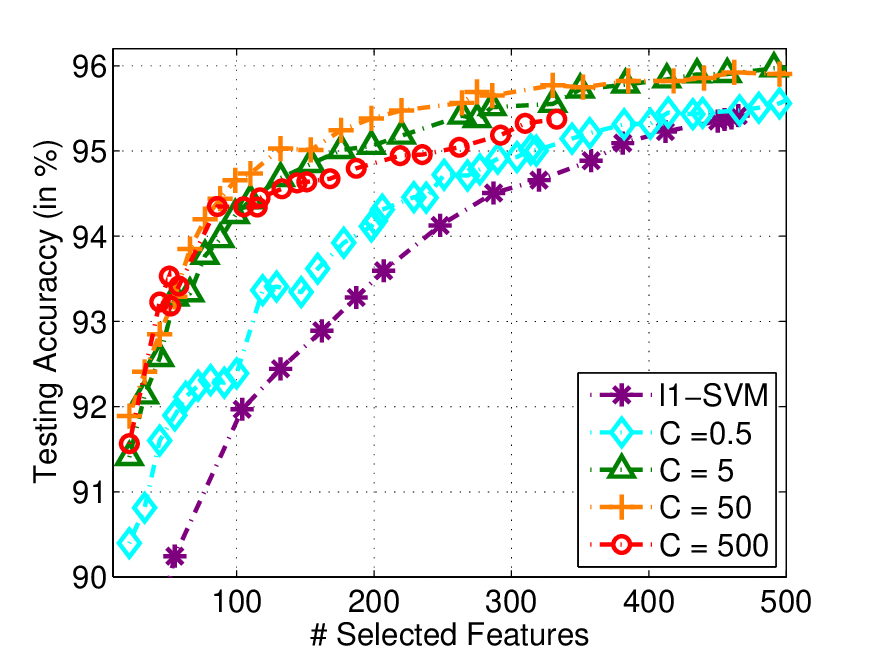}}
 \caption{Sensitivity of the parameter $C$ for FGM
on {\tt real-sim} and {\tt astro-ph} datasets. }
  \label{fig:senitiviy_C}
\end{figure*}

In the first experiment, we study the sensitivity of $C$. FGM with
suitable $C$ can reduce the feature selection bias. However, \aje{if} $C$ is
too large, \aje{overfitting} may \aje{occur}. To demonstrate this,
we test $C \in \{0.5, 5, 50, 500\}$. The testing accuracy of FGM
under different $C$'s is reported in Figure~\ref{fig:senitiviy_C}.
From Figure~\ref{fig:senitiviy_C}, the testing accuracy with \aje{a small}
$C$ \aje{is worse in general} than that with a large $C$. The reason \aje{for this} is that, when $C$ is small, feature selection bias may \aje{occur due to
an underfitting} problem. However, when $C$ is \aje{sufficiently} large,
increasing $C$ may not necessarily improve the performance. More
critically, if $C$ is too large, \aje{an overfitting problem may
occur}. For example, on the {\tt astro-ph} dataset, FGM with $C =
500$ performs much worse \aje{in general} than FGM with $C = 5$ and $C =
50$. An important observation is that, on both datasets, FGM with
different $C$'s generally performs better than the l1-SVM.

Recall that a large $C$ may lead to slower convergence \aje{speeds due to
an increase in the} Lipschitz constant of $F(\ww,b)$. In practice,
we suggest choosing $C$ in the range of $[1, 100]$. In Section
\ref{sec:real-world}, we have set $C = 10$ for all datasets. \aje{In}
this setting, FGM has demonstrated superb performance over the
competing methods. \aje{In contrast}, choosing the regularization
parameter for $\ell_1$-methods is more difficult. Therefore, FGM is
convenient \aje{for performing model selection}.

\begin{figure*}[htp]
  \centering
  \subfigure[$\#$ Selected Features Versus $B$]{
    \label{fig:senitiviy_feat_add} 
    \includegraphics[trim = 1.5mm 0mm 6mm 2mm,  clip,  width=2.35in]{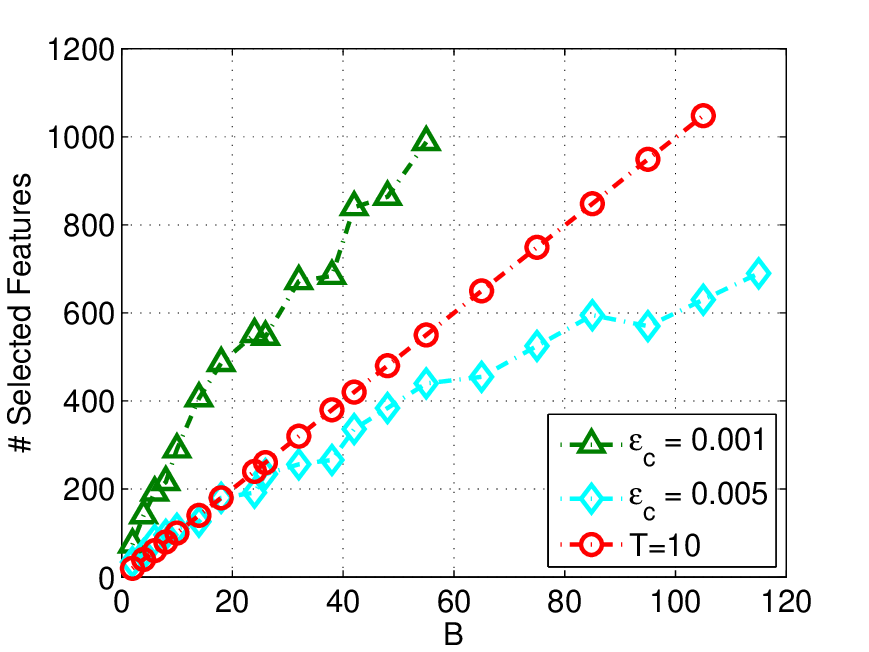}}
    \hspace{0.1in}
\subfigure[Testing Accuracy Versus $B$]{
    \label{fig:senitiviy_acc_add} 
    \includegraphics[trim = 1.5mm 0mm 6mm 2mm,  clip,  width=2.35in]{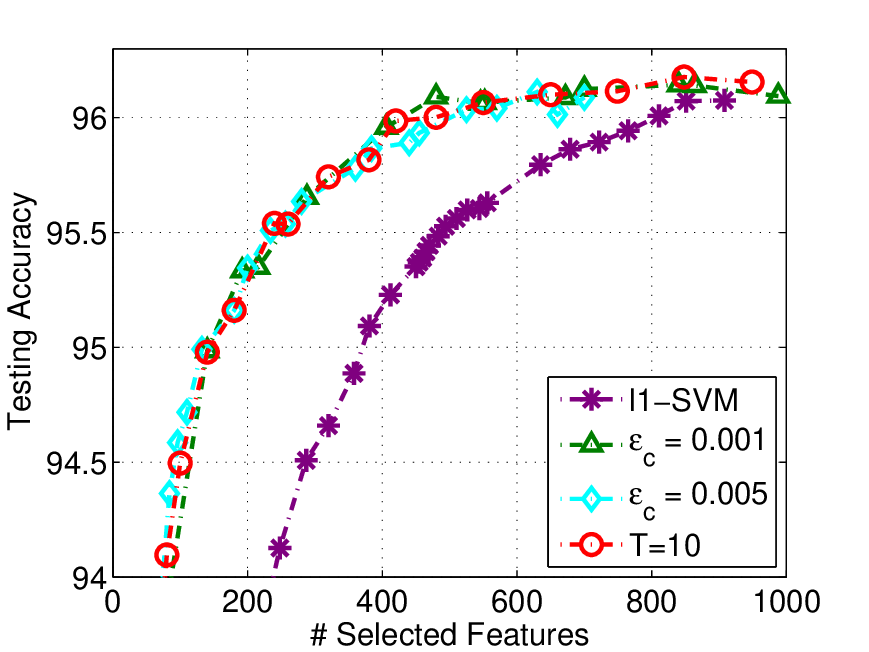}}
     \subfigure[Training Time (in seconds)]{
    \label{fig:senitiviy_time_add} 
    \includegraphics[trim = 1.5mm 0mm 6mm 2mm,  clip,  width=2.35in]{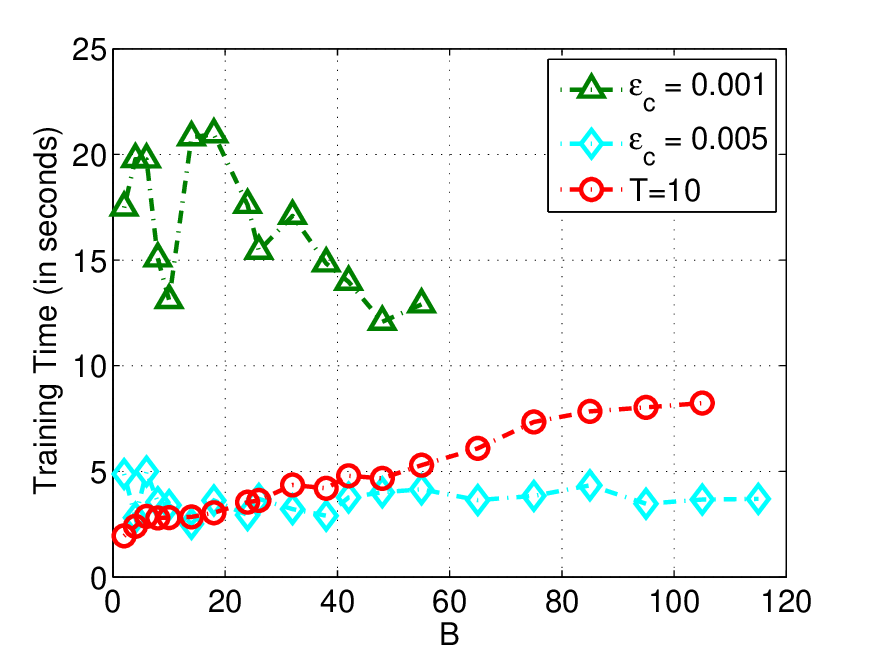}}
 \caption{Sensitivity of the parameter $B$ for FGM
on {\tt astro-ph} dataset, where FGM is stopped once
$(F(\ww_{t-1},b) - F(\ww_{t},b))/ F(\ww_{0},b) \leq \epsilon_{c}$.}
  \label{fig:senitiviy_B_add}
\vskip -0.2in
\end{figure*}

In the second experiment, we study the sensitivity of parameter $B$
for FGM under two stopping conditions:  (1) \aje{when} the condition
$(F(\ww_{t-1},b) - F(\ww_{t},b))/ F(\ww_{0},b) \leq \epsilon_{c}$ is
achieved \aje{and} (2) \aje{when} a maximum $T$ iterations is achieved where $T=10$. Here, we test two values of $\epsilon_{c}$, namely\aje{,} $\epsilon_{c} =
0.005$ and $\epsilon_{c} = 0.001$. The number of selected features,
the testing accuracy and the training time versus different $B$ \aje{values} are
reported in Figure~\ref{fig:senitiviy_feat_add},
\ref{fig:senitiviy_acc_add} and \ref{fig:senitiviy_time_add},
respectively.

In Figure \ref{fig:senitiviy_feat_add}, given the number of selected
feature $\#~ \text{features}$, the number of required iterations is
\aje{approximately} $\lceil\frac{\#~ \text{features}}{B}\rceil$ under the first
stopping criterion. In this sense, FGM with $\epsilon_{c}=0.001$
takes more than 10 iterations to terminate \aje{and} thus will choose more
features. As a result, it needs more time for optimization with
the same $B$, as shown in Figure~\ref{fig:senitiviy_time_add}. \aje{In contrast}, FGM with $\epsilon_{c} = 0.005$ requires fewer iterations (smaller than 10 when $B > 20$). Surprisingly, as
shown in Figure~\ref{fig:senitiviy_acc_add}, FGM with fewer
iterations (where $\epsilon_{c} = 0.005$ or $T=10$) \aje{obtains a} similar
testing accuracy \aje{as} FGM with $\epsilon_{c} = 0.001$, but shows
much better training efficiency. This observation indicates that we
can set a small number outer iterations (for example $5\leq T \leq
20$) to trade-off the training efficiency and the feature selection
performance.

\begin{figure*}[htp]
  \centering
   \subfigure[Training Time (in seconds) Versus $B$ ]{
    \label{fig:senitiviy_time} 
    \includegraphics[trim = 1.5mm 0mm 6mm 2mm,  clip, height = 1.75in,  width=2.35in]{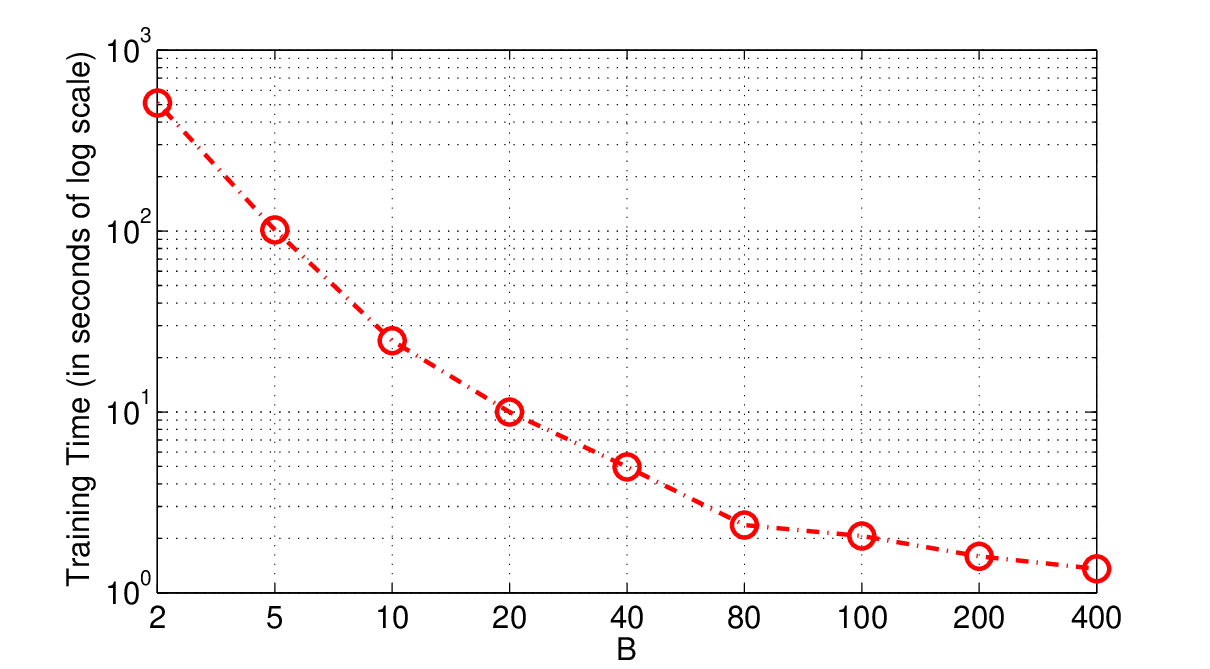}}
\hspace{0.1in}
  \subfigure[Testing Accuracy Versus $B$]{
    \label{fig:senitiviy_acc} 
    \includegraphics[trim = 1.5mm 0mm 6mm 2mm,  clip, height = 1.75in,  width=2.35in]{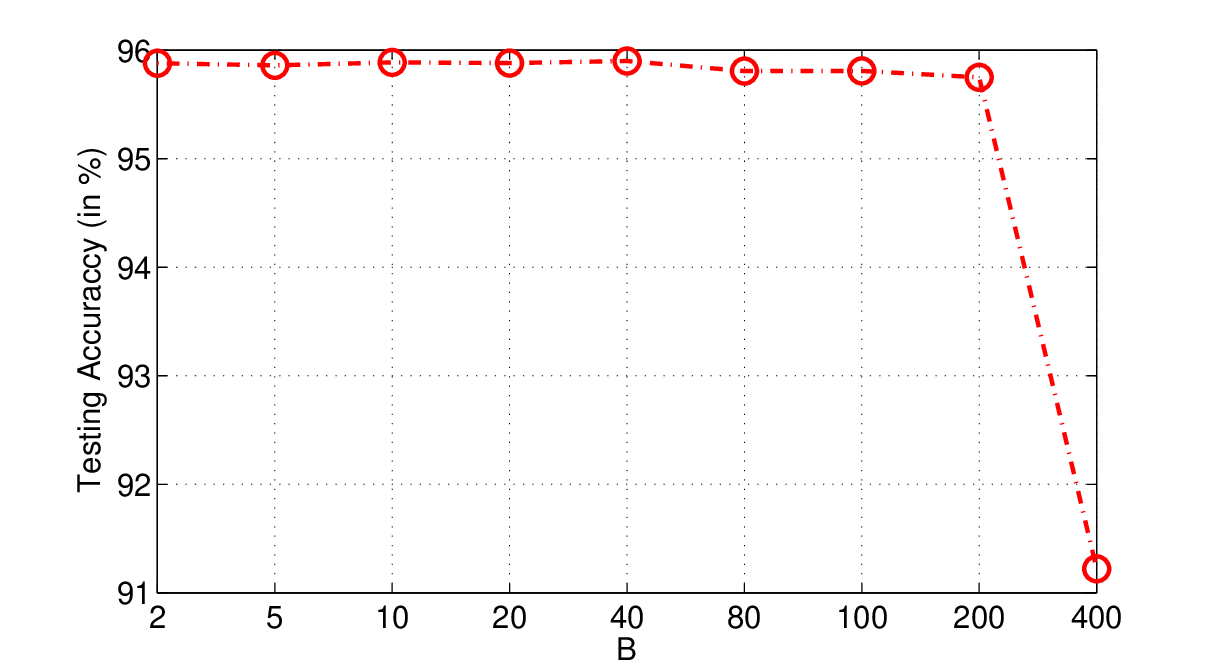}}
     \caption{Sensitivity of the parameter $B$ for FGM
on {\tt astro-ph} dataset. Given a parameter $B$, we stop FGM once
400 features are selected. }
  \label{fig:senitiviy_B}\vskip -0.3in
\end{figure*}

In the third experiment, we  study the influence of the parameter
$B$ on the performance of FGM on the {\tt astro-ph} dataset.
Specifically, we stop FGM once 400 features are selected with
different $B$'s. The training time and testing accuracy w.r.t.
different $B$'s are shown in Figure~\ref{fig:senitiviy_time} and
\ref{fig:senitiviy_acc}, respectively.

From Figure \ref{fig:senitiviy_time}, choosing a large $B$ in
general leads to better training efficiency. Particularly, FGM with
$B=40$ is \aje{approximately} 200 times faster than FGM with $B = 2$. Recall that active-set methods in~\citep{Roth2008,Francis2009HKL} can be
considered as special cases of FGM with $B=1$. Accordingly, we can
conclude that FGM with a properly selected $B$ can be much faster
than active-set methods~\cite{Roth2008,Francis2009HKL}. However, it
should be pointed that,  if $B$ is too large, the performance may
degrade. For instance, if we choose $B = 400$, the testing accuracy
dramatically degrades, which indicates that the selected 400
features are not the optimal ones. In summary, choosing a suitable
$B$ (e.g.\aje{,} $B\leq 100$) can much improve the efficiency while
maintaining promising generalization performance.

\subsection{Ultrahigh Dimensional  Feature
Selection via Nonlinear Feature Mapping}\label{sec:subexpploy} 

In this experiment, we compare the efficiency of {FGM}  and
$\ell_1$-SVM  on nonlinear feature \aje{selection} with polynomial
feature mappings on two medium dimensional datasets and a high
dimensional dataset. The comparison methods are denoted by
{PROX-PFGM}, {PROX-PSLR} and {l1-PSVM}.\footnote{The
\aje{code for} l1-PSVM \aje{is} available at:
http://www.csie.ntu.edu.tw/$\sim$cjlin/libsvmtools/$\#$fast$\_$training$\_$
testing$\_$for$\_$degree$\_$2$\_$polynomial$\_$mappings$\_$of$\_$data.}
The details of the studied datasets are shown in Table~\ref{polydataset}, where $m_{Poly}$ denotes the dimension of the
polynomial mappings and $\gamma$ is the polynomial kernel parameter
used in this experiment. The {\tt mnist38}
 dataset consists of the digital images of $3$ and $8$ from the
{\tt mnist}
dataset.\footnote{http://www.csie.ntu.edu.tw/~cjlin/libsvmtools/datasets/.}
For the {\tt kddb} dataset, we only use the first $10^6$ instances.
\aje{Last}, we change the parameter $C$ for {l1-PSVM} to obtain
different \aje{numbers} of features.

\begin{table}[t]
\vspace{-0.02in}
\begin{center}
\begin{small}
\begin{tabular}{|c|c|c|c|c|c|}
\hline
   Dataset &  $m$ & $m_{Poly}$ & $n_{train}$ & $n_{test}$ &  $\gamma$  \\
 \hline
{\tt mnist38} &     784 &$O(10^5)$&     40,000 &     22,581& 4.0\\
 \hline
 {\tt real-sim} &     20,958 & $O(10^8)$&     32,309 &     40,000 & 8.0\\
\hline
{\tt kddb} &  4,590,807  &    $O(10^{14})$ & $1000,000$ &     748,401 & 4.0\\
\hline
\end{tabular}
\caption{Details of the datasets using polynomial feature mappings.}
\label{polydataset} 
\end{small}
\end{center}
\vskip -0.1in
\end{table}

\begin{figure*}[htp]
  \centering
    \subfigure[{\tt mnist38} ]{
    \label{fig:poly_mnist38_time} 
    \includegraphics[trim = 1.5mm 0mm 6mm 2mm,  clip,  width=1.90in]{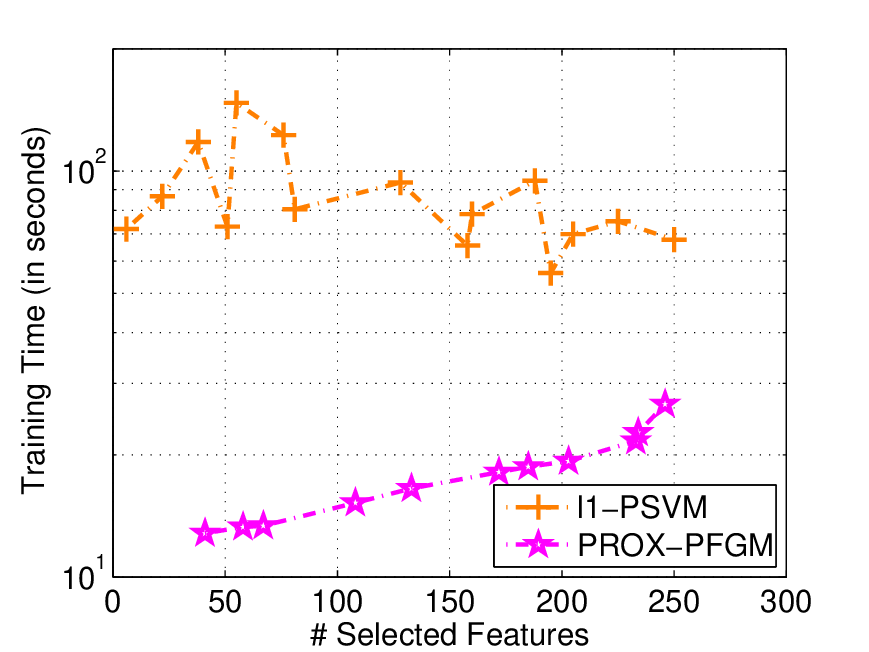}}
    \subfigure[{\tt real-sim}]{
    \label{fig:poly_real_sim_time}
    \includegraphics[trim = 1.5mm 0mm 6mm 2mm,  clip,  width=1.90in]{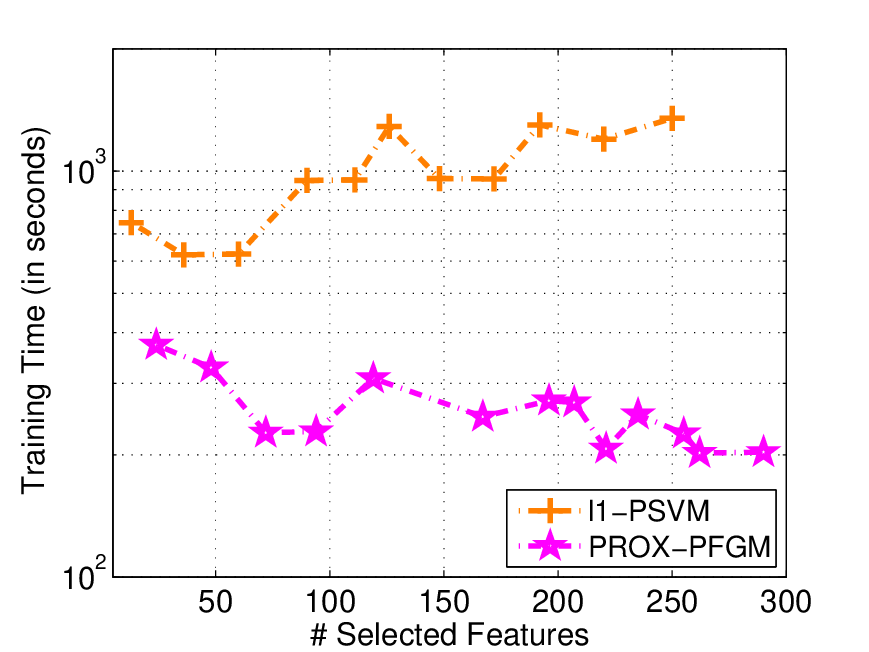}}
 \subfigure[{\tt kddb} ]{
   \label{fig:poly_kddb_time}
   \includegraphics[trim = 1.5mm 0mm 6mm 2mm,  clip,   width=1.90in]{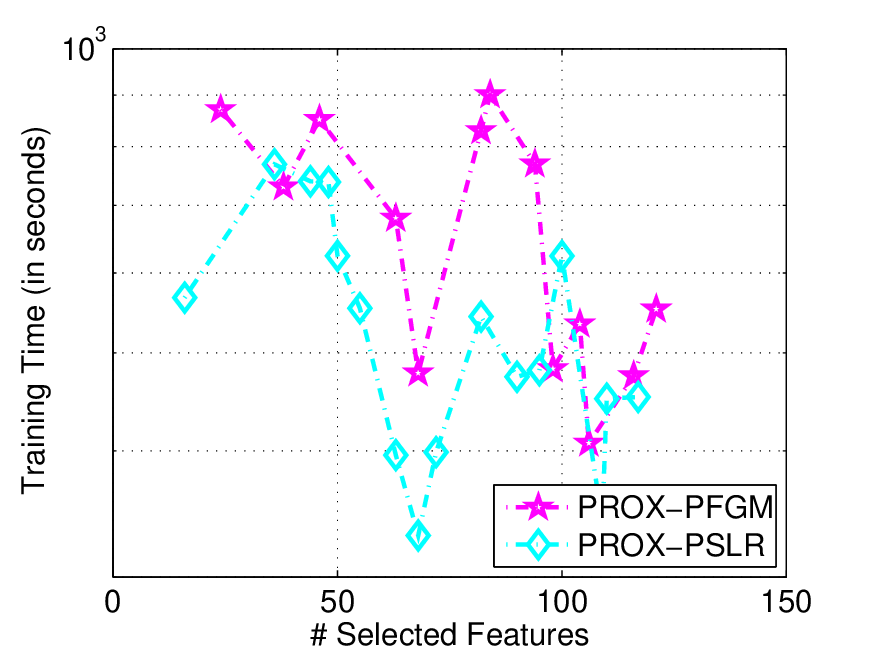}}

  \caption{Training \aje{times} of different methods on nonlinear feature selection using polynomial mappings. }
    \label{fig:poly_feature_time}\vskip -0.0in
\end{figure*}

\begin{figure*}[htp]
  \centering
  \subfigure[{\tt mnist38} ]{
    \label{fig:poly_subfig_feat_spar_mnist38} 
    \includegraphics[trim = 1.5mm 0mm 3mm 0mm,  clip,   width=1.90in]{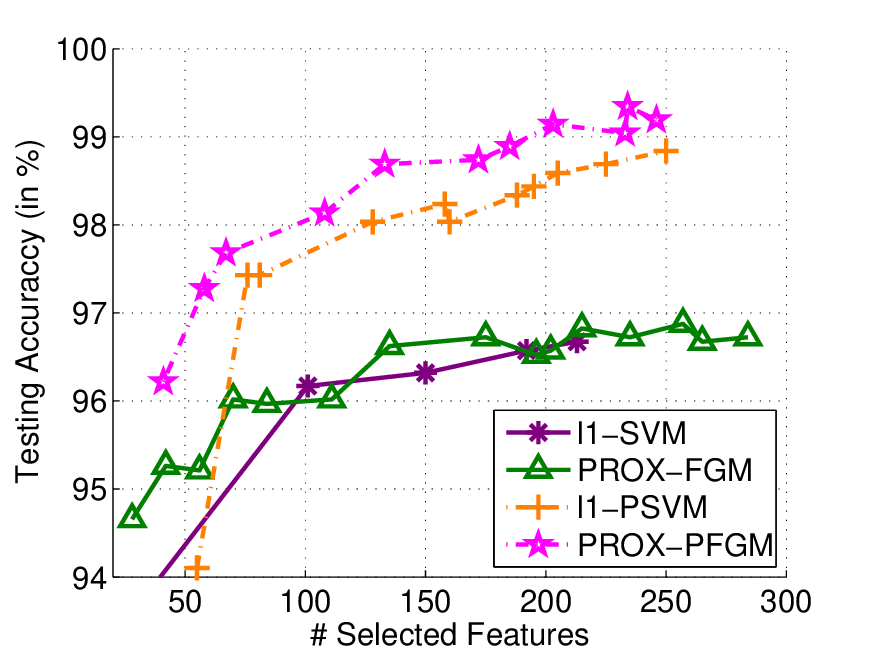}}
      \subfigure[ {\tt real-sim}]{
    \label{fig:poly_subfig_feat_spar_real} 
    \includegraphics[trim = 1.5mm 0mm 3mm 0mm,  clip,   width=1.90in]{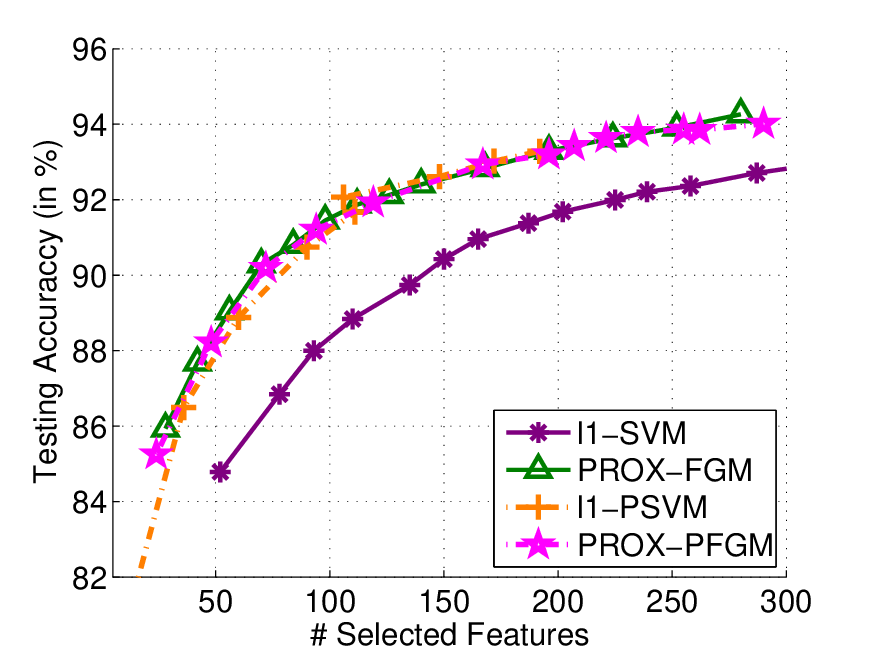}}
 \subfigure[{\tt kddb} ]{
    \label{fig:poly_subfig_feat_spar_kddb} 
    \includegraphics[trim = 1.5mm 0mm 6mm 2mm,  clip,  width=1.90in]{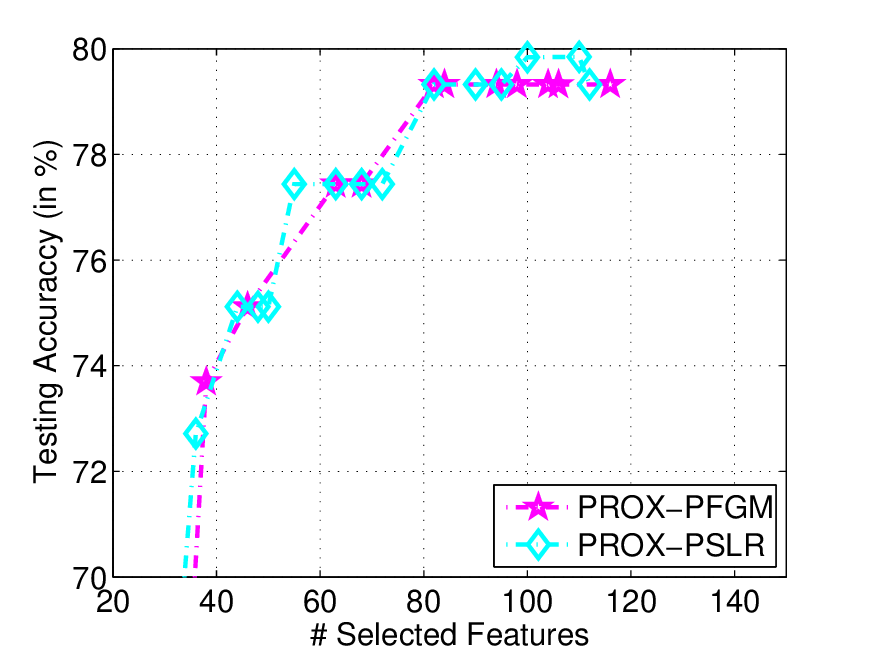}}
  \caption{Testing \aje{accuracies} of different methods on nonlinear feature selection using polynomial mappings. }
  \label{fig:poly_feature_acc}
\end{figure*}

The training time and testing accuracy  on different datasets are
reported in Figure~\ref{fig:poly_feature_time} and
\ref{fig:poly_feature_acc}, respectively. Both {PROX-PFGM} and
{l1-PSVM} can address the two medium dimensional problems. However,
{PROX-PFGM} shows much better efficiency than {l1-PSVM}. Moreover,
{l1-PSVM} is \aje{unfeasible} on the {\tt kddb} dataset due to the
ultrahigh dimensionality. Particularly, {l1-PSVM} needs more
than 1 TB \aje{of} memory to store a dense $\w$, which is \aje{unfeasible} for a
common PC. Conversely, this difficulty can be effectively addressed
by FGM. Specifically, {PROX-PFGM} completes the training within 1000
seconds.

From the figures, the testing accuracy on the {\tt mnist38} dataset
with polynomial mapping is much better than that of linear methods,
which \aje{demonstrates} the usefulness of the nonlinear feature
expansions. On the {\tt real-sim} and {\tt kddb} datasets however,
the performance with polynomial mapping does not show significant
improvements. A possible reason is that these two datasets are
linearly separable.

\subsection{Experiments for Group Feature Selection}
\label{sec:exp2}

In this section, we study the performance of the proposed methods
for group feature selection on a synthetic dataset and two
real-world datasets. Here, only the logistic loss is studied since it
has been widely used for group feature \aje{selection} on classification
tasks~\citep{Roth2008,Liu2010}. To demonstrate the sensitivity of
the parameter $C$ to {FGM}, we vary $C$ to select different \aje{numbers}
of groups under the stopping tolerance $\epsilon_c = 0.001$. For
each $C$, we test  $B \in \{2, 5, 8, 10\}$.  The tradeoff parameter
$\lambda$ in {\textsf{SLEP}} is chosen from $[0 ,1]$, where a larger
lambda leads to more sparse solutions~\citep{Liu2010}. Specifically,
we set $\lambda$ in $[0.002, 0.700]$ for {FISTA} and {{ACTIVE}} and
set $\lambda$ in $[0.003, 0.1]$ for {BCD}.

\subsubsection{Synthetic Experiments on Group Feature Selection}
In the synthetic experiment, we generate a random matrix
${\bX}\in\R^{4,096\times 400,000}$\aje{, where} each entry follows the
i.i.d. Gaussian distribution $\mathcal N (0,1)$. After, we directly
group the 400,000 features into 40,000 groups of equal
size~\citep{Jenatton2011}, namely\aje{,} each feature group contains 10
features. We randomly choose $100$ groups as the
ground-truth informative groups. Specifically, we generate a sparse
vector $\w$ where only the entries of the selected groups are
nonzero values sampled from the i.i.d. Gaussian distribution
$\mathcal N (0,1)$. \aje{Last}, we produce the output labels by $\y =
\text{sign}({\bX}\w)$. We generate 2,000 testing points in the same
manner.

\begin{figure*}[htp]
\center
    {\includegraphics[trim = 10mm 5.5cm 10mm 2.0cm,  clip,  width=0.74\textwidth]{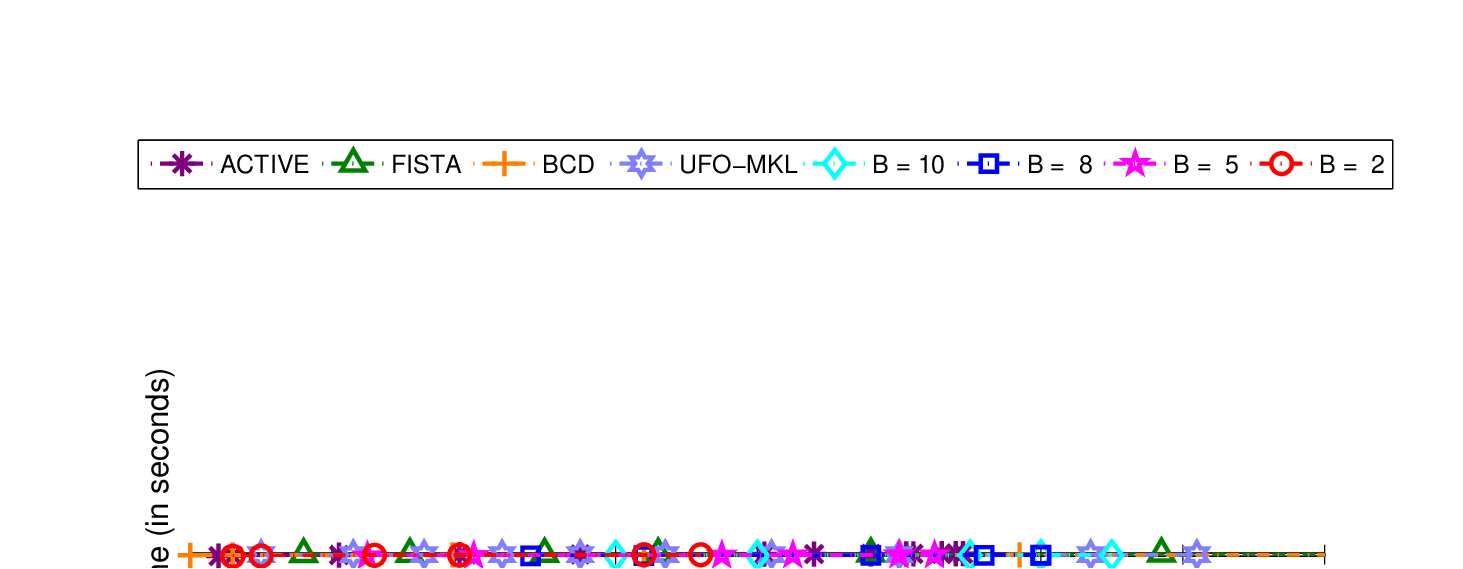}}\vspace{-6mm} \\
  \centering
  \subfigure[Testing accuracy]{
  \label{fig:subfig_group_acc} 
    \includegraphics[trim = 1.5mm 0mm 6mm 2mm,  clip,  width=1.9in]{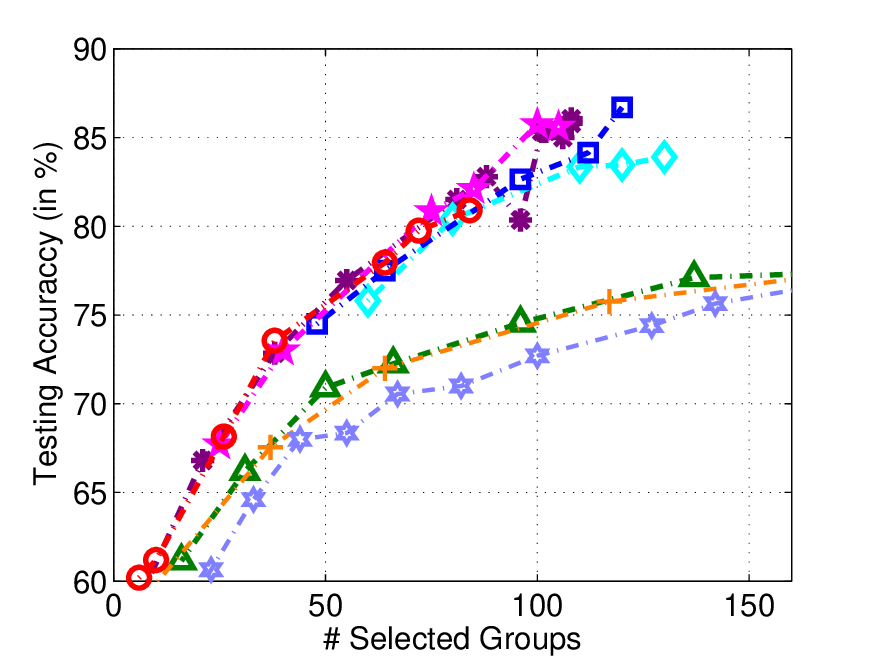}}
    \subfigure[Recovered features]{
  \label{fig:subfig_group_num}
    \includegraphics[trim = 1.5mm 0mm 6mm 2mm,  clip,  width=1.9in]{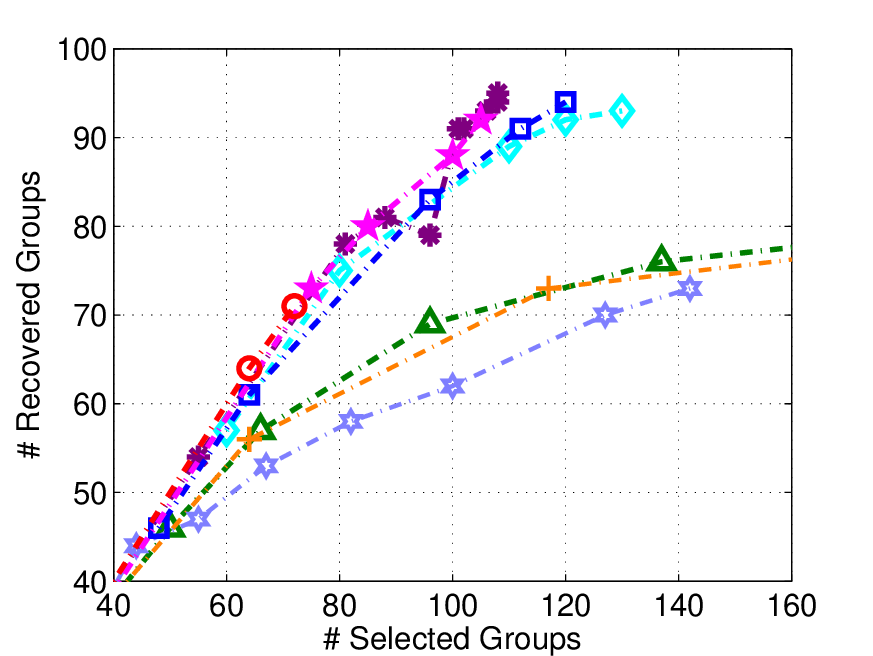}}
\subfigure[Training time]{
 \label{fig:subfig_group_time} 
    \includegraphics[trim = 1.5mm 0mm 6mm 2mm,  clip,  width=1.9in]{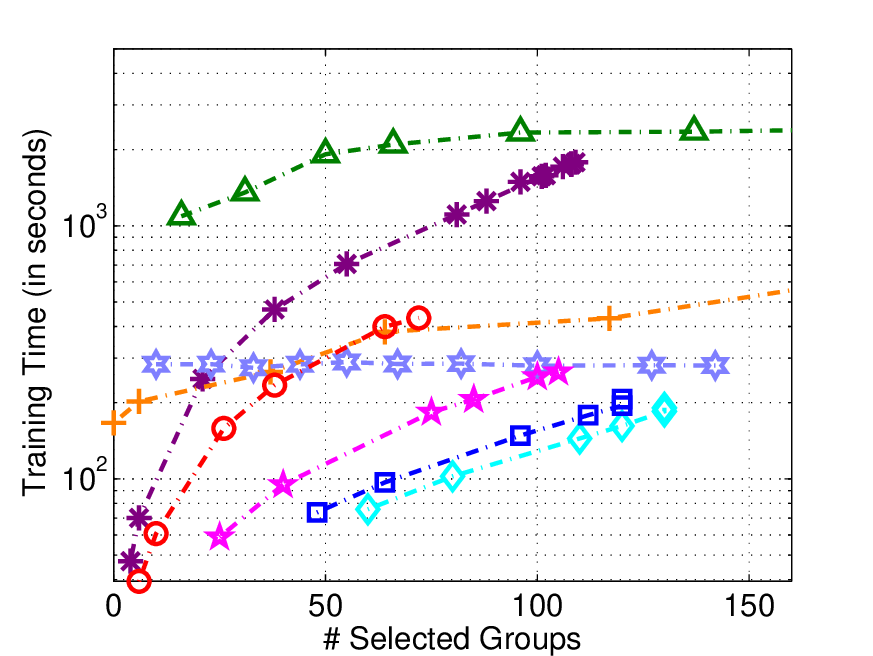}}
  \caption{Results of group feature selection on the synthetic dataset.}
  \label{toy_group_large}
\end{figure*}

 The testing accuracy,
training time and number of recovered ground-truth groups are
reported in Figure~\ref{fig:subfig_group_acc},
\ref{fig:subfig_group_num} and \ref{fig:subfig_group_time},
respectively. {Here\aje{,} only the results within 150 groups are included
since we only have  100 informative ground-truth groups.} From
Figure~\ref{fig:subfig_group_acc},  {FGM} achieves better testing
accuracy than {FISTA}, {BCD} and UFO-MKL. The reason \aje{for this} is that FGM
can reduce the group feature selection bias. From
Figure~\ref{fig:subfig_group_time}, in general, {FGM} is much more
efficient than {FISTA} and {BCD}. Interestingly, the active-set
method (ACTIVE) also shows good testing accuracy compared with
{FISTA} and {BCD}. However, from Figure~\ref{fig:subfig_group_time},
its efficiency is limited since it only includes one element per
iteration. Accordingly, when selecting a large number of groups on
\emph{Big Data}, \aje{the} computational cost becomes unbearable. For
UFO-MKL, although its training speed is fast, its testing accuracy
is generally worse than others. \aje{Last}, with a fixed $B$ for {FGM},
the number of selected groups will increase when $C$ becomes large.
This is because, with a larger $C$, one \aje{places} more importance on
the training errors. Accordingly, more groups are required to
achieve lower empirical errors.

\begin{table*}[htp]
\vspace{-0.02in}
\begin{center}
\begin{small}
\begin{tabular}{|c|c|c|c|c|c|c|c|c|c|c|}
\hline \multirow{2}{*}
{Dataset}&\multirow{2}{*}{$m$}&\multirow{2}{*}{$n_{train}$}&\multicolumn{3}{c|}{Size
of training set (GB)}
&\multirow{2}{*}{$n_{test}$}&\multicolumn{4}{c|}{Size of testing set(GB)}\\
\cline{4-6}\cline{8-10}
& && Linear&{ADD}&HIK&&Linear&{ADD}&HIK\\
\hline\hline
{ aut} &20,707&40, 000&0.027&{{0.320}}&{0.408}&{22,581}&0.016&{0.191}&{0.269}\\
\hline
{ rcv1} &47,236&677,399&0.727&{{8.29}}&{9.700}&{20,242}&0.022&{0.256}&{0.455}\\
\hline
\end{tabular}
\caption{Details of the datasets used for HIK kernel feature
expansion and Additive kernel feature expansion. For HIK kernel
feature expansion, each original feature is represented by a group
of 100 features\aje{,} while for Additive kernel feature expansion, each
original feature is represented by a group of 11 features.
}\label{group_generate}
\end{small}
\end{center}
\vskip -0.1in
\end{table*}

\subsubsection{Experiments on Real-World Datasets}
\begin{figure*}[h]
\center
    {\includegraphics[trim = 10mm 5.5cm 10mm 2.0cm,  clip,  width=0.74\textwidth]{group_large_legend.eps}}\vspace{-6mm} \\
  \centering
  \subfigure[{\tt aut-ADD}]{
    \label{fig:aut_real_group_acc} 
    \includegraphics[trim = 1.5mm 0mm 6mm 2mm,  clip,  width=2.35in ]{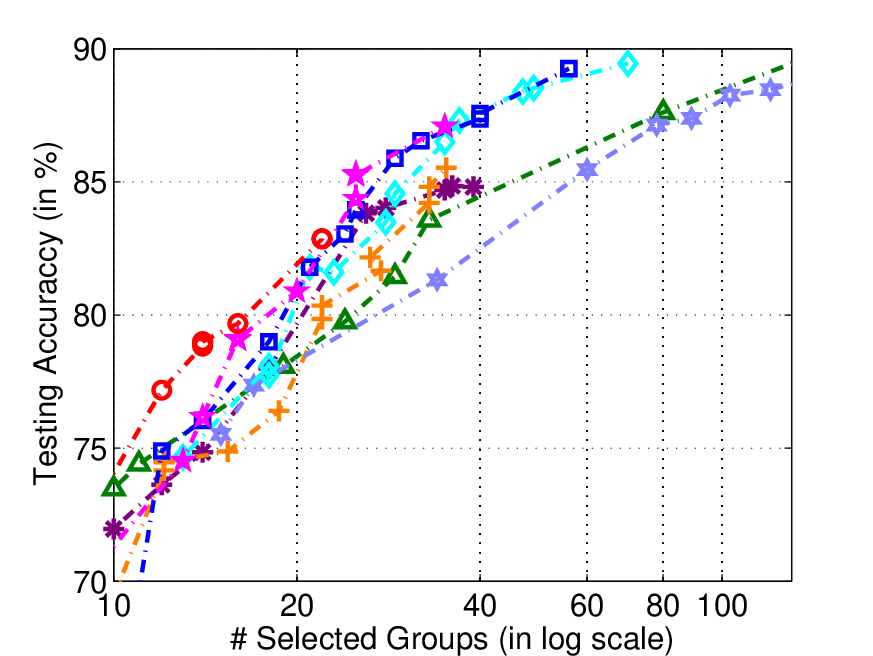}}
  \hspace{0.1in}
  \subfigure[{\tt rcv1-ADD}]{
    \label{fig:rcv1_real_group_acc} 
    \includegraphics[trim = 1.5mm 0mm 6mm 2mm,  clip,  width=2.35in ]{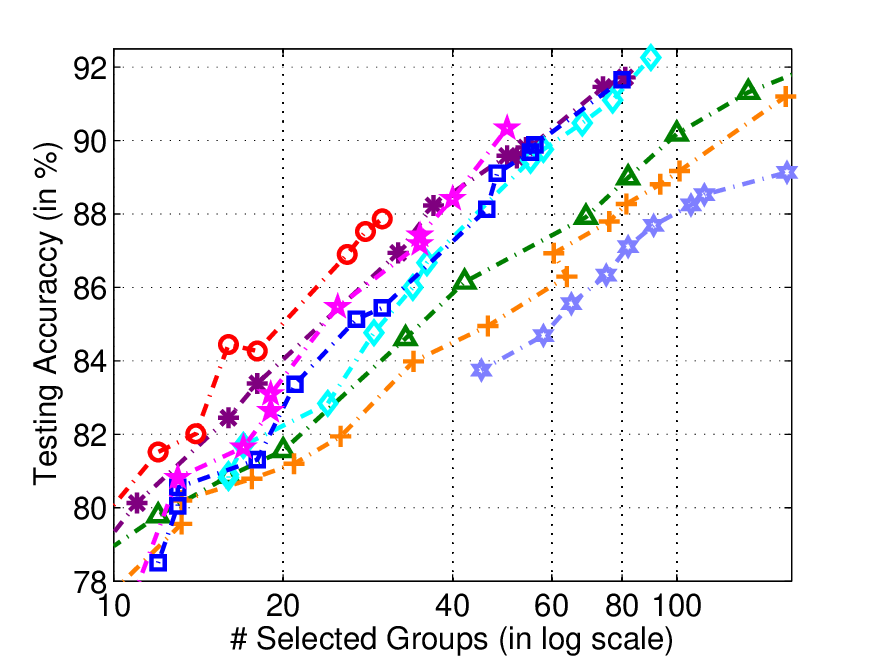}}\\
  \subfigure[{\tt aut-HIK}]{
    \includegraphics[trim = 1.5mm 0mm 6mm 2mm,  clip,
    width=2.40in]{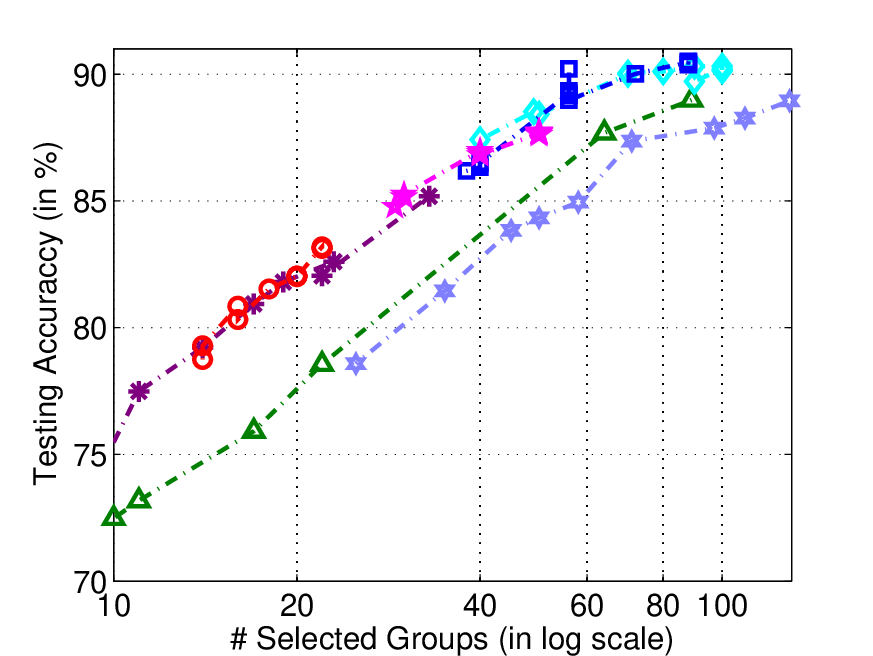}}
  \hspace{0.1in}
  \subfigure[{\tt rcv1-HIK}]{
    \includegraphics[trim = 1.5mm 0mm 6mm 2mm,  clip,  width=2.35in ]{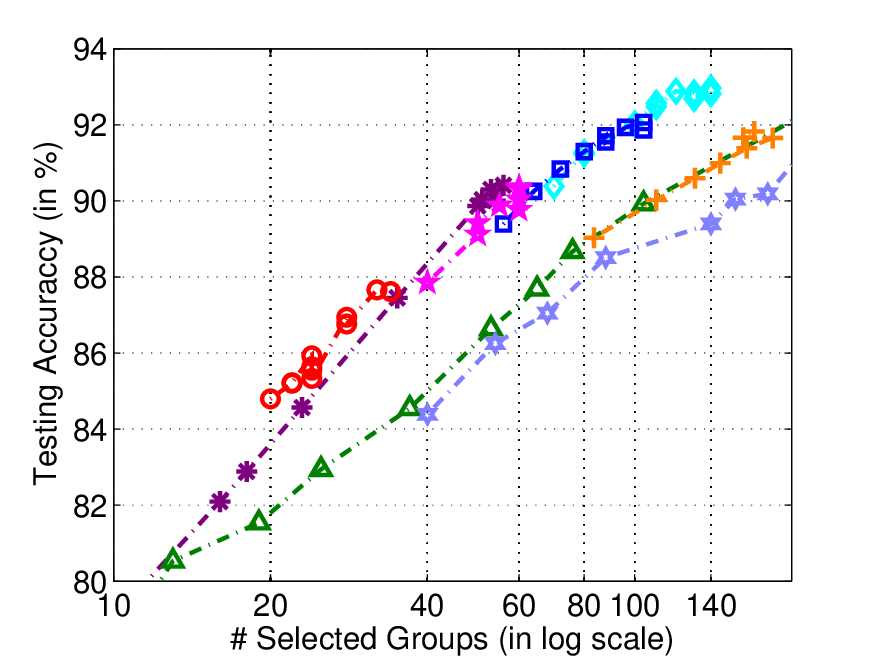}}
  \caption{Testing \aje{accuracies} on group feature selection tasks. The groups are generated by HIK
  or additive feature mappings.
  The results of {BCD} on {\tt aut-HIK} is not reported due to the heavy computational cost.}
  \label{fig:group_real_acc}
\vskip -0.20in
\end{figure*}

\begin{figure*}[h]
\center
    {\includegraphics[trim = 10mm 5.5cm 10mm 2.0cm,  clip,  width=0.74\textwidth]{group_large_legend.eps}}\vspace{-6mm} \\
  \centering
  \subfigure[{\tt aut-ADD}]{
    \label{fig:add_aut_time} 
    \includegraphics[trim = 1.5mm 0mm 6mm 2mm,  clip,  width=2.35in ]{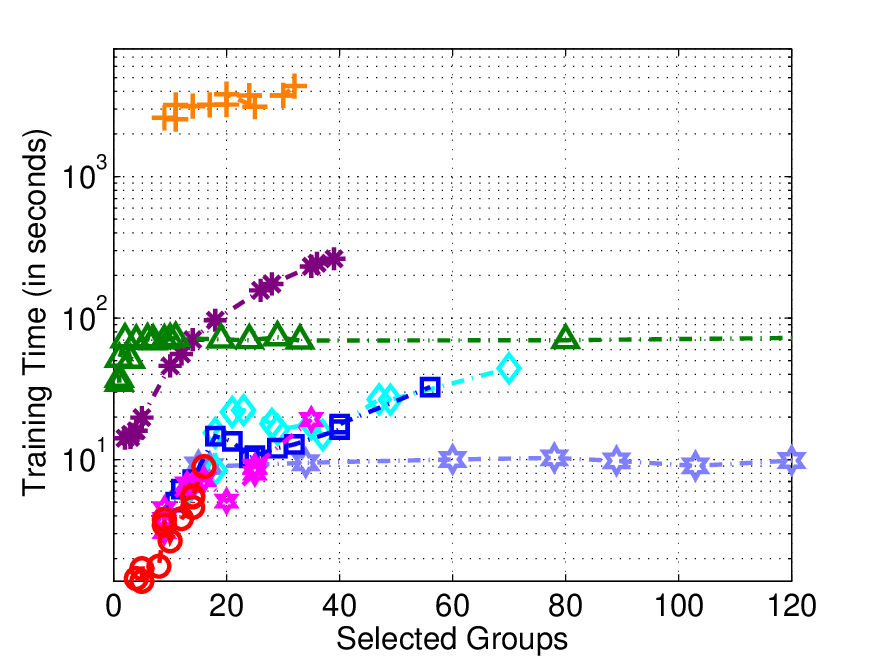}}
  \hspace{0.1in}
  \subfigure[{\tt rcv1-ADD}]{
    \label{fig:rcv1_aut__time} 
    \includegraphics[trim = 1.5mm 0mm 6mm 2mm,  clip,  width=2.35in ]{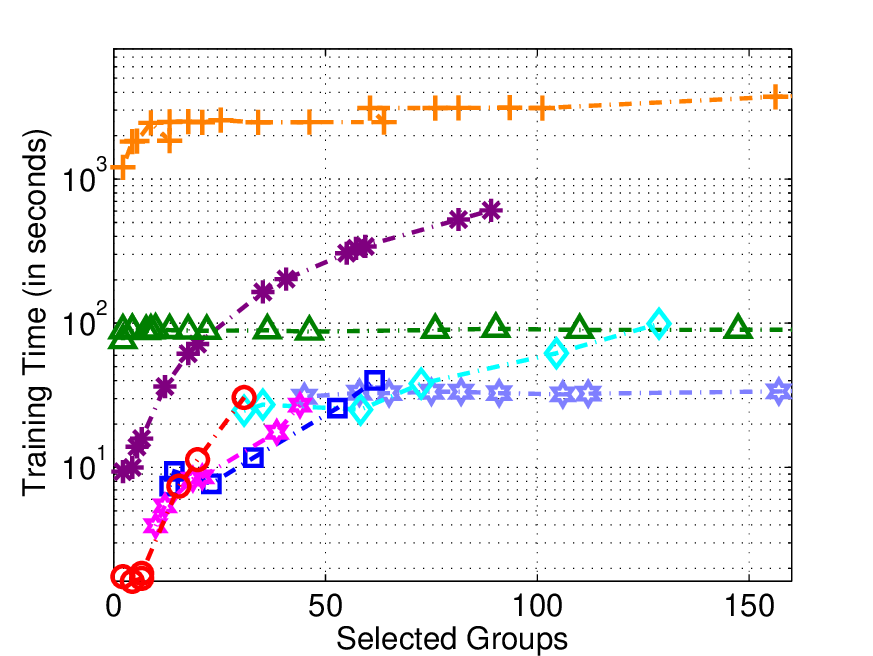}}\\
  \subfigure[{\tt aut-HIK}]{
    \label{fig:aut_hik_time} 
    \includegraphics[trim = 1.5mm 0mm 6mm 2mm,  clip,  width=2.35in ]{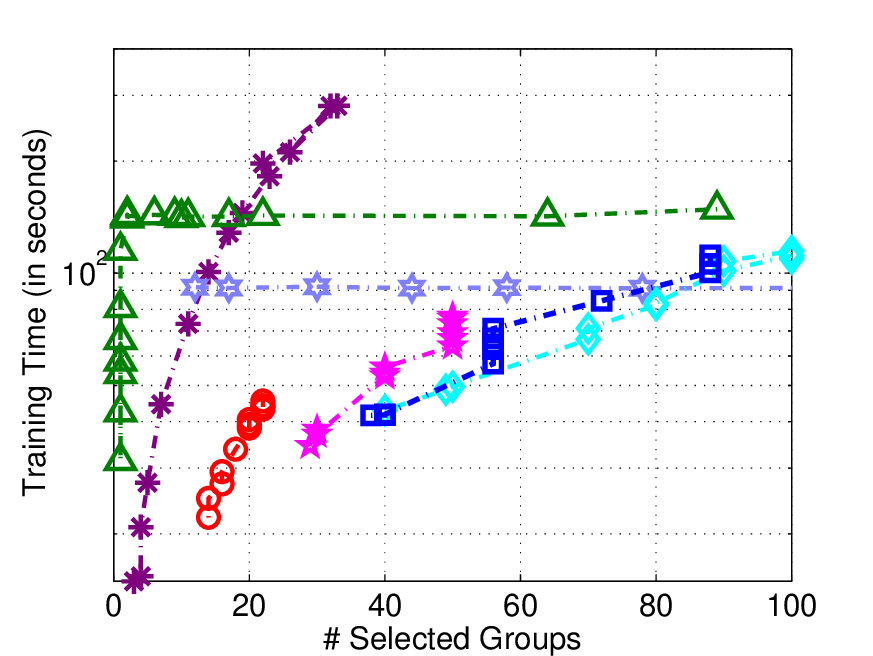}}
 \hspace{0.1in}
  \subfigure[{\tt  rcv1-HIK}]{
    \label{fig:rcv1_hik_time} 
    \includegraphics[trim = 1.5mm 0mm 6mm 2mm,  clip,  width=2.35in ]{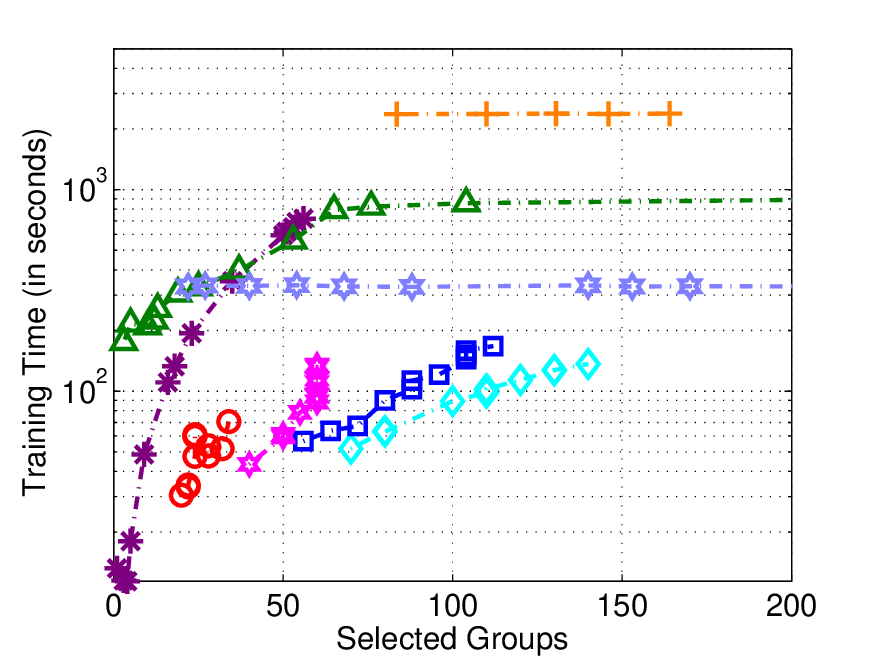}}
  \caption{Training \aje{times} on group feature selections. }
  \label{fig:group_real_time}
\vskip -0.15in
\end{figure*}

In this section, we verify the effectiveness of {FGM} for group
feature selection on two real-world datasets, namely\aje{,} {\tt aut-avn}
and {\tt rcv1}. In real-applications, the group prior of features
comes in different ways. In this paper,  we produce the feature
groups using the explicit kernel feature
expansions~\citep{wu2012efficient,Vedaldi2010}, where each original
feature is represented by a group of approximated features. Such
expansion can vastly improve the training efficiency of kernel
methods while \aje{maintaining} good approximation performance in many
applications, such as in computer vision~\citep{wu2012efficient}.
For simplicity, we only study the HIK kernel
expansion~\citep{wu2012efficient} and the additive Gaussian kernel
expansion \citep{Vedaldi2010}. In the experiments, for fair
\aje{comparison},  we pre-generate the explicit features for two
datasets. The details of the two datasets and the feature expansions
are listed Table~\ref{group_generate}. We can observe that\aje{,} after
feature expansion, the storage requirement dramatically increases.
Figure \ref{fig:group_real_acc} and \ref{fig:group_real_time} report
the testing accuracy and training time of different methods,
respectively.

From  Figure \ref{fig:group_real_acc}, {FGM} and the active set
method achieve superior performance over {FISTA}, {BCD} and
{UFO-MKL} in terms of testing accuracy. Moreover, from Figure
\ref{fig:group_real_time}, {FGM} \aje{has} much better efficiency than
the active set method. It is worth mentioning that,  due to the
unbearable storage requirement, the feature expansion cannot be
explicitly stored when dealing with \aje{ultrahigh-dimensional} \emph{Big
Data}. Accordingly, {FISTA} and {BCD}, which require the explicit
presentation of data, cannot work in such cases. However, the
proposed feature generating paradigm can effectively address this
computational issue since it only involves a sequence of small-scale
optimization problems.

\section{Conclusions}\label{sec:Discussion}

In this paper, an adaptive feature scaling (AFS) scheme has been
proposed to select the most informative features by introducing a
feature scaling vector $\d\in[0, 1]^p$ \aje{into} the input features. To
explicitly control the sparsity, we impose an $\ell_1$-norm
constraint $||\d||_1\leq B$ on $\d$, where $B$ represents the least
number of features to be selected. To address the resultant
non-convex problem, we \aje{first} transform it to a convex SIP problem
and then propose a Feature Generating Machine (FGM) to solve it.
FGM iteratively includes a group of informative features/groups
and then solves a much reduced MKL subproblem. FGM has been extended
to solve general MKL problems with additive kernels. The global
convergence of FGM has been proved.

 Compared with $\ell_1$-norm methods and other existing feature
selection methods, FGM has two major advantages. \aje{First}, with the
separate control of the complexity and sparsity, the feature
selection bias of $\ell_1$-norm methods can be greatly reduced by
FGM with proper stopping conditions. \aje{Second}, since only a small
subset of features or kernels are involved in the subproblem
optimization, {FGM} is particularly suitable for the \aje{ultrahigh-dimensional} feature selection task. Furthermore, to make FGM
scalable to \emph{Big Data}, we propose to solve the primal form of
the subproblem \aje{using} a modified APG method. Due to the new
optimization scheme, some efficient cache techniques are also
developed to improve the training efficiency. Accordingly, FGM can
efficiently tackle ultrahigh-dimensional \emph{Big Data} for
which most of the existing methods are \aje{unfeasible}. It is worth
mentioning that, by avoiding storing all the kernels with explicit
kernel feature expansions, FGM vastly reduces the unbearable memory
demands of general MKL learning tasks with many kernels. Thereafter, many intractable tasks \aje{using} previous MKL methods now become \aje{feasible}.

\aje{Last},  comprehensive experiments have been conducted to verify
the performance of the proposed methods on both linear feature
selection and group feature selection tasks. {Extensive experiments
on synthetic datasets and real-world datasets have demonstrated that
{FGM} archives superior performance compared with the baseline
methods in terms of testing accuracy and training efficiency. }

In this paper, the proposed methods have tackled \emph{Big Data}
problems with \aje{a} million training examples ($O(10^7)$) and 100 trillion
features ($O(10^{14})$). Recall that the subproblem of FGM can be
possibly solved through SGD methods. In the future, we will explore
SGD methods to further improve the training efficiency for bigger
data of ultra-large size where $n$ is in \aje{the} billion or trillion scale.

\section*{Acknowledgments}

We would like to acknowledge the valuable comments and useful
suggestions by the Action Editor and the four anonymous reviewers.
We would like to express our gratitude to Dr. Xinxing Xu and Dr.
Shijie Xiao for the proofreading and comments. This research was in
part supported by Singapore A*star under Grant SERC 112 280 4005.

\newpage

\section*{Appendix A. Proof of Theorem \ref{thm:primal_dual}}

\begin{proof}
The proof parallels the results in \citep{bach2004mkl} and \aje{uses} the
conic duality theory. Let $\Omega(\ww) =
\frac{1}{2}\left({\|\ww_h\|}{}\right)^2$ and define the cone $\mQ_B
= \{(\u,v)\in \R^{B+1}, \|\u\|_2\leq v\}$. Furthermore, let $z_h =
\|\ww_h\|$,   we have $\Omega(\ww) = \frac{1}{2} \left(\sum_{h=1}^t
{ \|\ww_t\|}\right)^2 = \frac{1}{2} z^2$ with $z = \sum_{h=1}^{t}
{z_h}{}$. Apparently, we have $z_h\geq0$ and $z\geq0$. \aje{Last},
problem (\ref{eq:l21}) can be transformed \aje{into} the following problem:
\begin{eqnarray}
\min\limits_{z,\ww} && \frac{1}{2}z^2 + P(\ww,b),~~~ \text{s.t.} ~
\sum\limits_{h=1}^{t} {z_h}{}\leq z, ~~~(\ww_t,z_h) \in \mQ_B,
\nonumber
\end{eqnarray}
where $\ww = [\ww_1', ..., \ww_t']'$. The Lagrangian function of
(\ref{eq:l21}) regarding the squared hinge loss can be written as \aje{follows}:
\begin{small}
\begin{eqnarray}\label{eq:fg_Lagrangian}
\!\!  \lefteqn{\L(z,\ww,\bxi,b,\ba,\gamma,\bzeta,\bvarpi)} \nonumber\\
\!\!&\!=\!&\! \frac{1}{2}z^2+ \frac{C}{2}
\sum_{i=1}^{n} \xi_i^2 - \sum_{i=1}^{n} \a_i
\left(y_{i}(\sum{}\ww_h'\x_{ih} -b) -1 + \xi_i\right) +
\gamma(\sum\limits_{h=1}^{t}  {z_h}-z)-
\sum\limits_{h=1}^{t}(\bzeta_h'\ww_h + \varpi_h z_h),\nonumber\\
\end{eqnarray}
\end{small}
where  $\ba$, $\gamma$, $\bzeta_t$ and $\varpi_t$ are the Lagrangian
dual variables to the corresponding constraints. The KKT condition
can be expressed as \aje{follows:}
\begin{equation}
   \begin{array}{ll}
\nabla_{z}\L = z-\gamma = 0 & \Rightarrow z= \gamma; \nonumber\\
\nabla_{z_h}\L ={\gamma}{}-\varpi_h = 0
&\Rightarrow \varpi_h= {\gamma}{};\nonumber\\
\nabla_{\ww_h}\L =
 - \sum_{i=1}^{n} \a_i y_{i}\x_{ih} - \bzeta_h= 0 &\Rightarrow  \bzeta_h
 =-
 \sum_{i=1}^{n} \a_i y_{i}\x_{ih};\nonumber\\
\nabla_{\xi_i}\L = C \xi_i - \a_i = 0& \Rightarrow \xi_i =
 \frac{\a_i}{C};\nonumber\\
\|\bzeta_h\|\leq \varpi_h  &
 \Rightarrow \|\bzeta_h\|\leq {\gamma} {};\nonumber\\
\nabla_{b}\L  = 0 &
 \Rightarrow \sum_{i=1}^n {\alpha_i}y_i = 0.\nonumber
   \end{array}
\end{equation}
By substituting all the above results into (\ref{eq:fg_Lagrangian}),
we  can {obtain} \aje{the following}
\begin{eqnarray}\label{eq:fg_Lg}
 { \L(z,\ww,\ba,\gamma,\bzeta,\bvarpi)} =
-\frac{1}{2}\gamma^2 - \frac{1}{2C} \ba'\ba +1'\ba.\nonumber
\end{eqnarray}
Hence the dual problem of the $\ell_{2,1}^2$-regularized problem
regarding squared hinge loss can be written as \aje{follows}:
\begin{eqnarray}\label{eq:fg_dual}
 {\max\limits_{\gamma,\ba}} &&-\frac{1}{2}\gamma^2-  \frac{1}{2C} \ba'\ba +1'\ba \nonumber\\
s.t  && {}\Big\|\sum_{i=1}^{n} \a_i y_{i}\x_{ih} \Big\|\leq \gamma,~~ {h=1,\cdots,t},\nonumber\\
  && \sum_{i=1}^n {\alpha_i} y_i = 0, \a_i \geq 0,~~ {i=1,\cdots,n}.\nonumber
\end{eqnarray}
Let $\theta =  \frac{1}{2}\gamma^2 +  \frac{1}{2C} \ba'\ba -\ba'\1
$, $\ww_h = \sum_{i=1}^{n} \a_i y_{i}\x_{ih}$ and
 $f(\ba,\d_h) = \frac{1}{2}
{}\|\ww_h\|^2+\frac{1}{2C}\ba'\ba  -\ba'\1$,  we have \aje{the following}
\begin{eqnarray}
 {\max\limits_{\theta,\ba}} &&-\theta, \nonumber\\
s.t  && f(\ba,\d_h) \leq  \theta,~~  {h=1,\cdots,t},\nonumber\\
  && \sum_{i=1}^n {\alpha_i} y_i = 0, \a_i \geq 0,~~  {i=1,\cdots,n}, \nonumber
\end{eqnarray}
which indeed is in the form of problem (\ref{eq:QCQP_sub}) by
letting $\A$ be the domain of $\ba$. This completes the proof and
brings the connection between the primal and dual formulation.

By defining $0\log(0) = 0$, with the similar derivation above, we
can obtain the dual form of (\ref{eq:l21}) regarding the logistic
loss. Let $\ww = [\ww_1', ..., \ww_t']'$, we have \aje{the following}
\begin{eqnarray}
\min\limits_{z,\ww} && \frac{1}{2}z^2 + P(\ww,b),~~~ \text{s.t.} ~
\sum\limits_{h=1}^{t} {z_h}{}\leq z, ~~~(\ww_t,z_h) \in \mQ_B.
\end{eqnarray}
The Lagrangian function of (\ref{eq:l21}) regarding logistic  loss
can be written as \aje{follows}:
\begin{small}
\begin{eqnarray}\label{eq:fg_Lagrangian_logistic}
 \lefteqn{\L(z,\ww,\bxi,b,\ba,\gamma,\bzeta,\bvarpi)} \nonumber\\&=& \frac{1}{2}z^2+ C\sum_{i=1}^{n} \log(1+\exp(\xi_i)) - \sum_{i=1}^{n} \a_i
\left(y_{i}(\sum{}\ww_h'\x_{ih} -b) + \xi_i\right)+
\gamma(\sum\limits_{h=1}^{t}  {z_h}-z)-
\sum\limits_{h=1}^{t}(\bzeta_h'\ww_h + \varpi_h z_h),\nonumber\\
\end{eqnarray}
\end{small}
where  $\ba$, $\gamma$, $\bzeta_t$ and $\varpi_t$ are the Lagrangian
dual variables to the corresponding constraints. The KKT condition
can be expressed as \aje{follows:}
\begin{equation}\label{eq:ktt_logistic}
   \begin{array}{ll}
\nabla_{z}\L = z-\gamma = 0 & \Rightarrow z= \gamma; \nonumber\\
\nabla_{z_h}\L ={\gamma}{}-\varpi_h = 0
&\Rightarrow \varpi_h= {\gamma}{};\nonumber\\
\nabla_{\ww_h}\L =
 - \sum_{i=1}^{n} \a_i y_{i}\x_{ih} - \bzeta_h= 0 &\Rightarrow  \bzeta_h
 =-
 \sum_{i=1}^{n} \a_i y_{i}\x_{ih};\nonumber\\
\nabla_{\xi_i}\L =  \frac{C\exp(\xi_i)}{1+\exp(\xi_i)} - \a_i = 0&
\Rightarrow \exp(\xi_i) =
 \frac{\a_i}{C-\a_i};\nonumber\\
\|\bzeta_h\|\leq \varpi_h  &
 \Rightarrow \|\bzeta_h\|\leq {\gamma} {};\nonumber\\
\nabla_{b}\L  = 0 &
 \Rightarrow \sum_{i=1}^n {\alpha_i}y_i = 0.\nonumber
   \end{array}
\end{equation}

By substituting all the above results into
(\ref{eq:fg_Lagrangian_logistic}), we  {obtain} \aje{the following}
\begin{eqnarray}
 { \L(z,\ww,\ba,\gamma,\bzeta,\bvarpi)} =
-\frac{1}{2}\gamma^2 -  \sum_{i=1}^{n} (C-\a_i)\log({C-\a_i}) -
\sum_{i=1}^{n} \a_i \log({\a_i}). \nonumber
\end{eqnarray}
The dual form of the $\ell_{2,1}^2$-regularized problem regarding
logistic loss can be written as \aje{follows}:
\begin{eqnarray}
 {\max\limits_{\gamma,\ba}} &&-\frac{1}{2}\gamma^2-  \sum_{i=1}^{n} (C-\a_i)\log({C-\a_i}) -
\sum_{i=1}^{n} \a_i \log({\a_i}) \nonumber\\
\text{s.t.}  && {}\Big\|\sum_{i=1}^{n} \a_i y_{i}\x_{ih} \Big\|\leq \gamma,~~ {h=1,\cdots,t},\nonumber\\
  && \sum_{i=1}^n {\alpha_i} y_i = 0, \a_i \geq 0,~~ {i=1,\cdots,n}.\nonumber
\end{eqnarray}
Let $\theta =  \frac{1}{2}\gamma^2 + \sum_{i=1}^{n}
(C-\a_i)\log({C-\a_i}) + \sum_{i=1}^{n} \a_i \log({\a_i}) $, $\ww_h
= \sum_{i=1}^{n} \a_i y_{i}\x_{ih}$,
 $f(\ba,\d_h) = \frac{1}{2}
{}\|\ww_h\|^2+\sum_{i=1}^{n} (C-\a_i)\log({C-\a_i}) + \sum_{i=1}^{n}
\a_i \log({\a_i})$, then we have \aje{the following}
\begin{eqnarray}
 {\max\limits_{\theta,\ba}} &&-\theta, \\
\text{s.t.}  && f(\ba,\d_h) \leq  \theta,~~  {h=1,\cdots,t},\nonumber\\
  && \sum_{i=1}^n {\alpha_i} y_i = 0, ~~0 \leq \a_i \leq C,~~  {i=1,\cdots,n}. \nonumber
\end{eqnarray}

\aje{Last},  according to the KKT conditon, we can easily recover the dual variable $\ba$ by $\a_i =
\frac{C\exp(\xi_i)}{1+\exp(\xi_i)}$.
This completes the proof.
\end{proof}

\section*{Appendix B. Proof of Theorem \ref{thm:apg_con}}
The proof parallels the results in \citep{beck2009fast} and
includes several lemmas. \aje{First}, we define a one variable
function $Q_{\tau_b}(\v,b,v_b)$ w.r.t. $b$ as \aje{follows:}
\begin{small}
\begin{eqnarray}\label{eqn:prox_b}
Q_{\tau_b}(\v,b,v_b) &=&P(\v,v_b)+\langle \nabla_{b} P(\v,v_b),b-v_b
\rangle +  \frac{\tau_b}{2}\|b-v_b\|^2,
\end{eqnarray}
\end{small}
where we abuse the operators $\langle, \rangle$ and $\|\cdot\|$ for
convenience.
\begin{lemma}\label{lemma:subgradient}
$S_{\tau}(\u,\v) = \arg\min_{\ww}Q_{\tau}(\ww,\v,v_b)$ is the
minimizer of problem (\ref{eqn:prox}) at point $\v$, if and only if
there exists $g(S_{\tau}(\u,\v)) \in \partial
\Omega(S_{\tau}(\u,\v))$, the subgradient of $\Omega(\ww)$ at
$S_{\tau}(\u,\v)$, such that
\begin{eqnarray}
 g(S_{\tau}(\u,\v))  + \tau (S_{\tau}(\u,\v)-\v) + \nabla P(\v) = \0.
\end{eqnarray}
\end{lemma}
\begin{proof}
The proof can be completed by the optimality condition of
$Q_{\tau}(\ww,\v,v_b)$ w.r.t. $\ww$.
\end{proof}
\begin{lemma}\label{lemma:improve}
Let $S_{\tau}(\u,\v) = \arg\min_{\ww}Q_{\tau}(\ww,\v,v_b)$ be the
minimizer of problem (\ref{eqn:prox}) at point $\v$, and
$S_{\tau_b}(b) = \arg\min_{b} Q_{\tau_b}(\v,b,v_b)$ be the minimizer
of problem (\ref{eqn:prox_b}) at point $v_b$. Due to the line search
in Algorithm \ref{alg:apg}, we have \aje{the following}
\begin{eqnarray}
&&F(S_{\tau}(\u,\v),v_b) \leq Q_{\tau}(S_{\tau}(\u,\v),\v,v_b). \nonumber\\
&&P(\v,S_{\tau_b}(v_b)) \leq Q_{\tau_b}(\v,S_{\tau_b}(v_b),v_b).
\nonumber
\end{eqnarray}
and
\begin{eqnarray}\label{eq:ineq1}
F(S_{\tau}(\u,\v),S_{\tau_b}(b)) \leq
Q_{\tau}(S_{\tau}(\u,\v),\v,v_b) +\langle \nabla_{b}
P(\v,v_b),S_{\tau_b}(b)-v_b \rangle +
\frac{\tau_b}{2}\|S_{\tau_b}(b)-v_b||^2.
\end{eqnarray}
Furthermore, for any $(\ww',b)'$ we have \aje{the following}
\begin{eqnarray}\label{eq:ineq2}
F(\ww,b) - F(S_{\tau}(\u,\v),S_{\tau_b}(b)) &\geq& \tau_b \langle
S_{\tau_b}(b)-v_b,v_b-b \rangle +
\frac{\tau_b}{2}\|S_{\tau_b}(b)-v_b||^2 \nonumber\\  &&+ \tau
\langle S_{\tau}(\u,\v)-\v,\v-\ww \rangle + \frac{\tau}{2}\|
S_{\tau}(\u,\v)-\v||^2.
\end{eqnarray}
\end{lemma}
\begin{proof}
We only prove the inequality (\ref{eq:ineq1}) and (\ref{eq:ineq2}).
\aje{First}, recall that in Algorithm \ref{alg:apg}, we update
$\ww$ and $b$ separately. It follows that
\begin{eqnarray*}
\lefteqn{F(S_{\tau}(\u,\v),S_{\tau_b}(b))} \\
&=& \Omega(S_{\tau}(\u,\v)) +
P(S_{\tau}(\u,\v),S_{\tau_b}(v_b)) \\
&\leq& \Omega(S_{\tau}(\u,\v)) +
Q_{\tau_b}(S_{\tau}(\u,\v),S_{\tau_b}(b),v_b) \\
&=&
\Omega(S_{\tau}(\u,\v)) + P(S_{\tau}(\u,\v),v_b)+\langle \nabla_{b}
P(\v,v_b),S_{\tau_b}(b)-v_b \rangle +
\frac{\tau_b}{2}\|S_{\tau_b}(b)-v_b||^2 \\
&=& F(S_{\tau}(\u,\v),v_b)
 +\langle \nabla_{b} P(\v,v_b),S_{\tau_b}(b)-v_b \rangle +
\frac{\tau_b}{2}\|S_{\tau_b}(b)-v_b||^2 \\
&\leq&
Q_{\tau}(S_{\tau}(\u,\v),\v,v_b) +\langle \nabla_{b}
P(\v,v_b),S_{\tau_b}(b)-v_b \rangle +
\frac{\tau_b}{2}\|S_{\tau_b}(b)-v_b||^2.
\end{eqnarray*}
 This proves the inequality
in (\ref{eq:ineq1}).

Now we prove  the inequality (\ref{eq:ineq2}). \aje{First}, since
both $P(\ww,b)$ and $\Omega(\ww)$ are convex functions, we have \aje{the following}
\begin{eqnarray}
&&P(\ww,b) \geq P(\v,v_b) + \langle\nabla P(\v), \ww-\v\rangle  +
\langle\nabla_{b} P(\v,v_b), b-v_b\rangle, \nonumber\\
&&\Omega(\ww)\geq \Omega(S_{\tau}(\u,\v)) +
\langle\ww-S_{\tau}(\u,\v), g(S_{\tau}(\bg,\v))\rangle, \nonumber
\end{eqnarray}
where  $g(S_{\tau}(\u,\v))$ \aje{is} the subgradient of $\Omega(\ww)$ at
point $S_{\tau}(\u,\v)$. Summing up the above  inequalities, we
obtain \aje{the following}
\begin{small}
\begin{eqnarray}
\lefteqn{F(\ww,b)}\nonumber\\
 &\geq& P(\v,v_b) + \langle\nabla P(\v), \ww-\v\rangle  +
\langle\nabla_{b} P(\v,v_b), b-v_b\rangle + \Omega(S_{\tau}(\u,\v))
+ \langle\ww-S_{\tau}(\u,\v), g(S_{\tau}(\bg,\v))\rangle.\nonumber\\
\end{eqnarray}
\end{small}
In addition, we have \aje{the following}
\begin{scriptsize}
\begin{eqnarray}
&&F(\ww,b) - \left(Q_{\tau}(S_{\tau}(\u,\v),\v,v_b) +\langle
\nabla_{b} P(\v,v_b),S_{\tau_b}(b)-v_b \rangle +
\frac{\tau_b}{2}\|S_{\tau_b}(b)-v_b||^2\right)\nonumber\\
&=&P(\v,v_b) + \langle\nabla P(\v), \ww-\v\rangle  +
\langle\nabla_{b} P(\v,v_b), b-v_b\rangle + \Omega(S_{\tau}(\u,\v))
+ \langle\ww-S_{\tau}(\u,\v), g(S_{\tau}(\bg,\v))\rangle, \nonumber\\
&&- \left(Q_{\tau}(S_{\tau}(\u,\v),\v,v_b) +\langle \nabla_{b}
P(\v,v_b),S_{\tau_b}(b)-v_b \rangle +
\frac{\tau_b}{2}\|S_{\tau_b}(b)-v_b||^2\right) \nonumber\\
&=& P(\v,v_b) + \langle\nabla P(\v), \ww-\v\rangle  +
\langle\nabla_{b} P(\v,v_b), b-v_b\rangle + \Omega(S_{\tau}(\u,\v))
+ \langle\ww-S_{\tau}(\u,\v), g(S_{\tau}(\bg,\v))\rangle, \nonumber\\
&&- \left( P(\v,v_b)+\langle \nabla P(\v),S_{\tau}(\u,\v)-\v \rangle
+ \Omega(S_{\tau}(\u,\v)) + \frac{\tau}{2}\|S_{\tau}(\u,\v)-\v||^2
+\langle \nabla_{b} P(\v,v_b),S_{\tau_b}(b)-v_b \rangle +
\frac{\tau_b}{2}\|S_{\tau_b}(b)-v_b||^2\right)\nonumber\\
&=&  \langle \nabla P(\v) + g(S_{\tau}(\bg,\v)),\ww -
S_{\tau}(\u,\v)\rangle -
\frac{\tau}{2}\|S_{\tau}(\u,\v)-\v||^2 \nonumber\\
&& + \langle\nabla_{b} P(\v,v_b), b-S_{\tau_b}(b)\rangle -
\frac{\tau_b}{2}\|S_{\tau_b}(b)-v_b||^2.
\end{eqnarray}
\end{scriptsize}
With the relation $ S_{\tau_b}(b) = b - \frac{\nabla_{b}
P(\v,v_b)}{\tau_b}$ and Lemma~\ref{lemma:subgradient}, we obtain \aje{the following}
\begin{eqnarray}
\lefteqn{F(\ww,b) - F(S_{\tau}(\u,\v),S_{\tau_b}(b) )} \nonumber\\
&\geq& F(\ww,b) -
\left(Q_{\tau}(S_{\tau}(\u,\v),\v,v_b) +\langle \nabla_{b}
P(\v,v_b),S_{\tau_b}(b)-v_b \rangle +
\frac{\tau_b}{2}\|S_{\tau_b}(b)-v_b||^2\right). \nonumber\\ &\geq&
\langle \nabla P(\v) + g(S_{\tau}(\bg,\v)),\ww -
S_{\tau}(\u,\v)\rangle - \frac{\tau}{2}\|S_{\tau}(\u,\v)-\v||^2 \nonumber\\
&& + \langle\nabla_{b} P(\v,v_b), b-S_{\tau_b}(b)\rangle -
\frac{\tau_b}{2}\|S_{\tau_b}(b)-v_b||^2. \nonumber\\
&=&  \tau \langle  \v - S_{\tau}(\u,\v),\ww - S_{\tau}(\u,\v)\rangle
- \frac{\tau}{2}\|S_{\tau}(\u,\v)-\v||^2  \nonumber\\ &&+
\tau_b\langle v_b- S_{\tau_b}(b), b-S_{\tau_b}(b)\rangle -
\frac{\tau_b}{2}\|S_{\tau_b}(b)-v_b||^2.\nonumber\\
&=&  \tau \langle    S_{\tau}(\u,\v) - \v, \v- \ww\rangle +
\frac{\tau}{2}\|S_{\tau}(\u,\v)-\v||^2 \nonumber\\ &&+ \tau_b\langle
S_{\tau_b}(b) - v_b , v_b - b\rangle +
\frac{\tau_b}{2}\|S_{\tau_b}(b)-v_b||^2.\nonumber
\end{eqnarray}
This completes the proof.
\end{proof}
\begin{lemma}\label{lemma:seq}
Let $L_{bt} = \sigma L_t$, where $\sigma>0$. Furthermore, let us
define
\begin{small}
 \begin{eqnarray}
&&\mu^k = F(\ww^k, b^k) - F(\ww^*,b^*),\nonumber\\
&&\bnu^k = \rho^k \ww^k - (\rho^k - 1)  \ww^{k-1} - \ww^*, \nonumber\\
&&\upsilon^k= \rho^k b^k - (\rho^k - 1)  b^{k-1} - b^*, \nonumber
\end{eqnarray}
\end{small}
and then the following relation holds:
 \begin{small}
\begin{eqnarray}\label{eq:seq}
\frac{2(\rho^{k})^2\mu^k}{L^{k}}-\frac{(\rho^{k+1})^2\mu^{k+1}}{L^{k+1}}\geq
\left(||\bnu^{k+1}||^2-||\bnu^{k}||^2\right) + \sigma
 \left((\upsilon^{k+1})^2-(\upsilon^{k})^2\right).
\end{eqnarray}
\end{small}
\end{lemma}
\begin{proof}
Notice that we have $\ww^{k+1} = S_{\tau}(\u,\v^{k+1})$ and $b^{k+1}
= S_{\tau_b}(v_b^{k+1})$.  By applying Lemma~\ref{lemma:improve},
let $\ww = \ww^k$, $\v = \v^{k+1}, \tau = L^{k+1}$, $b = b^k$, $v_b
= v_b^{k+1}, \tau_b = L_b^{k+1}$, we have \aje{the following}
\begin{small}
\begin{eqnarray}
2(\mu^k-\mu^{k+1})
&\geq&  L^{k+1}\left(||\ww^{k+1}-\v^{k+1}||^2+ 2\langle\ww^{k+1}-\v^{k+1}, \v^{k+1}-\ww^{k}\rangle\right) \nonumber\\
&&+ L_{b}^{k+1}\left(||b^{k+1}-v_b^{k+1}||^2+ 2\langle
b^{k+1}-v_b^{k+1}, v_b^{k+1}-b^{k}\rangle\right).\nonumber
\end{eqnarray}
\end{small}
Multiplying both sides by $(\rho^{k+1}-1)$, we obtain \aje{the following}
\begin{small}
\begin{eqnarray}
2(\rho^{k+1}-1)(\mu^k-\mu^{k+1})
&\geq&  L^{k+1}(\rho^{k+1}-1)\left(||\ww^{k+1}-\v^{k+1}||^2+ 2\langle\ww^{k+1}-\v^{k+1}, \v^{k+1}-\ww^{k}\rangle\right) \nonumber\\
&&+ L_{b}^{k+1}(\rho^{k+1}-1)\left(||b^{k+1}-v_b^{k+1}||^2+ 2\langle
b^{k+1}-v_b^{k+1}, v_b^{k+1}-b^{k}\rangle\right).\nonumber
\end{eqnarray}
\end{small}
Also,   let $\ww = \ww^*$, $\v = \v^{k+1}, \tau = L^{k+1}$, $b =
b^k$, $v_b = v_b^{k+1}$, and $\tau_b = L_b^{k+1}$, we have \aje{the following}
\begin{small}
\begin{eqnarray}
-2\mu^{k+1}
&\geq&  L^{k+1}\left(||\ww^{k+1}-\v^{k+1}||^2+ 2\langle\ww^{k+1}-\v^{k+1}, \v^{k+1}-\ww^{*}\rangle\right) \nonumber\\
&&+ L_{b}^{k+1}\left(||b^{k+1}-v_b^{k+1}||^2+ 2\langle
b^{k+1}-v_b^{k+1}, v_b^{k+1}-b^{*}\rangle\right).\nonumber
\end{eqnarray}
\end{small}
Summing up the above two inequalities, we get \aje{the following}
\begin{small}
\begin{eqnarray}
\lefteqn{2\left((\rho^{k+1}-1)\mu^k-\rho^{k+1}\mu^{k+1}\right)}\nonumber\\
&\geq&  L^{k+1}\left(\rho^{k+1}||\ww^{k+1}-\v^{k+1}||^2+ 2\langle\ww^{k+1}-\v^{k+1}, \rho^{k+1}\v^{k+1}-(\rho^{k+1}-1)\ww^{k}-\ww^{*}\rangle\right) \nonumber\\
&&+ L_{b}^{k+1}\left(\rho^{k+1}||b^{k+1}-v_b^{k+1}||^2+ 2\langle
b^{k+1}-v_b^{k+1},
\rho^{k+1}v_b^{k+1}-(\rho^{k+1}-1)b^{k}-b^{*}\rangle\right).\nonumber
\end{eqnarray}
\end{small}
Multiplying both sides by $\rho^{k+1}$, we obtain \aje{the following}
\begin{small}
\begin{eqnarray}
\lefteqn{2\left(\rho^{k+1}(\rho^{k+1}-1)\mu^k-(\rho^{k+1})^2\mu^{k+1}\right)}\nonumber\\
&\geq&  L^{k+1}\left((\rho^{k+1})^2||\ww^{k+1}-\v^{k+1}||^2+ 2\rho^{k+1}\langle\ww^{k+1}-\v^{k+1}, \rho^{k+1}\v^{k+1}-(\rho^{k+1}-1)\ww^{k}-\ww^{*}\rangle\right) \nonumber\\
&&+ L_{b}^{k+1}\left((\rho^{k+1})^2||b^{k+1}-v_b^{k+1}||^2+
2\rho^{k+1}\langle b^{k+1}-v_b^{k+1},
\rho^{k+1}v_b^{k+1}-(\rho^{k+1}-1)b^{k}-b^{*}\rangle\right).\nonumber
\end{eqnarray}
\end{small}
Since $(\rho^{k})^{2} = (\rho^{k+1})^{2} - \rho^{k+1}$,  it follows
that
\begin{small}
\begin{eqnarray}
\lefteqn{2\left((\rho^{k})^2\mu^k-(\rho^{k+1})^2\mu^{k+1}\right)}\nonumber\\
&\geq&  L^{k+1}\left((\rho^{k+1})^2||\ww^{k+1}-\v^{k+1}||^2+ 2\rho^{k+1}\langle\ww^{k+1}-\v^{k+1}, \rho^{k+1}\v^{k+1}-(\rho^{k+1}-1)\ww^{k}-\ww^{*}\rangle\right) \nonumber\\
&&+ L_{b}^{k+1}\left((\rho^{k+1})^2||b^{k+1}-v_b^{k+1}||^2+
2\rho^{k+1} \langle b^{k+1}-v_b^{k+1},
\rho^{k+1}v_b^{k+1}-(\rho^{k+1}-1)b^{k}-b^{*}\rangle\right).\nonumber
\end{eqnarray}
\end{small}
By applying the equality $||\u-\v||^2 + 2\langle\u-\v,\v-\w\rangle =
||\u- \w||^2 - ||\v-\w||^2 $, we have \aje{the following}
\begin{small}
\begin{eqnarray}
\lefteqn{2\left((\rho^{k})^2\mu^k-(\rho^{k+1})^2\mu^{k+1}\right)}\nonumber\\
&\geq&  L^{k+1}\left(||\rho^{k+1}\ww^{k+1}-(\rho^{k+1}-1)\ww^{k} -\ww^*||^2 - ||\rho^{k+1}\v^{k+1} - (\rho^{k+1}-1)\ww^{k}-\ww^*||^2\right) \nonumber\\
&&+ L_{b}^{k+1}\left(||\rho^{k+1}b^{k+1}-(\rho^{k+1}-1)b^{k}
-b^*||^2 - ||\rho^{k+1}v_b^{k+1} -
(\rho^{k+1}-1)b^{k}-b^*||^2\right).\nonumber
\end{eqnarray}
\end{small}
With $\rho^{k+1}\v^{k+1} = \rho^{k+1}\ww^k + (\rho^{k}-1)(\ww^k -
\ww^{k-1})$, $\rho^{k+1}v_b^{k+1} = \rho^{k+1}b^k + (\rho^{k}-1)(b^k
- b^{k-1})$ and the definition of $\bnu^{k}$, it follows that
\begin{small}
\begin{eqnarray}
2\left((\rho^{k})^2\mu^k-(\rho^{k+1})^2\mu^{k+1}\right) \geq L^{k+1}
\left(||\bnu^{k+1}||^2-||\bnu^{k}||^2\right) + L_{b}^{k+1}
\left((\upsilon^{k+1})^2-(\upsilon^{k})^2\right).\nonumber
\end{eqnarray}
\end{small}
Assuming that there exists a $\sigma> 0$
 such that $L_{b}^{k+1} = \sigma L^{k+1}$, we get \aje{the following}
 \begin{small}
\begin{eqnarray}
 \frac{2\left((\rho^{k})^2\mu^k-(\rho^{k+1})^2\mu^{k+1}\right)}{L^{k+1}}
\geq  \left(||\bnu^{k+1}||^2-||\bnu^{k}||^2\right) + \sigma
 \left((\upsilon^{k+1})^2-(\upsilon^{k})^2\right).\nonumber
\end{eqnarray}
\end{small}
Since $L^{k+1} \geq L^{k}$ and $L_{b}^{k+1} \geq L_{b}^k$, we have \aje{the following}
 \begin{small}
\begin{eqnarray}
\frac{2(\rho^{k})^2\mu^k}{L^{k}}-\frac{(\rho^{k+1})^2\mu^{k+1}}{L^{k+1}}\geq
\left(||\bnu^{k+1}||^2-||\bnu^{k}||^2\right) + \sigma
 \left((\upsilon^{k+1})^2-(\upsilon^{k})^2\right).\nonumber
\end{eqnarray}
\end{small}
This completes the proof.
\end{proof}

\aje{Last}, with Lemma \ref{lemma:seq}, following the proof of Theorem~4.4 in \citep{beck2009fast}, we can obtain that
\begin{small}
\begin{eqnarray*}
F(\ww^k,b^k) - F(\ww^*,b^*) \leq \frac{2
L^k||\ww^0-\ww^*||^2}{(k+1)^2} + \frac{2 \sigma
L^k(b^0-b^*)^2}{(k+1)^2} \leq  \frac{2
L_t||\ww^0-\ww^*||^2}{\eta(k+1)^2} + \frac{2
L_{bt}(b^0-b^*)^2}{\eta(k+1)^2}.
\end{eqnarray*}
\end{small}
This completes the proof.

\section*{Appendix C: Linear Convergence of Algorithm \ref{alg:apg} for the Logistic Loss}
In Algorithm \ref{alg:apg}, by fixing $\varrho^{k}= 1$, it is
reduced to the proximal gradient
method~\citep{nesterov2007gradient}, and it attains a linear
convergence rate for the logistic loss, if $\bX$ satisfies the
following \textit{Restricted Eigenvalue
Condition}~\citep{ZhangJMLR2010}:
\begin{deftn}\label{def:infsup}
~\citep{ZhangJMLR2010} Given an integer $\kappa>0$, a design matrix
$\bX$ is said to satisfy the \textit{Restricted Eigenvalue
Condition} at sparsity level $\kappa$, if there exists positive
constants $\gamma_{-}(\bX,\kappa)$ and $\gamma_{+}(\bX,\kappa)$ such
that
\begin{small}
\begin{eqnarray}
\gamma_{-}({\bX},\kappa) = \inf\left\{\frac{\ww^{\top}\bX^{\top}\bX\ww}{\ww^{\top}\ww}, \ww \neq  \0, ||\ww||_0 \leq \kappa\right\},\\
\gamma_{+}(\bX,\kappa) =
\sup\left\{\frac{\ww^{\top}\bX^{\top}\bX\ww}{\ww^{\top}\ww}, {\ww}
\neq \0, ||{\ww}||_0 \leq \kappa\right\}.
\end{eqnarray}
\end{small}
\end{deftn}
\begin{remark}
For the logistic loss, if $\gamma_{-}({\bX},tB)\geq \tau
>0$, Algorithm~\ref{alg:apg} with  $\varrho^{k}= 1$ attains a linear convergence rate.
\end{remark}
\begin{proof}
 Let $\xi_i = -y_i
(\sum_{h=1}^{t}\ww_h'\x_{ih}-b)$. According to~\citep{Yuan2011}, the
Hessian matrix for the logistic loss can be obtained by
\begin{small}
\begin{eqnarray*}\label{eq:strongly}
\nabla^2 P(\ww) = C \bX' \bD\bX,
\end{eqnarray*}
\end{small}

\vspace{-0.15in} \noindent where $\bD$ is a diagonal matrix with
diagonal element $\bD_{i,i} =
\frac{1}{1+\exp(\xi_i)}(1-\frac{1}{1+\exp(\xi_i)}) >0$. Apparently,
$\nabla^2 P(\ww, b)$ is upper bounded on a compact set due to the
existence of $\gamma_{+}(\bX,\kappa)$. Let $\sqrt{\bD}$ be the
square root  of $\bD$. Then if $\gamma_{-}({\bX},tB)\geq \tau
>0$, we have $\gamma_{-}({\sqrt{\bD}\bX},tB)>0$ due to
$\bD_{i,i}>0$. In other words, the logistic loss is strongly convex
if $\gamma_{-}({\bX},tB)>0$. Accordingly,  the linear convergence
rate can be achieved according to~\citep{nesterov2007gradient}.
\end{proof}

\section*{Appendix D: Proof of Proposition \ref{prop:Bnorm}}
\begin{proof}
Proof of argument (I): We prove it by contradiction. \aje{First},
suppose $\d^*$ is a minimizer and there exists an $l\in\{1\ldots
m\}$, such that $w_l = 0$ but $d^*_l>0$. Let $0< \epsilon <d^*_l $,
and choose one $j\in \{1\ldots m\}$ where $j\neq l$, such that
$|w_j|>0$. Define new solution $\widehat{\d}$ in the following way:
$$\widehat{d}_j= d^*_j+d^*_l-\epsilon, ~~\widehat{d}_l = \epsilon,~~~\textrm{and},$$
 $$~~ \widehat{d}_k = d^*_k, ~~~~  \forall k\in \{1\ldots m\}\backslash \{j,l\}.$$
 Then it is easy to check that $$\sum_{j=1}^m
\widehat{d_j}=\sum_{j=1}^m d^*_j\leq B.$$ In other words,
$\widehat{\d}$ is also a feasible point. However, since
$\widehat{d}_j = d^*_j+d^*_l-\epsilon \geq d^*_j$, it follows that
$$\frac{w_j^2}{\widehat{d}_j} < \frac{w_j^2}{{d}^*_j}.$$
Therefore, we have \aje{the following} $$\sum_{j=1}^{m} \frac{w_j^2}{\widehat{d}_j} <
\sum_{j=1}^{m} \frac{w_j^2}{d^*_j},$$ which \aje{contradicts} the
assumption that $\d^*$ is the minimizer.

On the other hand, if $|w_j|>0$ and $d^*_j=0$, by the definition,
$\frac{x^2_j}{0}=\infty$. As we expect to get the finite minimum, so
if $|w_j|>0$, we have $d^*_j>0$.

{(II): \aje{First}, the argument  holds trivially when $\|\w\|_0 =
{\kappa} \leq B$.

If $\|\w\|_0 = {\kappa}  > B$, without loss of generality, we assume
$|w_i|
> 0$  for the first ${\kappa}$ elements. From the argument (I), we have
$1\geq d_i >0$ for $i\in\{1\ldots\kappa\}$ and
$\sum_{i=1}^{{\kappa}}d_i \leq B$. Note that $\sum_{i=1}^{{\kappa}}
\frac{{w}_i^2}{d_i}$ is convex regarding $\d$. The minimization
problem can be written as \aje{follows}:
\begin{eqnarray}\label{eq:obj_var}
\min_{\d} \sum_{i=1}^{{\kappa}} \frac{{w}_i^2}{d_i},
~~~~~~\mathrm{s.t.} ~~~~ \sum_{i=1}^{{\kappa}}d_i \leq B, ~~d_i>0,
~~1-d_i \geq 0.
\end{eqnarray}
The KKT condition of this problem can be written as \aje{follows}:
\begin{eqnarray}
&&-w_i^2/d_i^2+ \gamma -\zeta_i +\nu_i = 0, \\
&&\zeta_i d_i =
0, \\
&& \nu_i(1-d_i) = 0, \label{eq:d_complementary} \\
&&\gamma(B-\sum_{i=1}^{\kappa}d_i)=0, \label{eq:B}\\
&&\gamma\geq0, \zeta_i \geq 0, \nu_i \geq 0, \forall ~i\in
\{1\ldots\kappa\},
\end{eqnarray}
where $\gamma, \zeta_i$ and $\nu_i$ are the dual variables for the
constraints $\sum_{i=1}^{{\kappa}}d_i \leq B$, $d_i>0 $ and $1-d_i
\geq 0$\aje{,} respectively. For those $d_i>0 $, we have $\zeta_i = 0$ for
$\forall i\in
 \{1\ldots{\kappa}\}$ due to the KKT condition. Accordingly, by the first equality in KKT
condition, we must have \aje{the following} $$d_i = |w_i|/ \sqrt{\gamma+ \nu_i},
~~~~\forall i \in
 \{1\ldots{\kappa}\}.$$
  Moreover, since $\sum^{\kappa}_{i=1} d_i
\leq B<\kappa$, there must exist some $d_i < 1$ with $\nu_i = 0$
(otherwise $\sum^{\kappa}_{i=1} d_i$ will be greater than $B$).
\ivor{Here $\nu_i = 0$  because of the condition
(\ref{eq:d_complementary})}. This observation implies that
$\gamma\neq 0$ since each $d_i$ is bounded. {Since $d_i\leq 1$, the
condition $\sqrt{\gamma + \nu_i} \geq \max\{|w_i|\}$ must hold for
$\forall i \in \{1\ldots\kappa\}$.} Furthermore, by the
complementary condition (\ref{eq:B}), we must have \aje{the following}
$$\sum_{i=1}^{\kappa}d_i =B.$$  By
substituting $d_i = |w_i|/ \sqrt{\gamma+ \nu_i}$ back to the
objective function of (\ref{eq:obj_var}), it becomes \aje{the following}
$$\sum_{i=1}^{\kappa} {|w_i|} {\sqrt{\gamma+\nu_i}}.$$
To get the minimum of the above function, we are required to set the
nonnegative $\nu_i$ as small as possible.

Now we complete the proof with the assumption
${\|\w\|_1}/\max\{|w_i|\} \geq {B}$. When setting $\nu_i = 0$, we
get $d_i = \frac{|w_i|}{\sqrt{\gamma}}$ and
$\sum_{i=1}^{\kappa}\frac{|w_i|}{\sqrt{\gamma}} = B$. It is easy to
check that ${\sqrt {\gamma}} = {\|\w\|_1}/{B}\geq \max\{|w_i|\}$ and
${d_i} = {B|w_i|}/{\|\w\|_1} \leq 1$}, which satisfy the KKT
condition. Therefore, the above $\d$ is an optimal solution.  This
completes the proof of the argument (II).

{(III):} {With the results of (II), if $\kappa\leq B$, we have 
$\sum_{j=1}^{m} \frac{w_j^2}{d_j}=\sum_{j=1}^{\kappa} w_j^2.$
Accordingly, we have $\|\w\|_B = \|\w\|_2$. And if $\kappa> B$ and
${\|\w\|_1}/\max\{w_j\} \geq {B}$, we have \aje{the following} \[\sum \frac{w_j^2}{d_j} =
\sum \frac{|w_j|}{d_j} |w_j| = \frac{\|\w\|_1}{B}\sum|w_j| =
\frac{(\|\w\|_1)^2}{B}.\] Hence we have $\|\w\|_B=\sqrt{\sum
\frac{w_j^2}{d_j}} = \frac{ \|\w\|_1}{\sqrt{B}}$.} This completes
the proof.
\end{proof}

\end{document}